\documentclass[letterpaper]{article}
\usepackage{amsmath}
\usepackage{amssymb}
\usepackage{natbib,alifeconf}  
\usepackage{url,hyperref,cleveref}
\usepackage{booktabs}
\usepackage{glossaries}
\usepackage{adjustbox}
\usepackage{xcolor}
\usepackage{enumitem}
\usepackage{epigraph}
\usepackage[T1]{fontenc}
\usepackage{tikz-cd}
\IfFileExists{ulem.sty}{\usepackage[normalem]{ulem}}{}
\usepackage{hyperref}

\usepackage[dvipsnames]{xcolor}\usepackage[draft,inline,nomargin]{fixme} \fxsetup{theme=color} 
\FXRegisterAuthor{cg}{cgg}{\color{RoyalBlue}CGG}

\title{ALIFE2024 template}

\title{A Matter of Time: Towards a General Theory of Agency}

\author{
    Amahury J. López-Díaz 
    \and
    Carlos Gershenson \\
    \mbox{}\\
    School of Systems Science and Industrial Engineering, Binghamton University, 4400 Vestal Pkwy E, Binghamton, NY 13902, USA\\
    alpez@binghamton.edu
} 

\begin{document}

\maketitle


\begin{abstract}
Agency is widely invoked in biology, cognitive science, artificial intelligence, and philosophy, yet its organizational basis, its empirical thresholds, and the operational criteria that distinguish it from other teleonomic terminology remain unsettled. Building on temporally parametrized $(F, A)$-systems, we propose a multidimensional theory of biological agency grounded in relational biology, physical biosemiotics, and process ontology. Our central claim is that the precarious physical realization of self-reference is necessarily diachronic; constitutive constraints act, decay, and are regenerated over distinct characteristic timescales. By temporalizing organizational diagrams ordered by specified relation deletion, we obtain a structural partial order, rather than an evolutionary ladder, that distinguishes four defeasible conditions: autonomy as internal regeneration of constitutive constraints under material openness; goal-directedness as viability-biased maintenance; agency as endogenous anticipatory modulation of organism--environment coupling; and open-endedness as reconstruction of the variables, measurement relations, effectors, and norms through which future viability is defined. We translate these distinctions into a closure-sensitive mechanistic workflow and a provisional profile of operational signatures for semantic closure, measurement--control complementarity, anticipatory modulation, affordance reconstruction, syntactic open-endedness, and viability-corrected skill acquisition. Markov blankets and active inference are treated as derived modeling tools, while Bickhard's interactivism clarifies anticipatory error and normativity. Across chemical, cellular, multicellular, and artificial systems, our framework turns agency from an all-or-none attribution into a falsifiable, scale-explicit research program. 
\end{abstract}

\section{Introduction} 
Molecular biology, synthetic biology, and origin-of-life research have been extraordinarily successful. We can characterize many of the molecular mechanisms through which living systems build and maintain themselves, and we can synthesize an expanding subset of life's chemical components. However, a central problem remains unresolved, we cannot yet build a fully autonomous cell from scratch \citep{roosth2019synthetic}. As \citet{deduve2002life} has emphasized, all life known to us is cellular, so the unresolved problem is not merely to reproduce isolated reactions but to explain what cells do as integrated organizations. The remaining gap is therefore not simply a missing chemical inventory, it concerns organization, temporality, and control: how material processes collectively produce the constraints through which the system distinguishes relevant from irrelevant perturbations and acts so as to preserve or transform its own conditions of existence  \citep{jaeger2024naturalizing}.

“Agency,” the crux of the present manuscript, is routinely invoked across biology, cognitive science, artificial life, artificial intelligence, and philosophy as if its meaning were self-evident. In practice, however, the term oscillates among several partially overlapping notions, from self-maintenance \citep{rosslenbroich2024agency}, adaptive control \citep{kelso2016self}, and goal-directedness \citep{ball2023organisms}, up to anticipation \citep{walsh2015organisms}, autonomy \citep{virenque2024agency}, and context-sensitive action \citep{egbert2023behaviour}. As a consequence, we have a proliferation of diverse partial accounts. Some explain how systems remain organized, others how they pursue attractors or viability conditions, others how they infer or predict, but no consensus yet exists regarding the threshold at which a merely self-organizing or goal-directed system becomes an agent. To characterize agency, it is necessary to clarify how it relates to other relevant terminology in the literature. 

To do so, we must first recognize that the modern discussion of agency inherits an older question: \emph{what genuinely distinguishes the living from the non-living?} Molecular genetics encouraged the view that conventional physical and chemical laws would be sufficient once the relevant mechanisms were known \citep{watson1953molecular, kendrew1967phage}. Early computational evolution already displayed nontrivial novelty and coevolution \citep{barricelli1962numerical}, while \citet{neumann1966theory} showed that open-ended construction requires a distinction between a description and the machinery that interprets it. Later, complex systems approaches shifted attention from molecular parts to nonlinear interaction, emergence, and self-organization \citep{capra2014systems}. These developments clarified indispensable aspects of living systems, but neither molecular inventory nor fixed-rule self-organization by itself explains life's historically contingent capacity to alter the constraints and possibilities of future evolution \citep{pattee1995artificial}. The relevant contrast is therefore not between “simple” and “complex” systems, but between systems confined to a fixed externally specified rule space and systems capable of materially reconstructing the constraints through which new functions become possible~\citep{cariani1989design}.

At the same time, we must notice that renewed attention to niche construction \citep{trappes2022individualized}, organism--environment reciprocity \citep{read2024animals}, multicellularity \citep{newman2025agency}, and minimal cognition \citep{lyon2020minimal} have made agency newly central rather than philosophically optional. This renewed interest has been productive, but it has also exposed a conceptual instability: biological agency is often treated either as an intuitive label for behaviors we do not yet fully understand and that cannot be experimentally falsifiable \citep{difrisco2025biological}, or as a primitive property inherent to life presupposed wherever living organisms exhibit flexibility, purposiveness, or problem-solving capacities \citep{rosslenbroich2024agency}. The result of such an ambiguity is a field rich in suggestive descriptions but comparatively poor in principled experimental thresholds. Currently, three broad approaches organize much of the present debate, and none is sufficient on its own.

The first broad approach treats agency as a primitive gift of life, an irreducible feature that appears once matter crosses some fuzzy biological threshold (e.g.,~\citet{maturana1980}, and more recently~\citet{moreno2015biological}). This move preserves the distinctiveness of living systems, but it risks collapsing into descriptive vitalism if it does not explain how agency could emerge gradually from simpler organized matter. This should not be conflated with rejecting every use of \emph{methodological vitalism}. In ~\citet{walsh2018towards}'s usage, “methodological vitalism” is a materialist research stance that grants organisms explanatory roles not captured by a purely externalist description; it is not a commitment to a vital substance. Our disagreement is therefore not with organism-centered methodology, but with any account that invokes agency without specifying the organization, interventions, and viability conditions by which the ascription could be supported or rejected. Fundamentally, our aim is not to reject the autonomy tradition, but to prevent agency from functioning as an unexplained primitive.

The second broad approach identifies agency with robust goal-directedness or adaptive self-stabilization (e.g.,~\citet{ashby1952design}, and more recently~\citet{barandiaran2009defining}). This account correctly emphasizes compensation, equifinality, and viability. However, persistence and attractor restoration are too permissive as sufficient conditions. A flame, a hurricane, a thermostat, and a cell can all display stable or recurrent behavior, but the norms governing their persistence have different origins \citep{mossio2017makes}. If agency is ascribed whenever an observer can identify a preferred state, the concept loses its ability to distinguish internally generated norms from functions imposed by an external user or explanatory perspective \citep{sepulveda2024enactive}.

The third broad approach universalizes agency by treating it as a consequence of free-energy minimization or expected-state maintenance (e.g.,~\citet{friston2009free}, and more recently~\citet{kirchhoff2018markov}). Such formalisms may offer useful descriptions, but they cannot by themselves identify which systems produce the boundaries, constraints, and measurement--control relations that make those descriptions biologically meaningful \citep{nave2025drive}. Pendulums, power grids, candles, and organisms may all admit probabilistic steady-state descriptions. However, the explanatory task is to determine when the relevant organization is internally regenerated, precarious, and normatively tied to the system's own continued existence \citep{bruineberg2022emperor}. 

When comparing the philosophical foundations of each of these approaches, the old disputes between computationalism and enactivism inevitably resurface. Mostly guided by the early insights from~\citet{turing1936computable} and~\citet{church1936note, church1936unsolvable}, computationalist approaches are right to insist that living and cognitive systems involve structured organization, informational constraint, and internally differentiated operations that cannot be reduced to undifferentiated flux \citep{ashby1947principles, Turing1950computing}. They preserve, in other words, the intuition that agency must involve some form of internal selectivity, mediation, or organization of possibilities \citep{fodor1975language, marr2010vision}. But old-fashioned classical computationalism pays for this strength by treating syntax as if it were sufficient, thereby sidelining embodiment, material turnover, measurement, control, and the historically contingent emergence of semantic relevance \citep{cariani1989design}.

Founded by~\citet{varela1991embodied}, and later radicalized by~\citet{thompson2001radical},
enactivist approaches, in contrast to computationalism, are right to insist that cognition and agency are inseparable from embodied self-maintenance, organism--environment coupling, and precarious autonomy \citep{Clark1997,villalobos2013enactive, newen20184E, gallagher2023embodied}. They preserve what computationalism often misses: meaning is not merely manipulated but enacted from within a self-producing organization \citep{thompson2010mind, di2022laying}. Yet, enactivist frameworks have often remained comparatively weaker on the question of formal articulation \citep{ward2017introduction}. They tell us why agency cannot be reduced to disembodied symbol manipulation, but they do not always specify what kind of internal organization is needed for anticipatory, selectively adaptive action to arise \citep{villalobos2017post}. 

We must then clarify that the point of this paper is not to choose one side against the other, but to show that both traditions isolate genuine features of agency while leaving a crucial gap between them. What is needed is a framework that preserves the organizational and inferential insight of computationalism without detaching it from material self-production, and that preserves the embodied and normative insight of enactivism without leaving agency at the level of suggestive metaphors. Following ~\citet{pattee2001physics}, we start by proposing that life is distinguished from non-life by semantic closure, a self-referential organization in which symbols do not merely stand for functions but participate in the construction of the very mechanisms that interpret them \citep{pattee2012evolving}. Our concrete relational scaffold is ~\citet{hofmeyr2021biochemically}'s $(F, A)$-systems, where $F$ denotes \emph{fabrication} and $A$ denotes \emph{assembly}; the model is formally introduced in Section~\nameref{sec:intercausality}  and temporally extended in Section~\nameref{sec:temporalFA}. By incorporating sensing and response into $(F, A)$-systems through environmental affordances, it is possible to represent a semantically closed organization that can be both autopoietic and adaptive, i.e., closed to efficient causation while remaining open to formal causation \citep{rosen1991life}. 

Deleting specified relations from an $(F,A)$-organization generates a partial order of organizational diagrams, from elemental reaction networks to self-constructing chemical systems with proactive capabilities \citep{lopez2025closing}. The order is structural---it records which relations are retained, not an evolutionary ladder, a taxonomic ranking, or a monotonic increase in complexity. That previous result nevertheless generates a deeper question. If relational models are intrinsically time-invariant, how can they capture the temporally extended character of living organization without collapsing back into state-space approaches with fixed transition functions? This is the point at which the present paper begins. Its concern is not the origin of semantic closure as such, but the consequences of a temporally unfolding semantically closed organization. The claim defended here is that once closure is temporally articulated, it yields not merely persistence, but a primitive anticipatory organization capable of guiding adaptive action.

It is important to note that the introduction of a temporal dimension is not a cosmetic extension. It is motivated by the nature of self-reference itself. As~\citet{varela1975calculus} proved, the poles of self-reference cannot be grasped simultaneously but only through an alternation in which operator and operand exchange roles; closure is not fully intelligible outside temporal unfolding \citep{abramsky2025open}. Likewise, as~\citet{louie2013more} showed, anticipation is already implicit in the relational topology of closed biological systems, and thus temporal parametrization should not be understood as importing anticipation from outside, but as revealing how a closed organization becomes operationally anticipatory when its constitutive processes unfold across distinct characteristic timescales \citep{lopez2025closing}. 

Once closure is formalized and temporally unfolded, it allows us to distinguish four notions that are often conflated in the literature. Autonomy concerns the self-production and maintenance of an organization under materially open and precarious conditions \citep{mossio2017makes}. Goal-directedness concerns the maintenance or attainment of viability-supporting organization, whether understood in terms of emergent autopoietic persistence \citep{veloz2021goals} or attractor-like compensation of perturbations \citep{heylighen2023meaning}. Agency begins, on the focal biological route defended here, when such goal-directed organization acquires an endogenous anticipatory structure that selectively modulates organism--environment coupling in light of possible futures \citep{newman2025agency}. Open-endedness concerns the stronger capacity of an anticipatory organization to reconstruct its own future space of possibilities rather than merely navigate within a fixed one \citep{erwin2017topology}. These distinctions avoid two symmetric mistakes. Not every autonomous or goal-directed system is yet an agent, but neither must agency appear all at once as an unexplained property of fully formed organisms. The relations among the four conditions are therefore treated as defeasible and partially ordered, with empirically important side cases, rather than as a universal staged continuum.


Our strategy is to construct a theory that is biologically anchored, formally explicit, and capable of yielding experimental thresholds. Rather than stipulating agency from the outset, we ask which organizational conditions are retained or lost under diagrammatic weakening, which conditions are ordered by inclusion, and which remain incomparable. First, we clarify the causal vocabulary of the paper by arguing why closure to efficient causation, organizational closure, and semantic closure can be treated as convergent descriptions of the same self-producing organization. Second, we explain why such self-referential closure cannot be understood outside temporal unfolding. Third, we formalize temporally parametrized relational organization and its redescription as a history-dependent Asynchronous Dynamic Bayesian Network whose joint distribution is not fixed once and for all, but depends on context, substrate, and prior interactions. Fourth, we derive weaker temporalized organizations and use their structural partial order to distinguish proto-agential precursors from robust agency without construing the order as a historical sequence. \emph{A posteriori}, we use these organizational profiles to distinguish autonomy, goal-directedness, agency, and open-endedness. Finally, we show how this framework paves the way for broader applications in multicellularity, synthetic biology, and neuronal organization, while also enabling a constrained and biologically grounded use of active inference, thereby clarifying the sense in which our account is “computationally enactivist.”

\section{Life's Intercausality}
\label{sec:intercausality}

Before explaining why agency requires temporal unfolding, we must clarify what kind of organization is being unfolded. Throughout this paper, we are going to use the expressions \emph{closure to efficient causation}, \emph{organizational closure}, and \emph{semantic closure} as closely related descriptions of the same underlying biological property: organisms are not merely systems that undergo transformations, but systems that produce, maintain, and transform the constraints by which their own transformations remain possible \citep{lopez2025closing}. This section makes that equivalence explicit. We first introduce Aristotle's four causes as causal roles rather than metaphysical substances. We then show how these roles are reinterpreted in relational biology and physical biosemiotics. Finally, we explain why organizational closure is a robust property from which a general theory of agency can legitimately begin.

Table~\ref{tab:four-causes} introduces Aristotle's four causes and how they are employed in the present manuscript. Here, they serve complementary explanatory functions and should not be treated as ontologically separate ingredients, nor assumed to map one-to-one onto molecular parts \citep{austin2021form}. In early chemical organizations, material and formal roles may be fused in informed material causes, while efficient and formal roles may be fused in informed efficient causes \citep{hofmeyr2018causation}. The point of this vocabulary is to prevent the efficient-causal description of a transformation from being mistaken for a complete explanation of why the relevant constraints exist, how they are maintained, and in what sense their effects are functional for the system.

\begin{table*}[ht!]
\centering
\caption{Aristotelian causal roles as used in this paper. The roles are assignments within an explanatory model, not four additional forces or substances. A single biological structure may occupy more than one role, and the assignments can change with scale and target phenomenon.}
\label{tab:four-causes}
\begin{adjustbox}{max width=\textwidth}
\begin{tabular}{p{0.14\textwidth}p{0.25\textwidth}p{0.26\textwidth}p{0.25\textwidth}}
\toprule
Role & Guiding question & Biological interpretation & Typical examples \\
\midrule
Material cause & What is transformed or consumed? & Substrates and flows that make the organization materially possible. & Metabolites, ions, macromolecules, membranes, energy carriers, environmental resources. \\
Efficient cause & What locally brings about or constrains a transformation? & Processes or relatively conserved constraints that channel physical dynamics on the timescale of interest. & Enzymes, transporters, receptors, molecular motors, catalytic assemblies. \\
Formal cause & What specifies which transformation is admissible or selected? & Templates, codes, geometries, organizations, or instruction-like constraints. Formal causes may be free-standing or materially embodied \citep{hofmeyr2018causation}. & Nucleic-acid sequences, binding specificity, spatial organization, regulatory architecture. \\
Final cause & In virtue of what contribution is the transformation functional for the organization? & The viability-relevant consequence through which a component contributes to maintaining the organization that makes the component possible. & Membrane repair, metabolic regulation, error correction, controlled motility. \\
\bottomrule
\end{tabular}
\end{adjustbox}
\end{table*}

Historically, modern scientific explanations often privilege efficient causation: one event, mechanism, or process brings about another. This strategy is powerful, but it becomes insufficient when the explanandum is a living organization \citep{juarrero1999dynamics}. In organisms, the relevant question is not only what causes a process to occur, but how the causes responsible for that process are themselves produced, maintained, and rendered functional within the organization \citep{deacon2011incomplete}. A purely efficient-causal description can tell us how a protein catalyzes a reaction, how a membrane channel transports ions, or how a motor protein performs work, but it does not by itself explain how these efficient causes belong to an organization that makes their continued operation possible. This is the reason~\citet{rosen1991life} reintroduced Aristotle's four causes into theoretical biology. 

Rosen's point was not to abandon physical explanation, but to prevent biological explanation from being reduced to a single causal register, as in physics. A living system requires material causes, because it is made from matter and sustained by flows of energy \citep{morowitz1979energy}; efficient causes, because transformations are carried out by concrete processes and constraints \citep{montevil2015biological}; formal causes, because these transformations are shaped by organization, specificity, and instruction-like patterns \citep{polanyi1968life, pattee2017physical}; and final causes, because the parts of an organism are intelligible in terms of their contribution to the maintenance of the whole \citep{weber2002life}. In this sense, the four causes should be understood as complementary causal modes needed to describe organized self-production, not as four independent substances or forces.

While it is true that Aristotelian causes allow us to describe the intercausality of living organisms, their generality prevents us from concretizing their biological role, often ending up being less clear-cut in the target system than in a particular model of it \citep{de2017hylomorphism}. As declared in Table~\ref{tab:four-causes}, the four causes will be reintroduced operationally as follows. The \emph{material cause} is the substrate or transformable matter, including metabolites, ions, macromolecules, membranes, energy carriers, and environmental resources. The \emph{efficient cause} encompasses any local process or constraint that brings about a transformation, such as enzymes, molecular motors, transporters, receptors, and other structures that channel physical dynamics. The \emph{formal cause} is the pattern, rule, template, code, or constraint that specifies which transformations are possible or admissible. The \emph{final cause} is not an externally imposed purpose, but the contribution of a component or process to the continued existence of the organization that makes it possible. 

This causal-role vocabulary is compatible with recent Neo-Aristotelian approaches without committing the present model to hylomorphism as a complete ontology. Formal-causal explanation can illuminate how organized capacities depend on the configuration and coordinated activity of material parts \citep{austin2021form}, and organizational realism can be continuous with mechanistic biology rather than opposed to it \citep{de2017hylomorphism}. At the same time, abstract closure diagrams do not by themselves establish how formal organization is materially realized in a particular system \citep{difrisco2014hylomorphism}. Relational diagrams therefore identify candidate causal roles and entailments; whether a concrete biological organization realizes them must be tested through material, temporal, and intervention-based evidence. When realizing the various causal roles in a living organism, one must pay attention to the final cause.

In the twentieth century, purposive vocabulary was often treated as scientifically suspect, yet cybernetics helped separate goal-directed organization from conscious intention or an external designer \citep{rosenblueth1943behavior}. In contemporary biology, much of what Aristotle called final causation is expressed through the language of \emph{function} \citep{cornish2020contrasting}; a component is \emph{functional} insofar as its activity contributes to maintaining or reproducing the organization in which that activity is possible \citep{mossio2017makes}. As argued by~\citet{nicholson2013organisms}, in a conventional artifact, final causation is typically external: the function of a hammer, thermostat, or engine is assigned by a user or designer. In an organism, by contrast, final causation is internally grounded; the products of component processes contribute to the maintenance of the system in which those same processes can continue to operate.

Finality in this paper is therefore intrinsic and organizational, not a claim that evolution moves toward a predetermined end. However, it is one diagnostic dimension among several, not by itself a sufficient binary test separating every organism from every possible machine \citep{mossio2017makes}. In presently conventional human-made machines, constitutive repair, goals, and semantic interpretation remain predominantly externally anchored. If organic bodies are “machines” of some kind, they therefore constitute a peculiar class of devices that human beings are not yet capable of building \citep{marques2014rise}; this is an empirical and design claim, not a claim of substrate-specific impossibility. To make teleological language legitimate in biology without invoking an external designer, relational biology notices that biological components are not merely caused by other components, they also contribute to the conditions under which those components remain causally efficacious \citep[p. 244]{rosen1991life}. 

The central move of relational biology is thus to ask what happens when efficient causes are not externally supplied. In an organism, the relevant efficient causes are generated and maintained by the organization itself \citep{rosen1986causal}. This is the core meaning of \emph{closure to efficient causation}: every component that functions as an efficient cause within the system is materially produced and maintained by other processes within that same system. To represent this quality, Rosen's $(M,R)$ notation abbreviates \emph{metabolism} and \emph{repair}, referring to metabolic transformation and the replacement of the components that realize it. An $(M,R)$-system is closed to efficient causation when the efficient causes required by the system are generated or maintained through other transformations within the same organization \citep{rosen1958representation, rosen1958relational}. Hofmeyr's $(F,A)$ extension replaces the historically narrow one-gene--one-enzyme reading with fabrication and assembly, making explicit how components can be manufactured, assembled, and reassembled while formal constraints remain revisable \citep{hofmeyr2018causation, hofmeyr2021biochemically}. Thus, agency cannot be grounded in mere persistence or energetic throughput, but in an organization capable of producing the constraints through which it continues to act \citep{pattee1971physical} .

The expression \emph{organizational closure} names the same idea at a more general level. It refers to a regime in which the constraints that channel the system's dynamics are themselves produced and maintained by the dynamics they constrain \citep{Varela1979Principles}. It is worth noting that organizational closure is more fundamental than cyclicity, feedback, or self-organization \citep{bourgine1992towards}. The system must not only persist; it must participate in the production of the constraints by which its persistence remains possible \citep{nave2025drive}. This criterion is intentionally more demanding than recurrence or attractor stability. The problem with calling every stable system an agent is not merely terminological; it erases the distinction between a system whose norms are generated by its own organization and a system whose apparent goal is assigned by an observer, designer, or external field \citep{varela1978being}. 

Our theory therefore concerns \emph{biological agency} proper, for which autonomy is an organizational ground, while allowing weaker \emph{proto-agential} precursors before full closure. Viruses and prions illustrate why this distinction matters. Considered in isolation, they do not regenerate the complete set of constraints required for their continuation and are therefore not autonomous entities in the present sense. Within host-coupled organizations they can nevertheless participate in causally and evolutionarily significant agent-like processes. Furthermore, organizational closure is a substrate-neutral quality. A synthetic nonliving system could count as an agent if it realized comparable internally regenerated closure, viability norms, and anticipatory measurement--control relations \citep{beer2023theoretical}. Figures~\ref{fig:causal-role-profile} and~\ref{fig:threshold-poset} later summarize these relations as organizational profiles and a partial order.

Physical biosemiotics redescribes this same organizational fact in terms of substrate vehicles (“symbols”), non-holonomic constraints, measurement, and control, allowing us to better describe the interaction and origin of Aristotelian causal roles in life \citep{pattee2007laws, pattee2012causation}. Importantly, a symbol does not become biologically meaningful merely by existing as a physical pattern. It becomes meaningful only when embedded in an organization that can interpret it and use it to control dynamics \citep{pattee2012does, pattee1969physical}. Clearly, a purely materialistic description is insufficient. In the cell, genetic sequences function as relatively rate-independent symbolic structures, while the molecular machinery that reads, interprets, and acts upon those sequences operates through rate-dependent physical dynamics. Pattee's \emph{epistemic cut} names the necessary descriptive distinction between relatively rate-independent symbolic structures and the rate-dependent physical dynamics that interpret and act on them~\citep{pattee2001physics}. It is not an ontological division between two substances. The cut is “epistemic” because measurement and control require different functional descriptions even though both are physically realized \citep{pattee2012cell}. A nucleotide sequence does not become a symbol merely by having a pattern; it becomes symbolically efficacious only in an organization that constructs and maintains the machinery through which that pattern is measured, interpreted, and converted into action. Semantic closure occurs when the symbol vehicles help construct their interpreters and those interpreters, in turn, maintain and use the symbol vehicles \citep{pattee2001physics}. 

In this sense, both semantic closure and closure to efficient causation converge. A system is semantically closed when its symbols participate in the construction of the very mechanisms that interpret those symbols \citep{pattee2012evolving}. The genome helps build the machinery that reads the genome, the interpretive machinery gives functional significance to the genome, and both are embedded in the continued self-production of the cell \citep{pattee2012cell}. Thus, semantic closure is the biosemiotic description of organizational closure when the organization contains symbol-mediated measurement and control. This equivalence should be understood carefully. Closure to efficient causation, organizational closure, and semantic closure come from different theoretical traditions and emphasize different aspects of living organization. In the class of systems considered here, however, these descriptions converge on the same core property: the system produces and maintains the conditions under which its own components can function as meaningful constraints \citep{lopez2025closing}. Figure~\ref{fig:comparative-profiles} returns to the corresponding machine, host-dependent, biological, and collective profiles in the Discussion.

There is also a methodological reason to begin from organizational closure. As stated by~\citet{wimsatt2007re}, a property is tangible when it can be accessed, derived, or detected in multiple independent ways. Organizational closure satisfies this criterion: homeostats \citep{ashby1949electronic}, autopoietic structures \citep{VarelaEtAl1974, maturana1980}, Kantian wholes \citep{kauffman2000investigations, kauffman2019world}, $(M, R)$-systems \citep{rosen1958relational, rosen1958representation}, chemotons \citep{ganti2003chemoton}, hypercycles \citep{eigen2012hypercycle}, kinematic automata \citep{neumann1966theory}, closure-of-constraints approaches \citep{mossio2010organisational, montevil2015biological} and semantic-closure accounts \citep{barbieri1981ribotype, pattee2012clues} differ in formalism, historical origin, and biological emphasis, yet they repeatedly converge on the idea that living systems are self-producing organizations whose constraints are not merely externally imposed \citep{letelier2011homme, cornish2020contrasting}. 

This convergence does not mean that all these models are equivalent in detail. They are not, but the recurrence of organizational closure across these independent traditions suggests that it is not a parochial artifact of one formalism. It is a \emph{robust} feature for biological explanation \citep{jaeger2026reengineering}. For this reason, the general theory of agency proposed here begins from the organizational property that multiple theories of life have independently found necessary. This point also helps us to avoid descriptive vitalism. Agency should not be treated as a primitive gift that appears mysteriously once chemistry becomes biological \citep{walsh2018towards}. If organizational closure is robustly characterizable, then we can ask how different degrees and kinds of closure support different degrees and kinds of agency. Thus, rather than assuming agency \emph{ab initio}, we now ask how a closed organization must be temporally structured for anticipatory adaptive action to become possible.

\section{Time and the Physical Realization of Self-Reference} 
\label{sec:selfreference}
Having clarified the causal meaning of organizational closure, we can now ask why time enters the picture at all. One of the biggest motivations for~\citet{rosen1991life} to work with time-invariant category theory diagrams was to capture the organizational structure of living systems, rather than focusing on rate-dependent, mechanistic, dynamical descriptions \citep{varenne2013mathematical, lane2024robert}. If relational models were introduced precisely because they avoid the limitations of state-space dynamics with fixed transition functions, then temporal parametrization must not be treated as a return to the very framework relational biology was designed to overcome. The problem, rather, is to show that temporality is already implicit in self-referential organization. In this section, we argue that temporally unfolding a semantically closed organization is not an external dynamical supplement to a static diagram, but the minimal way in which organizational closure becomes operationally intelligible.

In Rosen's original terms, what matters for living organization is not only how a material system changes state, but how the efficient causes responsible for those changes are themselves produced and maintained within the system \citep{rosen1991life}. This is why relational biology provides a natural language for semantic closure (i.e., closure to efficient causation), because it represents the organization of production, replacement, and constraint without reducing it to a single transition function. Still, this strength also creates an apparent difficulty. A relational diagram can specify the organization of self-production, but organisms do not exist as timeless diagrams \citep{korbak2023self}. They grow, repair, develop, adapt, and evolve; their constraints persist only by being continually regenerated, and their functional roles are maintained only through processes that unfold at different rates. The static diagrams of relational biology describe invariant entailment relations. They do not imply that the material processes realizing those relations occur simultaneously or persist without maintenance \citep{louie2013more}.

The relevant question then is how a materially precarious organization can \emph{realize} self-construction when its components act, decay, and are replaced at different rates \citep{abramsky2025open}. The challenge, therefore, is not simply to “add dynamics” to relational biology and its closed diagrams. Doing so in the conventional way would risk returning to state-space models in which the relevant possibilities are already fixed in advance \citep{kauffman2021world}. Instead, we need to ask whether temporality can be reintroduced from within the logic of closure itself. A first approach to this problem was proposed by~\citet{korbak2023self}, who suggested the use of probabilistic graphical models to represent temporally unfolded $(M, R)$-systems. However, the author did not adequately justify the introduction of a temporal dimension beyond the need for representation. As we will elaborately explain in the following, here the key to this move is impredicativity.

\citet{varela1975calculus}'s \emph{calculus for self-reference} offers an illuminating operational image of this problem. In~\citet{SpencerBrown1969Laws}'s calculus of indications, drawing a distinction generates an indicational space: there is a marked side, an unmarked side, and a boundary whose crossing makes indication possible. Varela's move was then to ask what happens when such a distinction is re-entered into the very space that it distinguishes \citep{varela1975calculus}. Re-entry in this context is not merely another operation within a pre-given space, but a peculiar procedure in which an indication becomes implicated in the domain that it indicates. For this reason, as~\citet{reichel2011snakes} emphasizes, whereas distinction produces space, re-entry produces time; the system cannot be grasped as a completed spatial form without also being grasped as a process of return, recurrence, or self-indication.


\citet{varela1975calculus}'s \emph{autonomous state} captures this ambiguity. It can be read spatially, as a compact notation for re-entry, but it can also be read operationally, as an injunction to re-enter; not merely the name of a self-crossing form, but the demand that the form continue crossing itself \citep{reichel2011snakes}. This double reading is biologically important. An autonomous organization does not first exist as a completed object and then later interact with an environment. Rather, through its own operations, it produces a boundary, distinguishes itself from what it is not, and thereby brings forth a domain of possible perturbations. This is the sense in which \emph{structural coupling} helps clarify the question: an organism is materially and energetically open, but organizationally closed in the sense that perturbations acquire significance only through the organization that the system itself maintains \citep{Varela1979Principles, maturana1980}.


The above supports a processual reading of organizational closure, but it does not prove that every formal self-reference is intrinsically temporal. Nor does it, by itself, establish process ontology. The stronger biological argument comes from precariousness: a living constraint is effective only over a characteristic interval and must be regenerated by other processes before it ceases to function~\citep{mossio2013emergence}. The physical realization of closure is therefore diachronic even when its relational description is time-invariant. This observation is central for avoiding a common misunderstanding.~\citet{varela1975calculus}'s calculus is not simply a three-valued logic, nor a complete biological theory, nor a recipe for constructing autonomous systems \citep{reichel2011snakes}. Its importance lies in providing a formal notation for the relation between a life's structure and its dynamic unfoldment \citep{Varela1979Principles}. The point is not that living systems literally compute Varela's calculus, but that the calculus displays a structural feature that any adequate theory of autonomy must respect: a self-producing unity cannot be represented only as a static object, because its identity depends on the recurrence of the operations that continually produce it.

This observation allows us to clarify in what sense organizationally closed systems must be temporally unfolded. Temporality is added because impredicativity must be enacted by material processes that are never all present in an unchanging form \citep{jaeger2023fourth}. A catalyst acts on one timescale, is fabricated or repaired on another, and is coordinated with environmental sensing and response on still others. A temporal parametrization records these heterogeneous intervals of action and regeneration, turning an invariant diagram of organizational dependence into a model of how the organization remains coherent through turnover \citep{varela1975calculus, reichel2011snakes}. Therefore, time is not imposed on closure from the outside---as if one first had a complete organization and then decided to animate it, just as~\citet{korbak2023self} did. Rather, once impredicativity is achieved via semantic closure, the organization must be unfolded in time in order to be understood as an organization at all \citep{Varela1979Principles}. A closed relational model is therefore not a frozen picture of life, but a compressed representation of a circular process whose intelligibility depends on temporal traversal.

The foregoing allows us to clarify the status of the temporal parameterization of~\citet{korbak2023self} and of our extension thereof to $(F, A)$-systems \citep{lopez2025closing}. A thermostat, an autocatalytic loop, or a chemical cycle may be represented as a loop of transformations, but closure to efficient causation requires more than the recurrence of processes \citep{cardenas2010closure}. It requires that the constraints or efficient causes that make the processes possible are themselves produced and maintained within the organization \citep{montevil2015biological}. In this sense, closure is not only circular causality, but inter-level and precariously self-maintenance \citep{nave2025drive}. From this perspective, the arrows of a relational diagram should be read as a compressed account of processual dependence; they do not merely indicate abstract mappings, but processes whose organizational roles are maintained only across characteristic temporal scales. Temporal unfolding therefore makes explicit what the diagram presupposes: that closure is an achievement sustained through the continual regeneration of constraints \citep{mossio2010organisational}.

The connection between self-construction and time also highlights the connection of organizational closure and anticipation.~\citet{rosen1978anticipatory} defined anticipatory systems as systems that contain internal predictive models of themselves and their environments, allowing present behavior to depend on modeled future states.~\citet{louie2012anticipation} later developed this insight in relational terms, showing that anticipation is not an accidental add-on to living systems but is deeply connected to the topology of relational organization. If a closed organization contains, in some form, the conditions of its own continued production, then it already contains a primitive relation between present organization and future viability \citep{louie2013more}. Temporal parametrization does not create this topology \emph{ex nihilo}, but proposes one way of realizing it: processes with different characteristic lags constrain one another asynchronously, allowing current measurements and internally maintained states to bias later viable trajectories. 

Anticipation here is therefore neither clairvoyance nor a separately installed predictor. It is the acquisition of a historically maintained endogenous organization in which present processes condition future possibilities. Thus, anticipation is not a mysterious reversal of causality, but the organization-mediated use of present states, traces, and constraints to bias future trajectories before perturbations have fully unfolded \citep{louie2010robert}. This is the conceptual bridge to the next section. If self-construction requires temporal unfolding, and if temporal unfolding exposes the anticipatory structure implicit in semantic closure, then the next task is to formalize this unfolding. We therefore turn to temporally parametrized $(F, A)$-systems, where each constitutive mapping is associated with a characteristic timescale and the resulting organization can be redesignated as a non-fixed, history-dependent Asynchronous Dynamic Bayesian Network.  

\section{Temporally Parametrized \texorpdfstring{$(F,A)$}{(F, A)}-systems} 
\label{sec:temporalFA}
The previous section established that self-referential closure cannot be fully understood as a static relation. If a living organization is both product and producer, both constraint and constrained process, then its relational structure must be unfolded across the characteristic timescales through which its constitutive processes are produced, maintained, and transformed. The aim of this section is to make this move explicit for Hofmeyr's \((F,A)\)-systems. To do this, we begin by clarifying the mathematical conventions we are using from this point onward. Then, rather than simply repeating the full construction developed in~\citet{lopez2025closing}, we summarize the formal steps needed for the present argument: first, we recall why \((F,A)\)-systems provide a biologically realizable model of semantic closure; second, we extend them to include environmental sensing and response; third, we assign characteristic timescales to their constitutive mappings; and fourth, we redescribe the resulting temporal organization as a history-dependent Asynchronous Dynamic Bayesian Network (ADBN).

\subsection{Mathematical status and notation}
\label{subsec:math-status}
\begin{figure*}[t]
\centering
\includegraphics[width=0.96\textwidth]{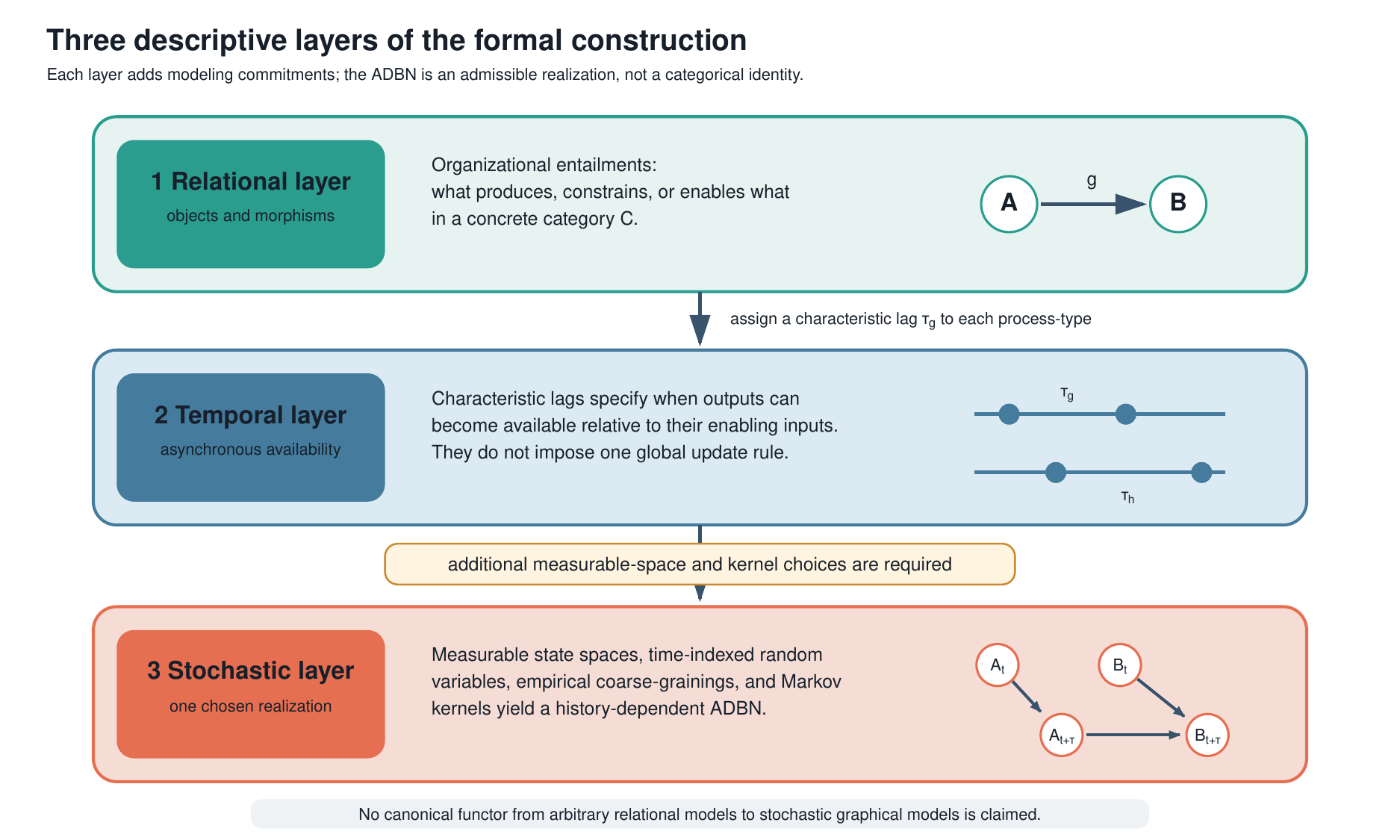}
\caption{Three descriptive layers of the construction. The relational diagram specifies organizational entailments; temporal parametrization assigns characteristic lags without imposing one global update rule; a stochastic realization then chooses measurable spaces, time-indexed random variables, and Markov kernels. Additional measurable-space and kernel choices are required, so the ADBN is one admissible realization rather than a categorical identity.}
\label{fig:formal-layers}
\end{figure*}

As illustrated in Figure \ref{fig:formal-layers}, three descriptive layers must be kept distinct.

\begin{enumerate}[label=(\roman*)]
\item \emph{Relational layer.} Capital Roman letters $A,B,C,\ldots$ denote objects in a concrete category $\mathcal C$, and Greek letters $\phi,\psi,\theta,\alpha,\beta$ denote admissible morphisms or process-types. At this layer, the diagrams specify organizational entailments and do not yet define random variables.
\item \emph{Temporal layer.} Each process-type $g$ is assigned a characteristic lag $\tau_g$. This indexing specifies when an output can become available relative to the inputs and constraints that enable it; it is not a global synchronous update rule.
\item \emph{Stochastic realization.} To obtain a probabilistic graphical model, each relational object $A$ is represented by a measurable state space $\mathcal X_A$, and a time-indexed random variable $\mathbf A_t:\Omega\rightarrow\mathcal X_A$ records a chosen empirical coarse-graining of that object at time $t$. Lower-case symbols $a_t\in\mathcal X_A$ denote realizations. Conditional dependencies are represented by Markov kernels chosen to respect the parent relations and lags specified by the temporalized diagram.
\end{enumerate}

The third layer does not follow functorially from the first two without additional assumptions. In particular, this paper does not establish a canonical functor from arbitrary Rosen--Hofmeyr relational models to a Markov category such as \textsc{FinStoch}. The ADBN should therefore be read as \emph{one admissible stochastic realization} of the temporal dependency structure, not as a theorem identifying the category-theoretic objects themselves with random variables. This restriction preserves the relational ontology while making the probabilistic claims mathematically explicit.

\subsection{Extending \texorpdfstring{$(F,A)$}{(F, A)}-systems}
\citet{hofmeyr2021biochemically}'s (Fabrication, Assembly)-systems refine~\citet{rosen1991life}'s (Metabolism, Repair)-systems by replacing the overly restrictive metabolism-repair vocabulary with a well-grounded biochemical distinction between fabrication and assembly. Their importance for the present paper is that they make explicit how a system can remain closed to efficient causation while remaining open to formal causation \citep[p. 12]{hofmeyr2021biochemically}. In such a system, the organization manufactures the efficient causes required for its own continuation, while \emph{free-standing} formal causes provide heritable and revisable specifications that can alter the range of admissible transformations \citep{hofmeyr2018causation}. In the jargon we previously introduced, $(F, A)$-systems are therefore not merely autocatalytic or self-maintaining; they are organizationally closed and semantically mediated, best fulfilling the criteria for standing in a full modeling relation to the causal entailments of a modern living cell \citep[p. 4]{hofmeyr2021biochemically}.

This is why $(F, A)$-systems are an appropriate starting point for a theory of agency. They already contain the minimal ingredients needed to distinguish living organizations from both dissipative structures and conventional machines: material openness, closure to efficient causation, internally grounded finality, and openness to formal causation. Yet, the original $(F, A)$-systems as a model of living organization still remains incomplete for the purposes of the present paper. Agency is not only about self-maintenance, but also about being coupled to an environment in a way that permits sensing, response, and context-sensitive modulation of its own activity \citep{fabregas2026organism}. Thus, before temporalizing the model, we first recall its ecological extension introduced in~\citet{lopez2025closing}.

As explained above, a self-manufacturing organization that lacks explicit sensing and response remains too internally focused to ground agency. For this reason, the $(F, A)$-system must be extended with processes that connect external signals, internal signals, and intracellular responses. Let $I=\{I_1,I_2,I_3,I_4\}$ be an extended set of free-standing formal causes, where $I_3$ and $I_4$ respectively encode the fabrication of sensing and response architectures. Associated with these instructions are the mappings

\[
X \xrightarrow{f_3} Y \xrightarrow{f_4} Z,
\]

where $X$ denotes external signals, $Y$ internal signals, and $Z$ intracellular responses. These processes are not simply additional metabolic reactions. They are the minimal formal representation of measurement and control at the organism--environment interface \citep{cariani1989design,  cariani2011semiotics}. In this way, we can extend Hofmeyr's $(F, A)$-systems via the following diagram. 

\begin{equation}
\label{d:hofmeyr-ext}
\begin{tikzcd}[column sep=0 cm]
& X \ar[r,thick] & Y \ar[r,thick] & Z \\
&& f = I \times \{f_1\}+f_3+f_4+f_2 \ar[lld,dashed,thick,start anchor={[xshift=-1ex]}]
\ar[rr,dashed,thick] \ar[ul,dashed,thick,start anchor={[xshift=6ex]}]
\ar[u,dashed,thick,start anchor={[xshift=6ex]}] & & D \ar[d,thick]   \\
A \ar[rr,thick] & & B_1+B_2+B_3+B_4 \ar[u,thick] & & \Phi \ar[ll,thick]
\end{tikzcd}
\end{equation}

In the extended diagram~(\ref{d:hofmeyr-ext}), $B_3$ and $B_4$ denote the components from which the sensing and response machinery are respectively assembled. The point is not that every organism literally implements this exact coarse-graining, but that any agentive organization must include some analog of this architecture: a way of being affected by the environment, a way of internally transforming that perturbation, and a way of acting back upon itself or the environment \citep{cariani1998epistemic}. In biosemiotic terms, this is the minimal interface through which material perturbations become measurable signs and through which internal states become control processes \citep{pattee1991measurement, pattee1993limitations}. In ecological terms, it opens the system to \emph{affordances}, understood not as pre-given external objects, but as relational possibilities for action enacted by an organism with a particular organization \citep{gibson1979ecological, kauffman2021world}.

The previous sections argued that semantic closure must be temporally unfolded because self-construction is processual. Formally, this means that each constitutive mapping in the extended $(F, A)$-system must be associated with a characteristic timescale. Let's begin by rewriting the extended organization~(\ref{d:hofmeyr-ext}) as

\begin{align}
    &A \xrightarrow{f} B \xrightarrow{\Phi_f} H(A,B), \nonumber \\
    &D \xrightarrow{f} H\left(B,H(A,B)\right), \nonumber \\
    &X \xrightarrow{f_3} Y \xrightarrow{f_4} Z.
    \label{eq:FA-explicit-agency}
\end{align}

Here $f=I\times\{f_1\}+f_2+f_3+f_4$, $I=\{I_1,I_2,I_3,I_4\}$, and $B=B_1+B_2+B_3+B_4$. Additionally, $H(X, Y)$ is a proper subset of the hom-set of all mappings from $X$ to $Y$. This notation compresses the self-manufacturing, interpretive, sensing, and response components into a single relational organization. To make the temporal structure more transparent, let $C=H(A, B)$ and $E=H(B,H(A, B))$, and rewrite the mappings as

\begin{align}
    &A \xrightarrow{\phi} B \xrightarrow{\psi} C, \nonumber \\
    &D \xrightarrow{\theta} E, \nonumber \\
    &X \xrightarrow{\alpha} Y \xrightarrow{\beta} Z.
    \label{eq:FA-simplified-agency}
\end{align}

Following the logic of~\citet{montevil2015biological}, we now assign a characteristic timescale $\tau_x$ to each process $x\in\{\phi,\psi,\theta,\alpha,\beta\}$:

\begin{align}
    &A \xrightarrow[\tau_{\phi}]{\phi} B \xrightarrow[\tau_{\psi}]{\psi} C, \nonumber \\
    &D \xrightarrow[\tau_{\theta}]{\theta} E, \nonumber \\
    &X \xrightarrow[\tau_{\alpha}]{\alpha} Y \xrightarrow[\tau_{\beta}]{\beta} Z.
    \label{eq:FA-temporal-agency}
\end{align}

This is the formal point at which relational organization becomes temporally articulated. The timescales should not be interpreted as external clock parameters imposed on an otherwise completed system. They express the fact that fabrication, assembly, interpretation, sensing, and response are maintained at different rates and that the organization persists only by coordinating these rates across time \citep{bich2022organization, bechtel2024situating}. Nevertheless, as well-argued by~\citet{montevil2015biological}, every biological constraint must be considered at two scales: the scale at which it constrains another process and the scale at which it is itself produced or maintained. Thus, temporal coherence in a closed organization requires more than the existence of multiple timescales. It requires that the production and operation of constraints remain mutually coordinated \citep{wilson2022scaffolding, kubiak2024timing}. 

\begin{figure*}[ht!]
\centering
\begin{adjustbox}{max width=\textwidth}
\begin{tikzcd}[column sep=5.5em, row sep=40ex, >={Latex}, thick]
A_t \arrow[r,"\phi_t"] & B_t \arrow[r,"\psi_t"] & C_t &
D_t \arrow[r,"\theta_t"] & E_t &
X_t \arrow[r,"\alpha_t"] & Y_t \arrow[r,"\beta_t"] & Z_t \\
A_{t+\tau} \arrow[r,"\phi_{t+\tau}"{name=phi_ttau}] &
B_{t+\tau} \arrow[r,"\psi_{t+\tau}"{name=psi_ttau}] & C_{t+\tau} &
D_{t+\tau} \arrow[r,"\theta_{t+\tau}"{name=theta_ttau}] & E_{t+\tau} &
X_{t+\tau} \arrow[r,"\alpha_{t+\tau}"{name=alpha_ttau}] &
Y_{t+\tau} \arrow[r,"\beta_{t+\tau}"{name=beta_ttau}] & Z_{t+\tau}
%
\arrow[from=1-3, to=phi_ttau,   bend right=18, dashed, shorten <=2pt, shorten >=2pt, "\tau_{\phi}"']
\arrow[from=1-3, to=theta_ttau, bend right=38, dashed, shorten <=2pt, shorten >=2pt, "\tau_{\theta}"]
\arrow[from=1-5, to=psi_ttau,   bend right=10, dashed, shorten <=2pt, shorten >=2pt, "\tau_{\psi}"']
\arrow[from=1-3, to=alpha_ttau, bend right=20,  dashed, shorten <=2pt, shorten >=2pt, "\tau_{\alpha}"]
\arrow[from=1-3, to=beta_ttau,  bend left=20,  dashed, shorten <=2pt, shorten >=2pt, "\tau_{\beta}"']
\end{tikzcd}
\end{adjustbox}
\caption{One admissible stochastic realization of the temporally parametrized (F, A)-systems.}
\label{fig:ADBN}
\end{figure*}
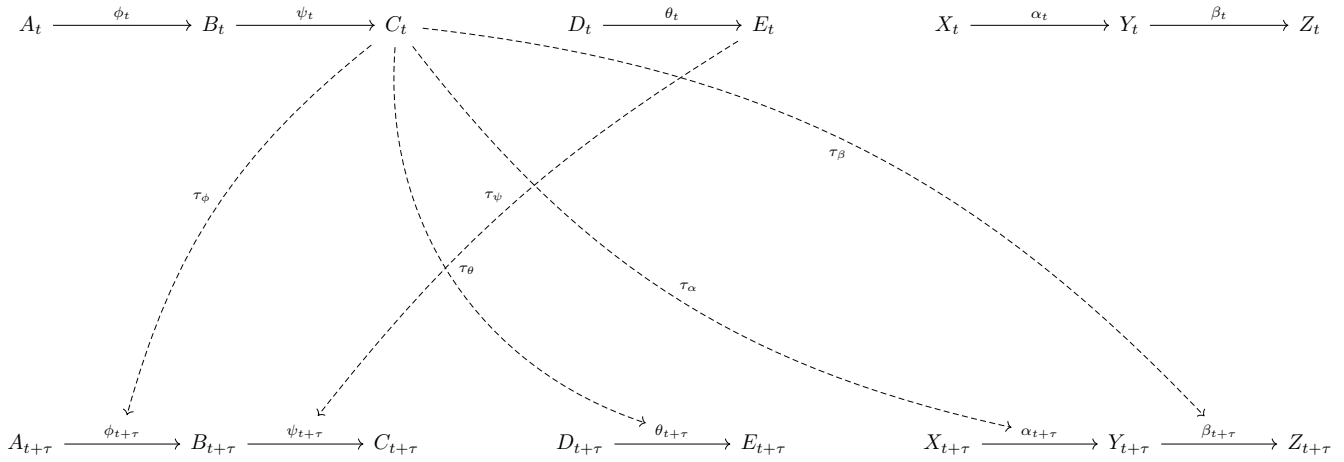

We can express this by defining $\Delta_{\phi}=\tau_{\phi}-\tau_{\psi}$, $\Delta_{\psi}=\tau_{\psi}-\tau_{\theta}$, $\Delta_{\theta}=\tau_{\theta}-\tau_{\psi}$, $\Delta_{\alpha}=\tau_{\alpha}-\tau_{\psi}$, and $\Delta_{\beta}=\tau_{\beta}-\tau_{\psi}$. Since the system is closed to efficient causation, at least one $\Delta_i$ must be positive and another $\Delta_j$, with $j\neq i$, must be negative \citep[p. 8--9]{montevil2015biological}. This condition prevents the temporalized organization from collapsing into a linear chain of externally ordered processes. It expresses the interdependence of processes that must be fast enough to act as constraints and slow enough to be maintained as constraints. Following~\citet{korbak2023self}, once the extended $(F, A)$-system~(\ref{d:hofmeyr-ext}) is temporally parametrized, it can be redescribed as an ADBN, as shown in Figure~\ref{fig:ADBN}. This is not because the organism is literally a Bayesian network, nor because Bayesian formalisms provide the metaphysical foundation of life. Rather, the ADBN is a formal redescription of the dependency structure induced by temporally unfolded closure; it allows us to represent how the status of one component-process at time $t$ constrains the possible circumstances of other component-processes at $t+\tau$, where $\tau=\tau(\tau_\phi,\tau_\psi,\tau_\theta,\tau_\alpha,\tau_\beta)$. This is the same formal move introduced in~\citet{lopez2025closing}, but here it is interpreted as the mathematical bridge between semantic closure, anticipation, and agency.

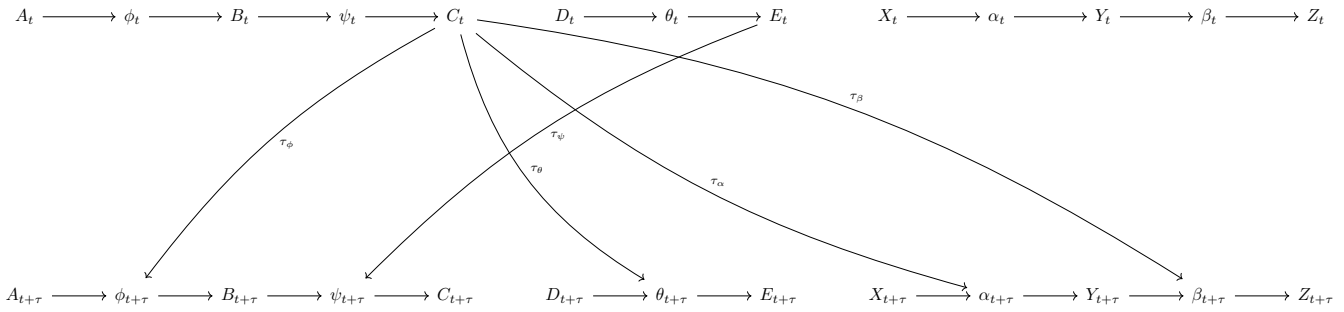
\begin{figure*}[ht!]
\centering
\begin{adjustbox}{width=\textwidth}
\begin{tikzcd}[column sep=3.2em, row sep=32ex, >={Latex}, thick]
A_t \arrow[r] & \phi_t \arrow[r] & B_t \arrow[r] & \psi_t \arrow[r] & C_t &
D_t \arrow[r] & \theta_t \arrow[r] & E_t &
X_t \arrow[r] & \alpha_t \arrow[r] & Y_t \arrow[r] & \beta_t \arrow[r] & Z_t \\
A_{t+\tau} \arrow[r] & \phi_{t+\tau} \arrow[r] & B_{t+\tau} \arrow[r] & \psi_{t+\tau} \arrow[r] & C_{t+\tau} &
D_{t+\tau} \arrow[r] & \theta_{t+\tau} \arrow[r] & E_{t+\tau} &
X_{t+\tau} \arrow[r] & \alpha_{t+\tau} \arrow[r] & Y_{t+\tau} \arrow[r] & \beta_{t+\tau} \arrow[r] & Z_{t+\tau}
%
\arrow[from=1-5, bend right=12, to=2-2,  shorten <=2pt, shorten >=2pt, "\tau_{\phi}"]
\arrow[from=1-5, bend right=22, to=2-7,  shorten <=2pt, shorten >=2pt, "\tau_{\theta}"]
\arrow[from=1-8, bend right=12, to=2-4,  shorten <=2pt, shorten >=2pt, "\tau_{\psi}"]
\arrow[from=1-5, bend right=12,  to=2-10, shorten <=2pt, shorten >=2pt, "\tau_{\alpha}"]
\arrow[from=1-5, bend left=12,  to=2-12, shorten <=2pt, shorten >=2pt, "\tau_{\beta}"]
\end{tikzcd}
\end{adjustbox}
\caption{Probabilistic Graphical Model associated with the temporal parametrization of extended (F, A)-systems.}
\label{fig:PGM}
\end{figure*}

The visual structure of the ADBN should be read carefully. In Figure~\ref{fig:ADBN}, the horizontal arrows represent the relational mappings internal to each time slice, while the cross-time arrows represent dependencies induced by the characteristic timescales of the organization. In the corresponding probabilistic graphical model (shown in Figure~\ref{fig:PGM}), both components and processes are treated as random variables, and dashed temporal dependencies are redescribed as probabilistic dependencies. In this sense, the ADBN does not replace the relational model as such, but it translates the relational model into a form in which temporal dependency, uncertainty, and context-sensitivity can be explicitly represented.

Furthermore, the ADBN permits the factorization of a joint distribution over the temporally indexed components and processes of the organization \citep[p. 45]{korbak2023self}. Using the notation adopted in Figures~\ref{fig:ADBN} and~\ref{fig:PGM}, the distribution takes the form

\begin{equation}
\resizebox{\columnwidth}{!}{$
\begin{aligned}
p(&\phi_{\tau}, \psi_{\tau}, \theta_{\tau}, \alpha_{\tau}, \beta_{\tau},
    A_{\tau}, B_{\tau}, C_{\tau}, D_{\tau}, E_{\tau}, X_{\tau}, Y_{\tau}, Z_{\tau}, \ldots, \\
  & \phi_{T}, \psi_{T}, \theta_{T}, \alpha_{T}, \beta_{T}, A_{T}, B_{T}, C_{T}, D_{T}, E_{T}, X_{T}, Y_{T}, Z_{T}) =
\\
&\prod_{t=0}^{T-\tau} \Big[
    p(A_{t+\tau})
    \, p(\phi_{t+\tau} \mid A_{t+\tau}, C_t)
    \, p(B_{t+\tau} \mid \phi_{t+\tau})
    \, p(\psi_{t+\tau} \mid B_{t+\tau}, E_t)
\\
&\qquad\quad
    p(C_{t+\tau} \mid \psi_{t+\tau})
    \, p(D_{t+\tau})
    \, p(\theta_{t+\tau} \mid D_{t+\tau}, C_t)
    \, p(E_{t+\tau} \mid \theta_{t+\tau})
    \, p(X_{t+\tau})
\\
&\qquad\quad
    p(\alpha_{t+\tau} \mid X_{t+\tau}, C_t)
    \, p(Y_{t+\tau} \mid \alpha_{t+\tau})
    \, p(\beta_{t+\tau} \mid Y_{t+\tau}, C_t)
    \, p(Z_{t+\tau} \mid \beta_{t+\tau}) \Big].
\end{aligned}
$}
\label{eq:joint-distribution-temporal-FA}
\end{equation}

Here \(T<\infty\) is the endpoint of the observation window, \(t\) indexes admissible update times, and \(\tau>0\) is the displayed lag between causally related time slices. When process-specific lags are suppressed for readability, \(\tau\) denotes the common display lag and the distinct lags remain those assigned in Figure~\ref{fig:PGM}. The ellipsis includes all intermediate time slices up to \(T\). Each factor \(p(\,\cdot\mid\cdot\,)\) denotes a conditional probability mass function or density---or, more generally, a Markov kernel on the measurable spaces chosen for the stochastic realization---and the factorization is completed by the initial distribution of any parent variables lying at or before the beginning of the window.

More generally, let $\mathcal V$ be the set of time-indexed random variables selected for a stochastic realization, and let $G_{0:T}$ be the resulting directed acyclic dependency graph over the finite observation window $0{:}T$. For each variable $V\in\mathcal V$ and time $t$, let $K_{V,t}$ be a Markov kernel from the values of its temporally lagged parents $\operatorname{Pa}_{G}(V,t)$ to the state space of $V_t$. The path distribution factorizes as
\begin{equation}
P(d\mathbf v_{0:T}\mid G_{0:T})
=
\prod_{t=0}^{T}
\prod_{V\in\mathcal V}
K_{V,t}\!\left(dv_t\mid \mathbf v_{\operatorname{Pa}_{G}(V,t)}\right).
\label{eq:kernel-factorization}
\end{equation}
Here \(P(\,\cdot\mid G_{0:T})\) is the probability measure on complete paths conditional on the selected dependency graph; \(\mathbf v_{0:T}\) is one complete trajectory of all variables over the observation window; \(d\mathbf v_{0:T}\) is the corresponding product-measure element; and \(\mathbf v_{\operatorname{Pa}_{G}(V,t)}\) is the realized configuration of the lagged parents of \(V_t\). The outer product ranges over observation times and the inner product over variables. Each normalized kernel \(K_{V,t}\) gives the conditional law of \(V_t\) given those parent values.
For the extended $(F,A)$-organization, the parent sets are selected so that material inputs, constitutive constraints, and environmental measurements appear with the lags indicated by the temporalized relational diagram. Equation~\eqref{eq:kernel-factorization} is deliberately generic: it separates the organizational claim---which dependencies must be represented---from the empirical choice of state spaces, kernels, and coarse-grainings used in a particular application. 

For the purposes of this manuscript, factorization \eqref{eq:joint-distribution-temporal-FA} is useful enough because it makes explicit how internal organization, environmental perturbation, and processual history jointly constrain future states. The terms involving $X$, $Y$, and $Z$ represent the organism--environment interface, while the terms involving $A$, $B$, $C$, $D$, and $E$ represent the self-manufacturing organization that gives such environmental coupling biological significance. The result is not a fixed transition function, but a temporally extended dependency structure. The distribution should therefore be interpreted in a restricted way. \emph{It is not} a complete law of the organism, and \emph{it is not} a universal principle from which life or agency can be deduced. \emph{It is} a formal redescription of a specific kind of organization once that organization has already been biologically identified as semantically closed. This restriction is crucial because it prevents the Bayesian formalism from becoming ontologically inflated \citep{nave2025drive}. In the present framework, relational closure comes first; probabilistic redescription comes second. The ADBN is a way of representing how closure unfolds through time, not a replacement for closure. 

The most important difference between the construction shown above and a conventional state-space model is that the derived probability distribution shown in equation~(\ref{eq:joint-distribution-temporal-FA}) should not be understood as fixed once and for all. A living organization does not merely update parameters inside a pre-given space of possibilities \citep{witherington2014self}. In contrast, through adaptation, development, and material turnover, a living organization may alter which variables matter, which couplings are relevant, which measurements are possible, and which actions are available \citep{pattee1972nature, longo2012no}. These changes must be distinguished rather than ranked: an ADBN may update probabilities over a fixed graph, reorganize graph topology, or reconstruct the repertoire of variables, substrates, interpretive relations, and control processes. 

This last point is essential for the theory of agency developed in this paper. If the induced ADBN were merely a fixed graphical model with parameter updates, it would not capture the kind of plasticity required for open-ended biological organization \citep{kauffman2019world, nave2025drive}. The stronger claim is that semantically closed systems can rebuild the very dependency structures through which they anticipate and act. The ADBN is therefore not a static internal model but a historically maintained and revisable organization. It is continuously regenerated by the same material and semiotic processes that it helps coordinate. This also clarifies the relation between the present framework and active inference. Our claim is not that free-energy minimization is a first principle of life, nor that a generative model is simply given in advance \citep{di2022laying}. Rather, a primitive generative model can be derived from the temporal unfolding of an already closed organization. Thus, in contrast to~\citet{korbak2021computational}, we propose that active-inference language must be used locally and carefully: organisms can be redescribed as probabilistically tailoring their internal and environmental couplings only because their relational organization already supplies the measurement and control architecture that makes such tailoring biologically meaningful. We will expand upon this argument and discussion later in this paper.

A temporalized $(F, A)$-system therefore gives a constructive interpretation of Rosennean anticipation. An anticipatory system is not merely a system that reacts quickly to perturbations, nor one that contains an externally inserted predictive module. It is a system whose present organization constrains possible future states through an internal model of its own conditions of persistence \citep{rosen1978anticipatory, louie2013more}. In the present framework, that internal model is not a fixed differential-equation surrogate, as originally suggested by~\citet{rosen2011anticipatory}; it is the revisable dependency structure generated by the temporal unfolding of semantic closure \citep[p. 9]{lopez2025closing}. This is why our ADBN approach matters for naturalizing agency. The organism's present state does not simply cause the next state according to a fixed transition rule. Rather, the present organization weights, canalizes, and constrains possible futures in light of its own continued viability \citep{juarrero1999dynamics}. Environmental signals become relevant because they enter a closed organization that can measure them, interpret them, and act upon them. 

Importantly, agency does not yet follow automatically from this construction, but the construction supplies the minimal formal space in which agency can be defined: the space between mere temporal dependency and anticipatory modulation of organism--environment coupling. The next section uses this formal apparatus to derive weaker temporalized organizations. The point will not be to decide, by stipulation, which systems are agents and which are not. Rather, by deleting specified relations from temporally parametrized $(F,A)$-systems, we can ask which organizational capacities depend on which relations and which resulting diagrams remain incomparable. The ADBN construction thereby supports a structural partial order of organizational conditions, not a scalar order from ``less'' to ``more'' agency.

\section{Which Organizational Conditions Support Agency?}
The previous section showed that temporally parametrized $(F,A)$-systems induce history-dependent probabilistic structures capable of redescribing the temporal organization of semantic closure. We now ask what happens when specified organizational relations are removed. The strategy follows the decomposition used in~\citet{lopez2025closing}, but the present question is different: the same diagrams are temporally unfolded to identify which proto-agential capacities depend on which retained relations. Mathematically, relation deletion generates a partial order from the extended diagram to weaker diagrams. Biologically, that order supplies hypotheses about organizational dependence, not a chronology. It does not assert that lineages historically acquired sensing--response, free-standing formal causes, informed efficient causes, and closure in one direction or sequence. The aim is to identify which dependency structures are present, absent, or incomparable in each regime, and therefore which agency claims each regime can or cannot support.

Importantly, the diagrams derived below form a partial order under the removal or addition of specified organizational relations. They should not be read as a universal chronological sequence, a scalar ranking of organisms, or a claim that evolution necessarily produces greater complexity, intelligence, or agency. Evolution can simplify, lose, redistribute, or externalize functions, and transitions can be reversible or bypassed \citep{levins1985dialectical, lynch2025complexity}. Here the term \emph{hierarchy} is retained only in Pattee's functional-control sense: relatively slow constraints can regulate faster dynamics across organizational levels \citep{pattee1972nature, pattee1991measurement}. Within living systems, these control relations are frequently heterarchical, reciprocal, and distributed rather than governed by a single top level \citep{bechtel2024situating}. Accordingly, the decomposition proposed below identifies partially ordered \emph{conditions of organization}, not stages of evolutionary progress. More parts or levels do not automatically imply more agency.

Furthermore, the derivation of weaker relational models already showed that the relations constituting semantic closure can dissociate \citep{lopez2025closing}. Some chemical organizations possess structured dependency without closure, some preserve memory-like traces without autonomy, and some realize closure without explicit environmental coupling. Temporalizing these diagrams allows us to ask a sharper question: not only what organization a system has, but what kind of future-oriented dependency structure that organization can support. This distinction matters because agency is not the same as self-maintenance \citep{rosslenbroich2024agency}. A system may persist, reproduce, or participate in autocatalytic amplification without possessing an endogenous anticipatory structure capable of modulating its coupling with the environment \citep{deacon2013teleology}. The ADBNs and associated joint distributions therefore locate distinct formal precursors of agency (memory, closure, constraint maintenance, environmental sensitivity, and revisable measurement--control, etc.) without asserting that they must arise in one sequence.

\subsection{Removing explicit sensing and response}

The first weakening removes the explicit sensing-response architecture from the extended $(F, A)$-system. In the full model represented by diagram~(\ref{d:hofmeyr-ext}), the mappings \(X \xrightarrow{f_3} Y \xrightarrow{f_4} Z\) depict the minimal interface through which environmental perturbations become internally measurable and internally generated states become control processes. Removing these mappings produces diagram~(\ref{d:hofmeyr_full}), which retains closure to efficient causation and free-standing formal causes, but no longer contains an explicit organism--environment measurement-control channel.

\begin{equation}
\label{d:hofmeyr_full}
\begin{tikzcd}
  & I^*\times\{f_1\}+f_2 \arrow[dl, dashed, thick] \arrow[r, dashed, thick]  & D \arrow[d, thick] \\
A \arrow[r, thick] & \arrow[u, thick] B_1+B_2 & \arrow[l, dashed, thick] \Phi_f 
\end{tikzcd}
\end{equation}

This weakening is important because it separates semantic closure from situated agency. Diagram~(\ref{d:hofmeyr_full}) can still be organizationally autonomous in the sense that the relevant efficient causes are internally produced and maintained. It can also remain open to formal causation, because free-standing formal causes still specify admissible transformations. Yet, without explicit sensing and response, the system cannot yet differentially modulate its coupling to the environment in the sense required for agency \citep{Robbins2012Primer}. It is therefore better interpreted as a semantically closed autonomous organization rather than as a fully agentive organization. The associated distribution~(\ref{eq:distro-FA-1}) preserves the main self-manufacturing dependency structure of the richer model while removing the environmental terms \(X,Y,Z,\alpha,\beta\). This loss is conceptually decisive. The distribution still redescribes a temporally coherent organization, but it no longer includes the variables through which environmental signals are converted into internal regulatory states and responses. Thus, what has been lost is not closure, but situated measurement-control.

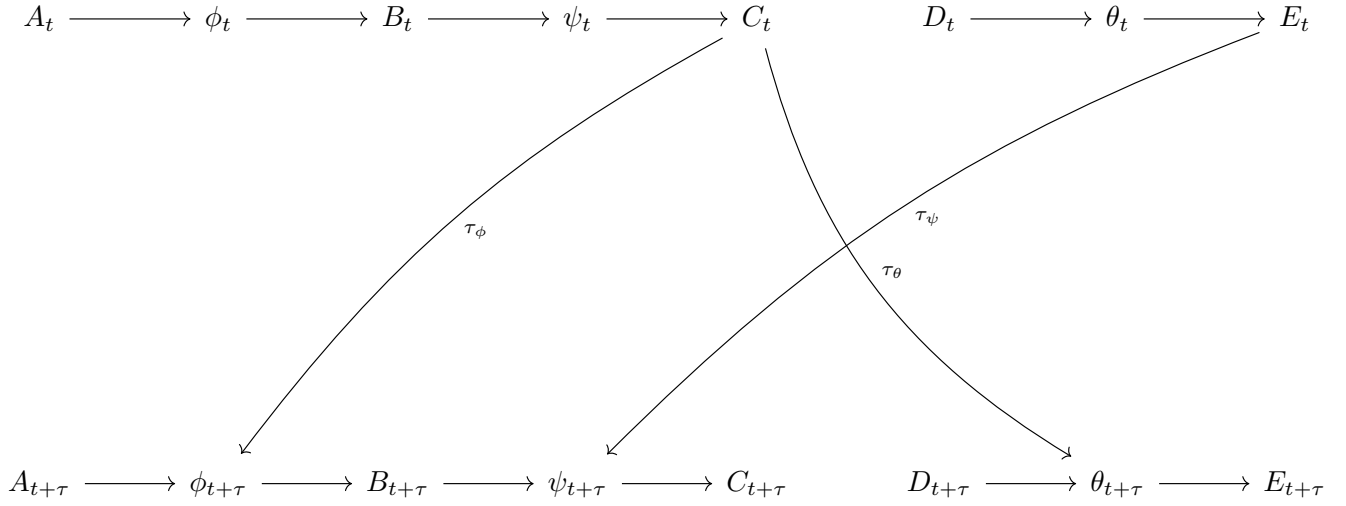
\begin{figure*}[ht!]
\centering
\begin{adjustbox}{width=\textwidth}
\begin{tikzcd}[column sep=3.2em, row sep=32ex, >={Latex}, thick]
A_t \arrow[r] & \phi_t \arrow[r] & B_t \arrow[r] & \psi_t \arrow[r] & C_t &
D_t \arrow[r] & \theta_t \arrow[r] & E_t  \\
A_{t+\tau} \arrow[r] & \phi_{t+\tau} \arrow[r] & B_{t+\tau} \arrow[r] & \psi_{t+\tau} \arrow[r] & C_{t+\tau} &
D_{t+\tau} \arrow[r] & \theta_{t+\tau} \arrow[r] & E_{t+\tau} 
%
\arrow[from=1-5, bend right=12, to=2-2,  shorten <=2pt, shorten >=2pt, "\tau_{\phi}"]
\arrow[from=1-5, bend right=22, to=2-7,  shorten <=2pt, shorten >=2pt, "\tau_{\theta}"]
\arrow[from=1-8, bend right=12, to=2-4,  shorten <=2pt, shorten >=2pt, "\tau_{\psi}"]
\end{tikzcd}
\end{adjustbox}
\caption{Probabilistic Graphical Model associated with the temporal parametrization of diagram~(\ref{d:hofmeyr_full}).}
\label{fig:PGM-1}
\end{figure*}

\begin{equation}
\resizebox{\columnwidth}{!}{$
\begin{aligned}
p(&\phi_{\tau}, \psi_{\tau}, \theta_{\tau},
    A_{\tau}, B_{\tau}, C_{\tau}, D_{\tau}, E_{\tau}, \ldots, \phi_{T}, \psi_{T}, \theta_{T}, A_{T}, B_{T}, C_{T}, D_{T}, E_{T}) =
\\
&\prod_{t \ge 0}^{T-\tau} \Big[
    p(A_{t+\tau})
    \, p(\phi_{t+\tau} \mid A_{t+\tau}, C_t)
    \, p(B_{t+\tau} \mid \phi_{t+\tau})
    \, p(\psi_{t+\tau} \mid B_{t+\tau}, E_t)
\\
&\qquad\quad
    p(C_{t+\tau} \mid \psi_{t+\tau})
    \, p(D_{t+\tau})
    \, p(\theta_{t+\tau} \mid D_{t+\tau}, C_t)
    \, p(E_{t+\tau} \mid \theta_{t+\tau}) \Big].
\end{aligned}
$}
\label{eq:distro-FA-1}
\end{equation}

\subsection{From free-standing formal causes to intrinsic material memory}

Following~\citet{lopez2025closing}, the next diagrammatic weakening removes the free-standing formal causes \(I^*=\{I_1,I_2\}\). In diagram~(\ref{d:hofmeyr_succesor}), the material causes \(A_1^*+A_2^*\) become informed material causes: substrates whose relevant syntax is materially inscribed rather than separately specified by independent symbolic structures \citep{deacon2014transition}. This corresponds to a plausible pre-digital regime in which organization is controlled by intrinsic physico-chemical regularities rather than by free-standing instructions \citep{pattee2012does}; it is not claimed to be a mandatory historical stage.

\begin{equation}
\label{d:hofmeyr_succesor}
\begin{tikzcd}
& f_1+f_2 \arrow[dl, color=black, dashed, thick] \arrow[r, color=black, dashed, thick]  & D \arrow[d, color=black, thick] \\
A_1^* + A_2^* \arrow[r, color=black, thick] & \arrow[u, color=black, thick] B_1+ B_2 & \arrow[l, color=black, dashed, thick] \Phi_f   
\end{tikzcd}
\end{equation}

This transition marks an important reduction in evolvability. Free-standing formal causes allow instructions to be recombined, inherited, and revised with a degree of independence from the material processes they constrain \citep{hofmeyr2018causation}. Informed material causes can still bias transformations, but their formal role remains embedded in their own material structure \citep{pattee2021symbol}. The system therefore retains a kind of proto-semantic organization, but its capacity for syntactic openness is much more limited. The distribution~(\ref{eq:distro-FA-2}) makes this limitation visible. Formally, it remains close to~(\ref{eq:distro-FA-1}), because the temporal dependencies among \(\phi\), \(\psi\), and \(\theta\) are still present. Biologically, however, the interpretation changes: the formal cause is no longer a separable instruction-like component but an intrinsic constraint of the material substrate. This is a weaker anticipation structure, because the system can still bias future states, but it cannot yet freely rewrite the symbolic basis through which those futures are specified \citep{egbert2023behaviour}.

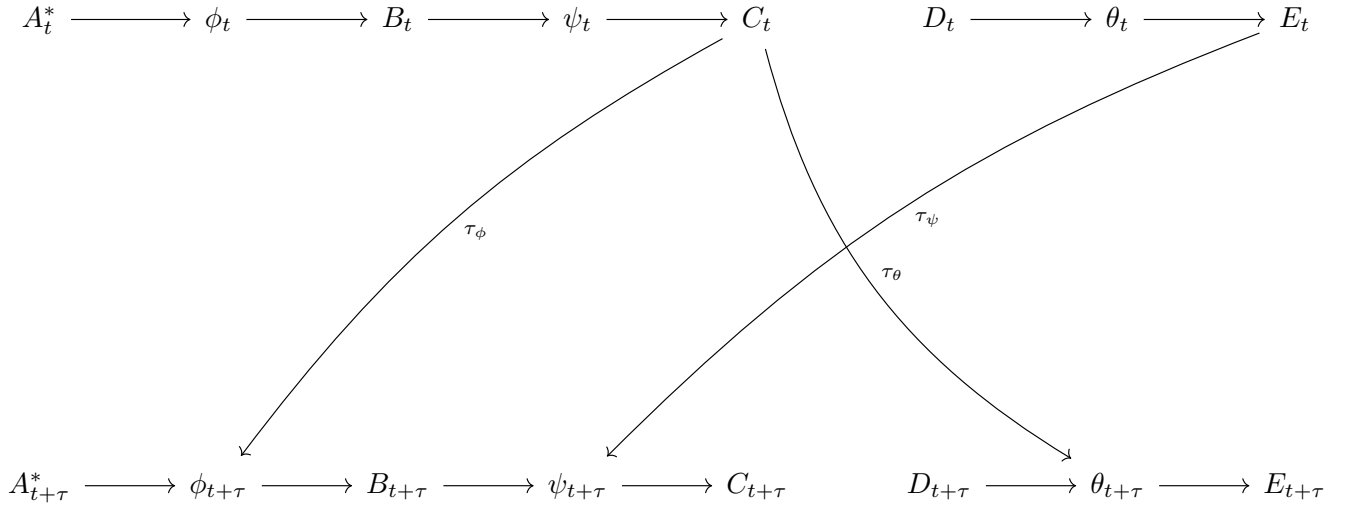
\begin{figure*}[ht!]
\centering
\begin{adjustbox}{width=\textwidth}
\begin{tikzcd}[column sep=3.2em, row sep=32ex, >={Latex}, thick]
A^{*}_t \arrow[r] & \phi_t \arrow[r] & B_t \arrow[r] & \psi_t \arrow[r] & C_t &
D_t \arrow[r] & \theta_t \arrow[r] & E_t  \\
A^{*}_{t+\tau} \arrow[r] & \phi_{t+\tau} \arrow[r] & B_{t+\tau} \arrow[r] & \psi_{t+\tau} \arrow[r] & C_{t+\tau} &
D_{t+\tau} \arrow[r] & \theta_{t+\tau} \arrow[r] & E_{t+\tau} 
%
\arrow[from=1-5, bend right=12, to=2-2,  shorten <=2pt, shorten >=2pt, "\tau_{\phi}"]
\arrow[from=1-5, bend right=22, to=2-7,  shorten <=2pt, shorten >=2pt, "\tau_{\theta}"]
\arrow[from=1-8, bend right=12, to=2-4,  shorten <=2pt, shorten >=2pt, "\tau_{\psi}"]
\end{tikzcd}
\end{adjustbox}
\caption{Probabilistic Graphical Model associated with the temporal parametrization of diagram~(\ref{d:hofmeyr_succesor}).}
\label{fig:PGM-2}
\end{figure*}

\begin{equation}
\resizebox{\columnwidth}{!}{$
\begin{aligned}
p(&\phi_{\tau}, \psi_{\tau}, \theta_{\tau},
    A^{*}_{\tau}, B_{\tau}, C_{\tau}, D_{\tau}, E_{\tau}, \ldots, \phi_{T}, \psi_{T}, \theta_{T}, A^{*}_{T}, B_{T}, C_{T}, D_{T}, E_{T}) =
\\
&\prod_{t \ge 0}^{T-\tau} \Big[
    p(A^{*}_{t+\tau})
    \, p(\phi_{t+\tau} \mid A^{*}_{t+\tau}, C_t)
    \, p(B_{t+\tau} \mid \phi_{t+\tau})
    \, p(\psi_{t+\tau} \mid B_{t+\tau}, E_t)
\\
&\qquad\quad
    p(C_{t+\tau} \mid \psi_{t+\tau})
    \, p(D_{t+\tau})
    \, p(\theta_{t+\tau} \mid D_{t+\tau}, C_t)
    \, p(E_{t+\tau} \mid \theta_{t+\tau}) \Big].
\end{aligned}
$}
\label{eq:distro-FA-2}
\end{equation}

\subsection{Toward autogenic closure}
The next weakening produces diagram~(\ref{d:hofmeyr_ss}), where \(D\) becomes an informed material cause \(D^*\), and \(f_2\) is replaced by a non-informed efficient cause \(f_2^*\). This regime remains especially important because it retains a form of reciprocal closure while reducing the machinery needed to support it. It therefore approximates the kind of pre-cellular organization described by~\citet{katla2025self}, in which self-maintenance is not yet genetically or digitally controlled, but still depends on a mutually reinforcing relation between transformation and constraint.

\begin{equation}
\label{d:hofmeyr_ss}
\begin{tikzcd}
& f_1+f_2^* \arrow[dl, color=black, dashed, thick] \arrow[r, color=black, dashed, thick]  & D^* \arrow[d, color=black, thick] \\
A_1^* + A_2^* \arrow[r, color=black, thick] & \arrow[u, color=black, thick] B_1+ B_2 & \arrow[l, color=black, dashed, thick] \Phi_f   
\end{tikzcd}
\end{equation}

In this sense, diagram~(\ref{d:hofmeyr_ss}) is one of the central thresholds in the section. It is not yet an agent, because it lacks explicit sensing-response and robust formal openness. Yet it is no longer a mere open reaction chain. It contains the minimal circularity needed for a system to maintain some of the conditions of its own continuation. In the language of agency, this is where proto-agency begins to become more than reactive chemistry: the system starts to preserve a dependency structure that makes its own future states non-arbitrary \citep{moreno2018minimal}. The associated distribution~(\ref{eq:distro-FA-3}) preserves the same general temporal form as the previous two distributions, but the starred variables indicate a change in causal interpretation. What was previously handled by free-standing or informed efficient causes is now carried by materially inscribed constraints. Thus, the system still has temporal dependency and primitive internal constraint, but the anticipatory organization is less plastic. This helps distinguish primitive autonomy from the richer form of agency defended in this paper.

\begin{equation}
\resizebox{\columnwidth}{!}{$
\begin{aligned}
p(&\phi_{\tau}, \psi_{\tau}, \theta_{\tau},
    A^{*}_{\tau}, B_{\tau}, C_{\tau}, D^{*}_{\tau}, E_{\tau}, \ldots, \phi_{T}, \psi_{T}, \theta_{T}, A^{*}_{T}, B_{T}, C_{T}, D^{*}_{T}, E_{T}) =
\\
&\prod_{t \ge 0}^{T-\tau} \Big[
    p(A^{*}_{t+\tau})
    \, p(\phi_{t+\tau} \mid A^{*}_{t+\tau}, C_t)
    \, p(B_{t+\tau} \mid \phi_{t+\tau})
    \, p(\psi_{t+\tau} \mid B_{t+\tau}, E_t)
\\
&\qquad\quad
    p(C_{t+\tau} \mid \psi_{t+\tau})
    \, p(D^{*}_{t+\tau})
    \, p(\theta_{t+\tau} \mid D^{*}_{t+\tau}, C_t)
    \, p(E_{t+\tau} \mid \theta_{t+\tau}) \Big].
\end{aligned}
$}
\label{eq:distro-FA-3}
\end{equation}

\begin{figure*}[ht!]
\centering
\begin{adjustbox}{width=\textwidth}
\begin{tikzcd}[column sep=3.2em, row sep=32ex, >={Latex}, thick]
A^{*}_t \arrow[r] & \phi_t \arrow[r] & B_t \arrow[r] & \psi_t \arrow[r] & C_t &
D^{*}_t \arrow[r] & \theta_t \arrow[r] & E_t  \\
A^{*}_{t+\tau} \arrow[r] & \phi_{t+\tau} \arrow[r] & B_{t+\tau} \arrow[r] & \psi_{t+\tau} \arrow[r] & C_{t+\tau} &
D^{*}_{t+\tau} \arrow[r] & \theta_{t+\tau} \arrow[r] & E_{t+\tau} 
%
\arrow[from=1-5, bend right=12, to=2-2,  shorten <=2pt, shorten >=2pt, "\tau_{\phi}"]
\arrow[from=1-5, bend right=22, to=2-7,  shorten <=2pt, shorten >=2pt, "\tau_{\theta}"]
\arrow[from=1-8, bend right=12, to=2-4,  shorten <=2pt, shorten >=2pt, "\tau_{\psi}"]
\end{tikzcd}
\end{adjustbox}
\caption{Probabilistic Graphical Model associated with the temporal parametrization of diagram~(\ref{d:hofmeyr_ss}).}
\label{fig:PGM-3}
\end{figure*}
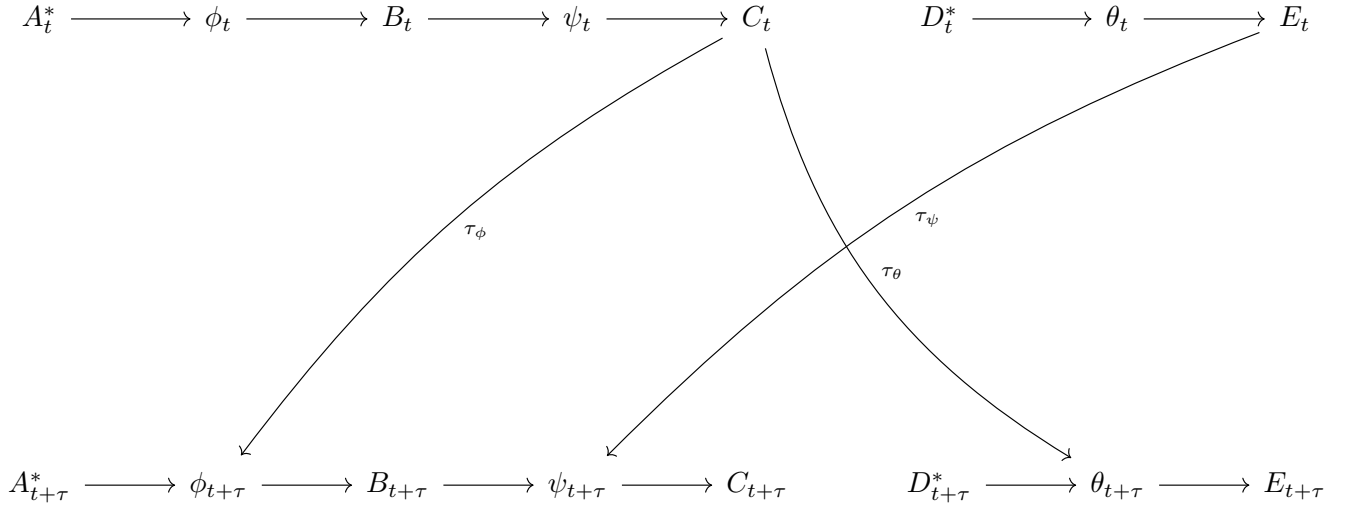

\subsection{Splitting the diagram: memory without closure and closure without agency}
Diagram~(\ref{d:hofmeyr_ss}) can be decomposed into two components. The first component, diagram~(\ref{d:component1}), has the form of a nested dependency that is not closed to efficient causation. It can preserve a kind of memory, because previous products constrain later transformations, but it does not internally produce the whole network of efficient causes required for autonomous self-maintenance \citep{hofmeyr2021biochemically}.

\begin{equation}
\label{d:component1}
\begin{tikzcd}
& f_1 \arrow[dl, color=black, dashed, thick] &  \\
A_1^* \arrow[r, color=black, thick] & \arrow[u, color=black, thick] B_1 & \arrow[l, color=black, dashed, thick] \Phi_{f_1}
\end{tikzcd}
\end{equation}

The distribution~(\ref{eq:distro-FA-3-1}) should therefore be interpreted as a proto-inferential structure rather than as an autonomous agentive structure. The term \(C_t\) constrains the later production of \(\phi_{t+\tau}\), which means the system is no longer a memoryless reaction \citep{duenas2019chemistry}. Yet the dependency does not close back on itself in the strong organizational sense. This is the kind of system that may already exhibit structured sensitivity to prior states, but whose organization is still externally dependent.

\begin{equation}
\resizebox{\columnwidth}{!}{$
\begin{aligned}
p(&\phi_{\tau}, \psi_{\tau}, A^{*}_{\tau}, B_{\tau}, C_{\tau}, \ldots, \phi_{T}, \psi_{T}, A^{*}_{T}, B_{T}, C_{T}) =
\\
&\prod_{t \ge 0}^{T-\tau} \Big[
    p(A^{*}_{t+\tau})
    \, p(\phi_{t+\tau} \mid A^{*}_{t+\tau}, C_t)
    \, p(B_{t+\tau} \mid \phi_{t+\tau})
    \, p(\psi_{t+\tau} \mid B_{t+\tau}) p(C_{t+\tau} \mid \psi_{t+\tau}) \Big].
\end{aligned}
$}
\label{eq:distro-FA-3-1}
\end{equation}

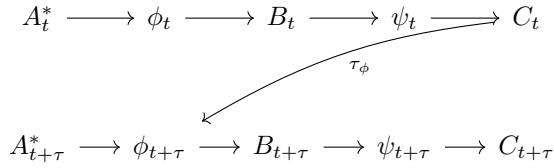
\begin{figure}[ht!]
\centering
\begin{adjustbox}{max width=\columnwidth, center}
\begin{tikzcd}[column sep=1.6em, row sep=7ex, >={Latex}, thick]
A^{*}_t \arrow[r] & \phi_t \arrow[r] & B_t \arrow[r] & \psi_t \arrow[r] & C_t \\
A^{*}_{t+\tau} \arrow[r] & \phi_{t+\tau} \arrow[r] & B_{t+\tau} \arrow[r] & \psi_{t+\tau} \arrow[r] & C_{t+\tau} 
%
\arrow[from=1-5, bend right=12, to=2-2,  shorten <=2pt, shorten >=2pt, "\tau_{\phi}"]
\end{tikzcd}
\end{adjustbox}
\caption{Probabilistic Graphical Model associated with the temporal parametrization of diagram~(\ref{d:component1}).}
\label{fig:PGM-3-1}
\end{figure}

The second component, diagram~(\ref{d:component2}), is more interesting for agency. Unlike the first component, it retains a minimal closure relation. It resembles the autogenic logic discussed by~\citet{deacon2011incomplete}, where reciprocal catalytic and self-assembling processes begin to generate a system that is more than merely autocatalytic. Autocatalysis can amplify a reaction network, but by itself it is only self-promoting, not yet self-regulating or self-preserving. Closure becomes decisive when the system begins to maintain the constraints that prevent its own dissipation.

The mapping to Deacon's autogen should not be taken to settle the origin of semiosis. \citet{deacon2011incomplete} interprets autogenic organization as exhibiting minimal interpretive competence \citep{deacon2021molecules, deacon2023minimal}, whereas Pattee argues that symbol grounding through a physically realized symbol--action relation precedes interpretation proper \citep{pattee2021symbol}. The accompanying commentaries likewise disagree about which energetic, informational, and triadic conditions are sufficient \citep{favareau2021facing, raczaszek2021complementarity, joslyn2022semiotic}. The present framework adopts the more conservative formulation. Autogenic closure is a plausible \emph{presemiotic or protosemiotic precursor}: it becomes semantically closed only when relatively rate-independent vehicles participate in constructing and maintaining the measurement--control constraints that give those vehicles functional significance. Structural correspondence with a closed reaction cycle is therefore necessary evidence, not sufficient evidence, for full biological semiosis.

\begin{equation}
\label{d:component2}
\begin{tikzcd}
f_2^* \arrow[r, color=black, dashed, thick]  & D^* \arrow[d, color=black, thick] \\
\arrow[u, color=black, thick] B_2 & \arrow[l, color=black, dashed, thick] \Phi_{f_2}   
\end{tikzcd}
\end{equation}

The revised distribution~(\ref{eq:distro-FA-3-2}) expresses a mutual enabling relation without duplicating conditional factors. The previous state of one process constrains the current production of the other, and vice versa. This is not yet full agency, but it is a minimal temporal form of self-maintaining anticipation: the system's present organization is conditioned by traces of its own prior organization. At this level, proto-agency becomes a real candidate rather than a metaphor, because the system begins to act as a locus of constraint maintenance \citep{deacon2021molecules}.

\small
\begin{equation}
\begin{aligned}
p(&\phi_{\tau}, \psi_{\tau}, B_{\tau}, D^{*}_{\tau}, \ldots,
    \phi_{T}, \psi_{T}, B_{T}, D^{*}_{T}) =
\\
&\prod_{t\ge 0}^{T-\tau} \Big[
    p(B_{t+\tau})
    \, p(\phi_{t+\tau} \mid B_{t+\tau}, \psi_t)
    \, p(D^{*}_{t+\tau} \mid \phi_{t+\tau})
    \, p(\psi_{t+\tau} \mid D^{*}_{t+\tau}, \phi_t)
\Big].
\end{aligned}
\label{eq:distro-FA-3-2}
\end{equation}

\begin{figure}[ht!]
\centering
\begin{adjustbox}{max width=\columnwidth, center}
\begin{tikzcd}[column sep=1.6em, row sep=7ex, >={Latex}, thick]
B_t \arrow[r] & \phi_t \arrow[r] & D^{*}_t \arrow[r] & \psi_t \\
B_{t+\tau} \arrow[r] & \phi_{t+\tau} \arrow[r] & D^{*}_{t+\tau} \arrow[r] & \psi_{t+\tau}
\arrow[from=1-4, bend left=14, to=2-2, shorten <=2pt, shorten >=2pt, "\tau_{\phi}"]
\arrow[from=1-2, bend right=14, to=2-4, shorten <=2pt, shorten >=2pt, "\tau_{\psi}"]
\end{tikzcd}
\end{adjustbox}
\caption{Probabilistic graphical model associated with the temporal parametrization of diagram~(\ref{d:component2})}
\label{fig:PGM-3-2}
\end{figure}
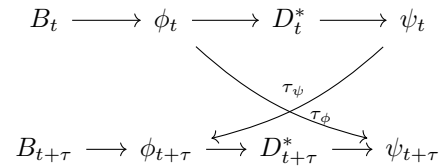

\subsection{Opening the loop: selection before closure}
According to~\citet{lopez2025closing}, a further diagrammatic weakening opens the closed component into diagrams~(\ref{d:no_closed}) and~(\ref{d:no_closed_2}). These diagrams retain structured dependency but lose closure to efficient causation. They matter evolutionarily because selection need not wait for full agency or autonomy \citep{kauffman1992origins}. Patterns can become differentially stabilized without being agents, provided there is variation, differential persistence, and some channel through which the pattern is reconstructed \citep{doolittle2018processes}. This regime connects the present formalism to broader evolutionary accounts of constraint maintenance \citep{poledna2026towards}. The claim is limited: selection can shape agency-relevant precursor relations without implying a universal progression toward agency. What is selected in this case is a dependency pattern capable of persisting across perturbation cycles, not yet an autonomous self.

\begin{equation}
\label{d:no_closed}
\begin{tikzcd}
& \Phi_{f_2} \arrow[dl, color=black, dashed, thick]  \\
B_2 \arrow[r, color=black, thick] & f_2^* \arrow[dl, color=black, dashed, thick]\\   
D^* \arrow[r, color=black, thick] & F
\end{tikzcd}
\end{equation}

\begin{equation}
\label{d:no_closed_2}
\begin{tikzcd}
& f_2^* \arrow[dl, color=black, dashed, thick]  \\
D^* \arrow[r, color=black, thick] & \Phi_{f_2} \arrow[dl, color=black, dashed, thick]\\   
B_2 \arrow[r, color=black, thick] & F
\end{tikzcd}
\end{equation}

The distribution~(\ref{eq:distro-FA-3-1-1}) captures this open dependency structure. A previous product \(C_t\) constrains the later process \(\phi_{t+\tau}\), but the system lacks the reciprocal closure required for autonomy. It is therefore misleading to call this organization agentive. It is better understood as a proto-agential precursor: a chemically realizable dependency structure in which history begins to matter, but not yet as the self-produced history of an autonomous system \citep{duenas2019chemistry}.

\small
\begin{equation}
\begin{aligned}
p(&\phi_{\tau}, \psi_{\tau}, B_{\tau}, C_{\tau}, D^{*}_{\tau}, F_{\tau}, \ldots,
    \phi_{T}, \psi_{T}, B_{T}, C_{T}, D^{*}_{T}, F_{T}) =
\\
&\prod_{t=0}^{T-\tau} \Big[
    p(B_{t+\tau})
    \, p(\psi_{t+\tau} \mid B_{t+\tau})
    \, p(C_{t+\tau} \mid \psi_{t+\tau})
\\
&\qquad\quad
    p(D^{*}_{t+\tau})
    \, p(\phi_{t+\tau} \mid D^{*}_{t+\tau}, C_t)
    \, p(F_{t+\tau} \mid \phi_{t+\tau})
\Big].
\end{aligned}
\label{eq:distro-FA-3-1-1}
\end{equation}

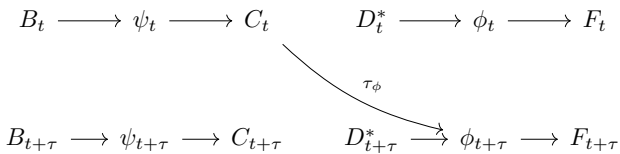
\begin{figure}[ht!]
\centering
\begin{adjustbox}{max width=\columnwidth, center}
\begin{tikzcd}[column sep=1.6em, row sep=7ex, >={Latex}, thick]
B_t \arrow[r] & \psi_t \arrow[r] & C_t &
D^{*}_t \arrow[r] & \phi_t \arrow[r] & F_t  \\
B_{t+\tau} \arrow[r] & \psi_{t+\tau} \arrow[r] & C_{t+\tau} &
D^{*}_{t+\tau} \arrow[r] & \phi_{t+\tau} \arrow[r] & F_{t+\tau} 
%
\arrow[from=1-3, bend right=15, to=2-5,  shorten <=2pt, shorten >=2pt, "\tau_{\phi}"]
\end{tikzcd}
\end{adjustbox}
\caption{Probabilistic Graphical Model associated with the temporal parametrization of diagram~(\ref{d:no_closed}).}
\label{fig:PGM-3-1-1}
\end{figure}

\subsection{Elemental reactions as the lower bound}
At the lower bound we arrive at elemental reactions of the form shown in diagram~(\ref{d:elemental}). Such reactions can be temporally indexed, but they do not by themselves generate endogenous memory, closure, anticipation, or self-maintaining organization. They are therefore useful not because they are agentive, but because they mark the baseline from which the richer partial order departs.

\begin{equation}
\label{d:elemental}
\begin{tikzcd}
& f \arrow[dl, color=black, dashed, thick]  \\
A \arrow[r, color=black, thick] & B 
\end{tikzcd}
\end{equation}

The associated distribution~(\ref{eq:distro-FA-elemental}) is correspondingly simple. It describes a transformation in which \(A_{t+\tau}\), \(f_{t+\tau}\), and \(B_{t+\tau}\) are related within a time-indexed process, but no previous organizational state constrains the current transformation. There is temporal succession, but not yet temporal self-reference. This is why elemental reactions may contribute to the origin of agency without themselves being proto-agents \citep{poole2017chemical}.

\small
\begin{equation}
\begin{aligned}
p(&f_{\tau}, A_{\tau}, B_{\tau} \ldots, f_{T}, A_{T}, B_{T}) = \\ &\prod_{t \ge 0}^{T-\tau} \Big[
    p(A_{t+\tau})
    \, p(f_{t+\tau} \mid A_{t+\tau})
    \, p(B_{t+\tau} \mid f_{t+\tau})\Big].
\end{aligned}
\label{eq:distro-FA-elemental}
\end{equation}

\begin{figure}[ht!]
\centering
\begin{adjustbox}{max width=\columnwidth, center}
\begin{tikzcd}[column sep=1.6em, row sep=7ex, >={Latex}, thick]
A_t \arrow[r] & f_t \arrow[r] & B_t  \\
A_{t+\tau} \arrow[r] & f_{t+\tau} \arrow[r] & B_{t+\tau} 
%
\end{tikzcd}
\end{adjustbox}
\caption{Probabilistic Graphical Model associated with the temporal parametrization of diagram~(\ref{d:elemental}).}
\label{fig:PGM-elemental}
\end{figure}

\subsection{Reading the decomposition as a partial order}
The weakening construction orders diagrams only by explicitly retained relations. Let \(D_1\preceq D_2\) when every organizational relation specified in \(D_1\) is retained in \(D_2\), while \(D_2\) may contain additional relations. This order is generally partial: two diagrams may preserve different relations and therefore be incomparable. It does not by itself specify evolutionary time, taxonomic rank, organismal complexity, or a universal sequence of stages. Biological histories may bypass a regime, externalize one of its constraints, combine relations in a different order, or lose a previously realized relation. The diagrams are consequently candidate organizational regimes, not successive taxa.

Along the focal biological route examined here, the regimes nevertheless support a disciplined interpretation. Elemental reactions provide transformations without cross-time self-reference. Open dependency structures can introduce history-sensitive constraint, while non-closed memory systems can preserve traces without autonomy. Autogenic closure introduces reciprocal constraint maintenance. Informed material causes can support materially embodied memory, whereas free-standing formal causes permit revisable syntactic specification. Situated semantic closure adds sensing and response to an organization that already produces the measurement--control relations through which environmental differences matter. Table~\ref{tab:decomposition-regimes} records these profiles without treating them as rungs on a ladder. Temporal dependency is not sufficient, because every reaction unfolds in time \citep{duenas2019chemistry, bartlett2022provenance}; memory may be externally scaffolded \citep{ehresmann2007memory, wilson2022scaffolding}; closure need not include explicit adaptive coupling \citep{maturana1980, di2005autopoiesis}; and sensors may be externally imposed \citep{cariani1989design, barandiaran2009defining}.

Proto-agency is therefore a profile of precursor conditions rather than a lower taxonomic level. It is supported when a temporally extended chemical organization preserves dependencies that bias later transformations in light of prior states \citep{deacon2014transition, deacon2021molecules}. Autonomy requires the stronger condition that the organization regenerates the constraints through which it continues to exist \citep{mossio2010organisational, moreno2018minimal}. Agency, in the focal sense defended here, further requires a revisable measurement--control interface that modulates organism--environment coupling in light of possible futures \citep{rosen2011anticipatory, fabregas2026organism}. These relations prepare the next section without implying that every actual system must traverse one historical route.


\begin{table*}[t]
\centering
\small
\caption{Organizational regimes obtained by weakening the temporally parametrized \((F,A)\)-system. The order records retained relations; it is neither a universal evolutionary sequence nor a ranking of complexity.}
\label{tab:decomposition-regimes}
\begin{tabular}{p{0.18\textwidth} p{0.22\textwidth} p{0.28\textwidth} p{0.22\textwidth}}
\toprule
\textbf{Regime} & \textbf{Diagrammatic form} & \textbf{Formal signature} & \textbf{Agency interpretation} \\
\midrule
Elemental reaction & Diagram~(\ref{d:elemental}) & Time-indexed transformation without cross-time self-reference & No agency; baseline transformation \\
Open dependency & Diagrams~(\ref{d:no_closed})--(\ref{d:no_closed_2}) & Structured dependency with history-sensitive constraint & Proto-agential precursor; selection can act on patterns \\
Non-closed memory & Diagram~(\ref{d:component1}) & Nested dependency without closure to efficient causation & Memory-like organization without autonomy \\
Autogenic closure & Diagram~(\ref{d:component2}) & Minimal reciprocal constraint maintenance & Candidate minimal autonomy; regeneration must be demonstrated \\
Intrinsic formal memory & Diagram~(\ref{d:hofmeyr_ss}) & Closure supported by informed material causes & Autonomous organization with limited formal revision \\
Semantic closure & Diagram~(\ref{d:hofmeyr_full}) & Closure to efficient causation plus formal openness & Autonomous semantic organization \\
Situated semantic closure & Diagram~(\ref{d:hofmeyr-ext}) & Closure plus sensing-response and revisable measurement-control & Robust primitive agency \\
\bottomrule
\end{tabular}
\end{table*}

\section{Agency Thresholds}
The preceding section derived a structural partial order of temporally parametrized organizations. It now allows us to ask when a proto-agential profile supports an attribution of agency. The answer cannot be obtained by counting components or locating a system on one scalar continuum \citep{newman2025agency}. Closure, viability, anticipation, environmental coupling, and formal reconstruction are distinct organizational conditions that can dissociate, especially in host-dependent, engineered, or collective systems \citep{sultan2022bridging}. Along the focal biological route, autonomy concerns self-maintenance under material openness; goal-directedness concerns viability-biased organization; agency concerns endogenous anticipatory modulation of coupling; and open-endedness concerns reconstruction of the space of possible relevance. The relations among these conditions are represented below as causal-role profiles and a defeasible threshold poset, not as a universal progression.

\subsection{Autonomy as closure under material openness}
Along the focal biological route, autonomy marks the constitutive condition at which a system no longer merely undergoes externally constrained transformations, but participates in producing and maintaining the constraints that make its own continuation possible. In the organizational tradition, this is captured by closure to efficient causation \citep{rosen1991life} or, equivalently, closure of constraints \citep{mossio2017makes}. A system is autonomous when the relevant efficient causes of its organization are not merely imposed from the outside, but are regenerated through the very processes they constrain \citep{Varela1979Principles}. This does not mean that an autonomous system is materially closed. On the contrary, organisms remain thermodynamically open: they exchange matter and energy with their environments and depend on flows they do not produce alone. The distinctive feature of autonomy is therefore not isolation, but the closure of constraint-maintaining relations under conditions of material openness \citep{montevil2015biological}. 

This is why autonomy distinguishes organisms from ordinary dissipative structures. A flame, vortex or B\'enard cell may display spontaneous order, but the constraints that sustain its organization are not themselves produced, repaired and transformed by the organization as a condition of its continued existence \citep{vasas2010lack}. The partial order derived above gives this distinction a formal expression. Elemental reactions and open dependency structures may unfold in time, preserve traces of prior states, or even participate in selective stabilization. Still, they do not thereby become autonomous. Autonomy begins only when a dependency structure closes on itself in such a way that the system maintains the constraints through which it remains a system \citep{deacon2014transition}. In this sense, closure is not already agency, but it is the organizational ground required for the robust biological agency defended along the focal route.

\subsection{Goal-directedness as viability-biased maintenance}
Goal-directedness adds a norm-selective condition that should not be conflated with autonomy. A system may be autonomous without yet exhibiting a rich form of agency, but once autonomy is precarious, its continued existence becomes normatively structured \citep{beer2023theoretical}. Some trajectories preserve the organization, while others lead to breakdown. Goal-directedness can therefore be understood as the tendency of an autonomous organization to maintain, recover or generate viability-supporting conditions \citep{veloz2021goals}. This conception helps avoid two extremes. On the one hand, goal-directedness should not be reduced to an externally imposed target, as in a machine whose goal is assigned by a designer \citep{nicholson2013organisms}. On the other hand, it should not be inflated into conscious intention. Minimal biological goal-directedness is neither external programming nor mental purpose, but the organization-dependent bias by which a system compensates perturbations, preserves its viability and maintains the conditions of its own continuation \citep{heylighen2023meaning}. 

The relevant contrast is between merely reaching a goal and becoming of a goal. As highlighted by~\citet{nicholson2013organisms}, in a non-living machine, the goal is usually specified in advance by an external observer. In contrast, in an autonomous organization, the most primitive goal is generated by the organization itself: to become and remain a viable unity. This is why autopoietic and chemical-organization approaches to goal-directedness are relevant here. They do not begin with a pre-given target, but with the emergence of a self-maintaining network whose continued existence defines what counts as success or failure. Nevertheless, goal-directedness is still not agency in the strong sense defended in this paper. As shown by~\citet{rosenblueth1943behavior}, a system may compensate perturbations and return to a viability-supporting attractor without possessing an endogenous anticipatory organization. Furthermore, a system may be robust, adaptive, and self-maintaining, while still acting primarily through restorative dynamics. The transition to agency requires more than the preservation of a viable organization; it requires the internal modulation of possible couplings in light of what has not yet occurred \citep{jaeger2024naturalizing}.

\subsection{Agency as endogenous anticipatory modulation}
Agency appears when goal-directed organization becomes anticipatory. The crucial point is not that the system predicts in the abstract, since many non-living systems can be described predictively by an observer \citep{clark2013whatever}. As argued by~\citet{juarrero1999dynamics}, the relevant question is whether present action is modulated by an internally generated, history-dependent structure that the organization itself produces, maintains and revises. Agency therefore begins when the system does not merely restore itself after perturbation, but differentially selects among possible organism--environment couplings in light of possible future consequences. The temporal parametrization developed above gives this claim a formal basis. Once the processes of a closed organization are associated with distinct characteristic timescales, the organization induces a dependency structure in which present variables are conditioned by traces of prior organization and by possible future-oriented constraints. This structure is not an externally imposed controller, but the temporal unfolding of the organization itself. Agency, in this sense, is not added to autonomy from the outside; it is autonomy becoming anticipatory through time \citep{nave2025drive}.

This also clarifies why ordinary feedback control is insufficient. A cybernetic regulator, such as a thermostat or a conventional machine, can correct deviations from a target state, but its variables, sensors, effectors and success criteria are externally specified. In biological agency, by contrast, the measurement--control architecture is itself part of the organization that must be produced and maintained \citep{cariani1993evolve, cariani2009strategies}. What is sensed, what is controlled, and what counts as viable are not fixed once and for all. They are historically stabilized and, in stronger forms of agency, revisable \citep{cariani2011semiotics}. The threshold from goal-directedness to agency can therefore be stated as follows. A system is goal-directed when its organization biases trajectories toward viability-supporting states. A system is agentive when that organization contains an endogenous anticipatory structure that modulates coupling before the relevant future state is fully actualized. This is why the ADBN redescription matters: it does not merely represent a stochastic dynamics, but a temporally structured organization in which prior interactions canalize future possibilities \citep{juarrero2023context}.

\subsection{Open-endedness as reconstruction of future possibilities}
Open-endedness marks a reconstructive condition that is stronger than anticipatory modulation alone. A merely adaptive agent may choose among available actions, update parameters or compensate perturbations within an already established repertoire. An open-ended agent can alter the variables, couplings, interpretive roles or formal constraints through which future relevance is defined. This distinction is essential for avoiding a weak conception of adaptation. Many systems adapt by changing parameters within a fixed state space \citep{hernandez2018undecidability}. Living systems, by contrast, often alter the effective space in which adaptation occurs: new sensors, new effectors, new codes, new constraints and new organism--environment relations can emerge historically \citep{cariani1989design}. 

In the language developed here, open-endedness requires not only closure to efficient causation and endogenous anticipation, but openness to formal causation: the capacity to transform the rules, descriptions and distinctions by which the system constructs its own future \citep{ellis2023efficient}. The decomposition profiles in Table~\ref{tab:decomposition-regimes} make the relevant contrasts explicit. Elemental reactions can exhibit temporal succession without self-reference; open dependency structures can be history-sensitive without autonomy; autogenic closure can maintain constraints without rich environmental measurement; and situated semantic closure can support measurement and control without necessarily reconstructing its own formal repertoire. Open-ended agency is therefore supported only when a situated anticipatory organization can revise that repertoire without destroying its organizational continuity.

\subsection{A threshold poset}
\begin{figure*}[t]
\centering
\includegraphics[width=0.96\textwidth]{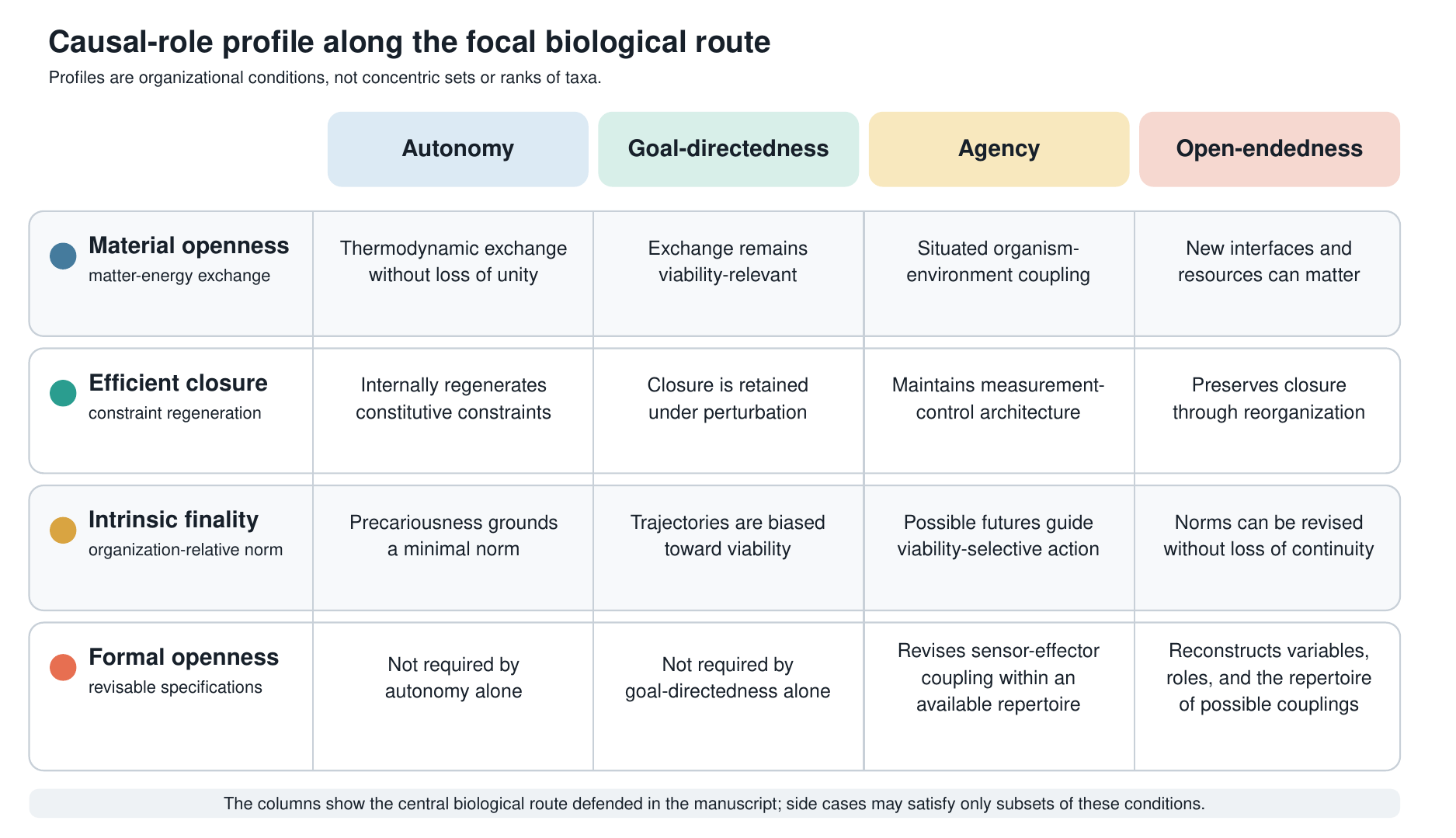}
\caption{Causal-role profiles along the focal biological route. The columns are organizational conditions, not concentric sets or taxonomic ranks. Side cases may realize only subsets of the four causal-role dimensions, and every profile remains relative to a declared scale and target phenomenon.}
\label{fig:causal-role-profile}
\end{figure*}

\begin{figure*}[t]
\centering
\includegraphics[width=0.94\textwidth]{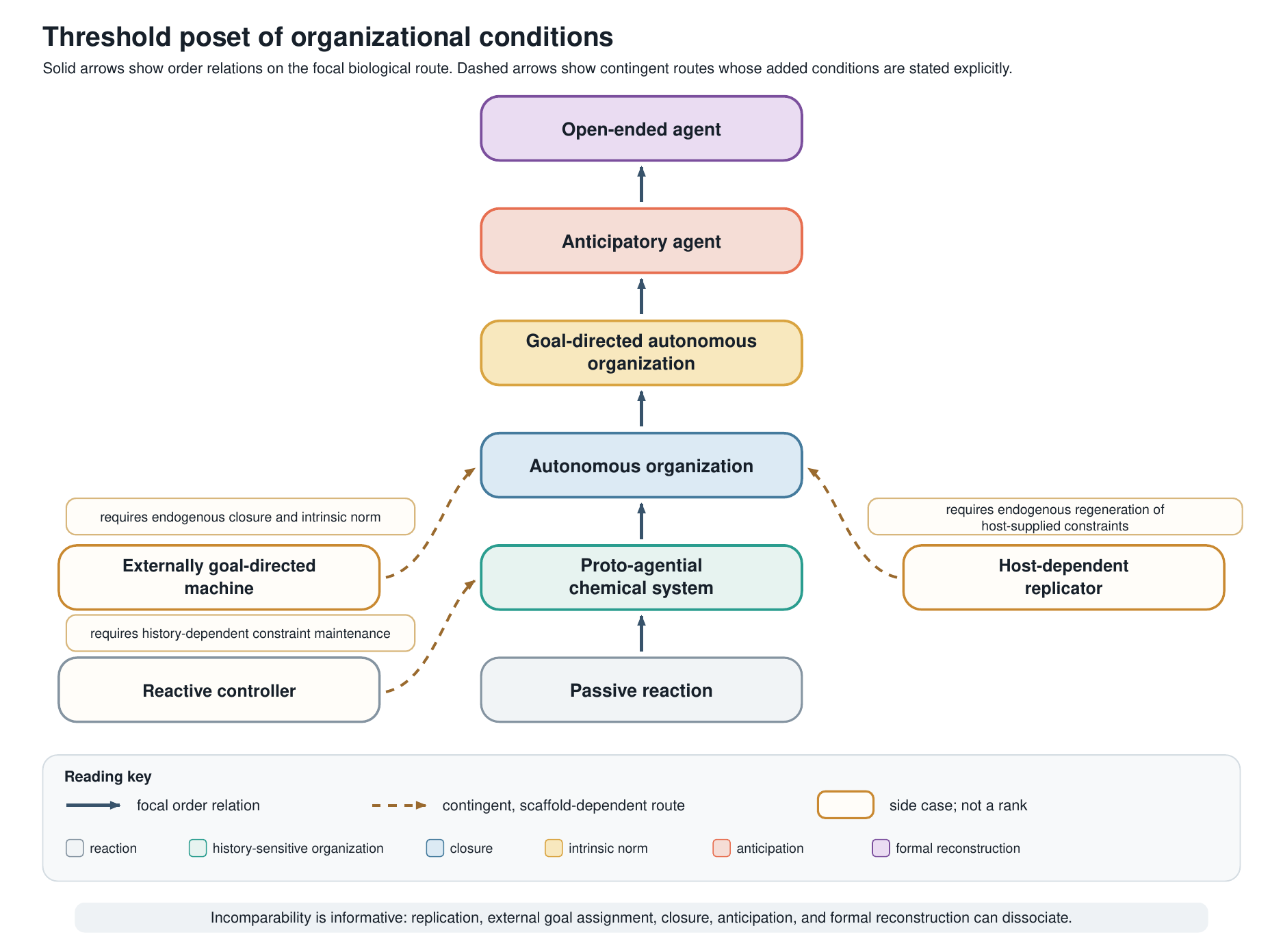}
\caption{Hasse-style diagram of the threshold poset. Solid arrows order specified organizational conditions along the focal biological route; dashed arrows mark contingent, scaffold-dependent routes. Host-dependent replicators and externally goal-directed machines are not forced onto the central route. The diagram is not an evolutionary ladder and makes no claim that every lineage traverses every node.}
\label{fig:threshold-poset}
\end{figure*}

Table~\ref{tab:agency-thresholds} gives operational definitions for the focal biological route. Autonomy is closure to efficient causation under material openness. Goal-directedness is viability-biased maintenance grounded in the organization. Agency is endogenous anticipatory modulation of organism--environment coupling. Open-endedness is reconstruction of the future space within which such modulation occurs. Figure~\ref{fig:causal-role-profile} shows how the four conditions recruit material openness, efficient closure, intrinsic finality, and formal openness without treating the conditions as concentric sets. Figure~\ref{fig:threshold-poset} then represents a defeasible partial order: solid arrows mark the relations defended along the focal biological route, while side nodes and dashed routes display cases in which replication, external goal assignment, closure, or anticipation dissociate. We use \emph{poset} rather than \emph{lattice} because no claim is made here that every pair of profiles has a meet and join. Proto-agential systems occupy profiles with some relevant precursors but do not yet satisfy the strong agency criterion \citep{moreno2018minimal}. A chemical system can preserve memory-like dependencies without autonomy \citep{duenas2019chemistry}; an autogenic system can maintain constraints without explicit sensing--response architecture \citep{deacon2021molecules}; and a semantically closed system can be autonomous without being richly open-ended \citep{lopez2025closing}. Incomparability is therefore a substantive result, not a defect of the taxonomy.

\begin{table*}[t]
\centering
\small
\caption{Proposed thresholds distinguishing autonomy, goal-directedness, agency and open-endedness.}
\label{tab:agency-thresholds}
\begin{adjustbox}{max width=\textwidth}
\begin{tabular}{p{0.18\textwidth} p{0.28\textwidth} p{0.28\textwidth} p{0.18\textwidth}}
\toprule
\textbf{Threshold} & \textbf{Organizational condition} & \textbf{Formal signature} & \textbf{Failure mode} \\
\midrule
Autonomy & The system produces and maintains the constraints required for its own continuation under material openness. & Closure to efficient causation / closure of constraints. & Dissipative order without self-maintained constraints. \\
Goal-directedness & The organization biases trajectories toward viability-supporting states or processes. & Viability-selective compensation and persistence across perturbations. & External goal assignment or mere relaxation to equilibrium. \\
Agency & The system modulates coupling through an endogenous anticipatory structure. & History-dependent ADBN with internally maintained measurement--control dependencies. & Fixed feedback regulation within externally specified variables. \\
Open-endedness & The anticipatory organization can reconstruct its own space of possible relevance. & Rewritable variables, couplings, constraints, formal causes or interpretive roles. & Parameter adaptation inside a fixed state space. \\
\bottomrule
\end{tabular}
\end{adjustbox}
\end{table*}

Moreover, this organizational atlas is intended to answer a central objection to biological agency: if agency cannot be distinguished from ordinary mechanism, predictive regulation or self-stabilization, then it risks becoming an empty label \citep{difrisco2025biological}. The present framework avoids that problem by specifying what agency adds. It does not add an immaterial force, a hidden vital principle, or an observer-dependent metaphor. It adds an organizational condition: the self-produced measurement--control architecture through which the system selects, revises and acts upon relevant variables. Mechanistic explanation remains necessary. The molecular, physiological, and dynamical details still matter, because without them there is no concrete system to analyze. Yet, ordinary mechanism and explanation are insufficient when treating the relevant parts, variables, and transition rules as externally fixed \citep{juarrero1999dynamics}. 

As we will further discuss in the next section, biological agency requires a broader mechanistic picture in which the mechanisms that explain local transformations are themselves embedded in an organization that produces and revises the constraints making those transformations functional for the system \citep{sultan2022bridging, bich2022organization}. We call this a \emph{closure-sensitive mechanistic explanation}: local mechanisms remain indispensable, but the explanatory target also includes how relevant constraints, boundaries, and measurement--control relations are produced and maintained \citep{bich2021mechanism, bechtel2023organisms}. The question is no longer only how a system moves through a predefined space of possibilities, but how an organizationally embedded mechanism helps construct, constrain, and revise that space \citep{jaeger2024naturalizing, bourrat2026agency}. This formulation is continuous with experimental biology and avoids using ``neomechanism'' in a sense that would collide with the established new-mechanist literature.

The conditions proposed here are conceptual so far, but are not intended to remain merely verbal. If autonomy, goal-directedness, agency, and open-endedness are real organizational distinctions, then they should leave empirical signatures. We should be able to ask whether a system maintains its own constraints, whether perturbation compensation is viability-selective, whether present action depends on internally generated future-oriented structure, and whether the system can revise the variables and couplings through which the environment becomes relevant. The next section turns these profiles into an operational research program. The goal is not to claim that all relevant measurements already exist, but to specify what a falsifiable theory of biological agency would require. In particular, we develop a closure-sensitive mechanistic workflow and propose a first metric suite for detecting semantic closure, measurement--control complementarity, anticipatory modulation, affordance reconstruction, syntactic open-endedness, and viability-corrected skill acquisition. 

\section{Operational Signatures of Biological Agency}
The previous section distinguished autonomy, goal-directedness, agency, and open-endedness as nested but non-equivalent thresholds, pointing out to a unification of multiple perspectives in the literature. Still, a conceptual taxonomy is actually not enough. As well-criticized by~\citet{difrisco2025biological}, if agency is to become a useful concept for experimental biology, rather than a metaphor for complex behavior, then the proposed thresholds must be translated into operational criteria that biologists can measure in their labs. The central question is therefore not only what agency is, but how one could detect, compare, perturb, and falsify claims about agency in concrete biological or synthetic systems \citep{walsh2018towards}. This section develops a first step toward such an operational theory. We do not claim that the metrics introduced below already constitute a complete experimental program. Rather, they should be understood as operational handles: preregisterable profile of operational signatures intended to guide comparative experiments. Each signature must be reported with uncertainty, sensitivity to coarse-graining, and null models. Our aim is to specify what should be measured if agency is grounded in self-produced constraints, measurement--control complementarity, endogenous anticipation, and the reconstruction of future possibilities. No single signature is sufficient; the strongest evidence comes from their convergence under intervention.

\subsection{Organizational closure and falsifiability}
The main difficulty in operationalizing organizational closure is that it is not a directly visible component, but a relation among processes, constraints, and levels of description \citep{jaeger2023fourth}. A molecule, membrane, enzyme, receptor, or signaling pathway is not a constraint merely by existing. It becomes a constraint relative to a process it canalizes, a time scale over which it persists, and an organization for which its contribution matters. Closure must therefore be operationalized perspectivally without becoming arbitrary \citep{weckstrom2023natural}. The claim is not that every microphysical process belongs to one closed loop, but that the constraints required to explain a declared target phenomenon are enabled, maintained, or regenerated by the organization whose behavior is being explained.

This is a \emph{closure-sensitive mechanistic explanation}, not a new kind of entity called a ``neomechanism.'' New-mechanist accounts rightly require organized components and activities that produce a phenomenon \citep{nicholson2012concept, craver2021constitutive}. The present proposal adds that, for living explananda, investigators must also test how the relevant components, boundaries, and constraints are produced and maintained over time \citep{leuridan2021diachronic, bich2021mechanism, bechtel2023organisms}. Organizational closure does not replace local mechanism; it specifies when local mechanisms are constitutively embedded in a precarious organization. Nor does closure follow from any single mutual-manipulability experiment. Fat-handed interventions, nonspecific damage, and common causes can create misleading interlevel dependencies \citep{baumgartner2016constitutive, craver2021constitutive}. The appropriate evidential standard is triangulation across interventions, regeneration assays, decompositions, and time scales \citep{wimsatt2007re}.

\begin{figure*}[t]
\centering
\includegraphics[width=0.97\textwidth]{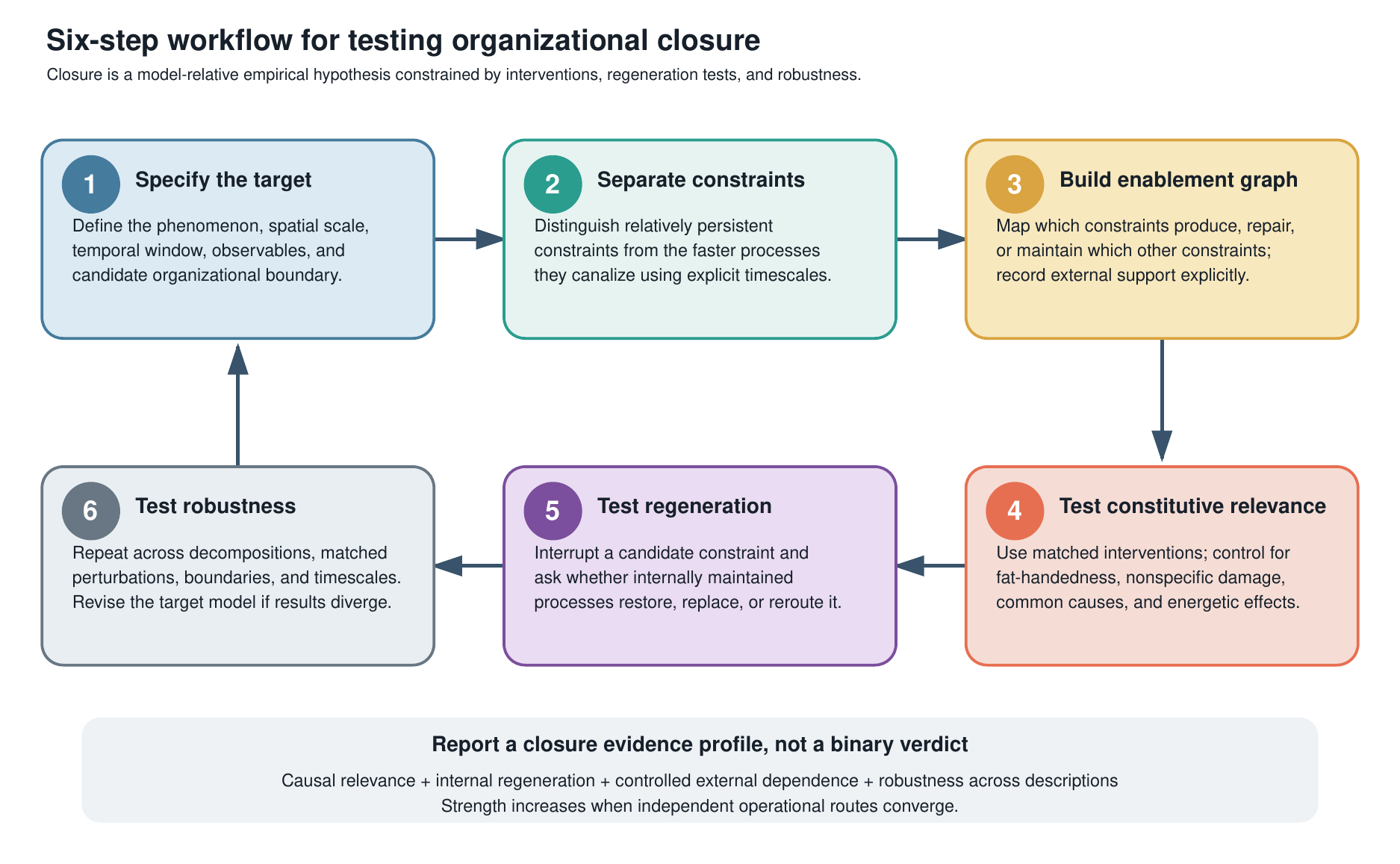}
\caption{Six-step workflow for testing organizational closure. The workflow begins with a declared target and returns to target specification when results depend strongly on boundary, coarse-graining, or time scale. Closure is supported by convergent evidence for constitutive relevance, internal regeneration, controlled external dependence, and robustness; no single intervention is an \emph{experimentum crucis}.}
\label{fig:closure-workflow}
\end{figure*}

We therefore propose the following six-step workflow, summarized in Figure~\ref{fig:closure-workflow}.

\begin{enumerate}[label=\arabic*.]
\item \emph{Specify the target.} Declare the phenomenon, spatial scale, temporal window, observables, viability variables, and candidate organizational boundary. Closure is always a hypothesis relative to this target.
\item \emph{Separate constraints from faster processes.} Identify relatively persistent structures that canalize faster dynamics and estimate the characteristic time scales on which those structures act, decay, and are regenerated.
\item \emph{Construct an enablement graph.} Let vertices denote experimentally identifiable processes or constraints and directed edges denote production, maintenance, repair, or modulation. Record externally supplied dependencies explicitly rather than absorbing them into the background.
\item \emph{Test constitutive relevance.} Perturb candidate constraints using matched interventions and control for fat-handedness, nonspecific damage, common causes, and global energetic effects. Ask whether the intervention selectively changes the target phenomenon in the manner predicted by the enablement graph.
\item \emph{Test regeneration and external dependence.} Interrupt a candidate constraint and determine whether internally maintained processes restore, replace, or reroute it. A closure claim is weakened when the allegedly internal constraint is dispensable, cannot be regenerated under the specified conditions, or is supplied by the experimenter or host.
\item \emph{Test robustness.} Repeat the analysis across plausible decompositions, perturbations, boundaries, and time scales. Report a closure evidence profile---causal relevance, internal regeneration, controlled external dependence, and robustness---rather than a context-free binary verdict.
\end{enumerate}

The workflow also clarifies the relation between state-space approaches and the present framework. State-space descriptions remain useful whenever the relevant variables, transition rules, and boundaries can be treated as fixed over the time scale of interest. Biological agency becomes visible when the organization changes which variables matter, which couplings are relevant, which constraints are maintained, or which future states remain viable \citep{spencer2026agency}. For the metrics below, investigators may identify a target organization \(O\), environmental variables \(E_t\), internal variables \(I_t\), sensor-like variables \(S_t\), effector-like variables \(A_t\), and viability variables \(V_t\), and then infer a time-lagged dependency structure \(\mathcal{G}_t\) using methods appropriate to the system. This graph is an experimentally inferred redescription, not the organism itself. A fixed regulator should principally show parameter adjustment within a stable dependency structure. Evidence for endogenous anticipation is stronger when perturbation induces internally maintained reorganization of sensor--effector relevance while viability is preserved, and when disrupting that organization selectively impairs adaptive action.

\subsection{Semantic Closure Index}
The first proposed metric is the Semantic Closure Index (SCI). Its purpose is to estimate the degree to which a system constructs or maintains its own \emph{interpreters}. In the language of physical biosemiotics, an interpreter is a constraint that connects symbolic or signal-like structures to functional consequences \citep{pattee2013epistemic}. In the language of relational biology, it is an efficient cause whose production is entailed by the organization itself \citep{hofmeyr2018causation}. The SCI therefore asks whether measurement--control relations are internally produced or externally imposed. Let \(\mathcal{I}_O\) be the set of candidate interpreters required for the organization \(O\) to maintain viability-relevant measurement--control relations. Let \(P(i)\) denote the set of processes that produce or maintain interpreter \(i\), and let \(W(i)\geq 0\) denote a weight proportional to the viability-relevance of \(i\), estimated by perturbation, causal influence, or loss of predictive performance, with \(\sum_{i\in\mathcal I_O}W(i)>0\). A first approximation of SCI is

\begin{equation}
    \mathrm{SCI}(O) = \frac{ \sum_{i \in \mathcal{I}_O} W(i)\,\chi_{\mathrm{internal}}(P(i))\,\chi_{\mathrm{functional}}(i) }{ \sum_{i \in \mathcal{I}_O} W(i) }. \label{eq:SCI}
\end{equation}

In eq.~(\ref{eq:SCI}), \(\chi_{\mathrm{internal}}(P(i))=1\) when the processes maintaining \(i\) are internal to the organization under study, and \(\chi_{\mathrm{functional}}(i)=1\) when perturbing \(i\) disrupts the relevant measurement--control relation. Otherwise, \(\chi_{\mathrm{internal}}(P(i))=0\) and \(\chi_{\mathrm{functional}}(i)=0\). The SCI is close to existing attempts to quantify organizational closure, such as~\citet{klyubin2008keep}, but it is narrower; it focuses specifically on closure of interpreters, not merely on cyclic dependence among processes. We can go further and generalize \eqref{eq:SCI} into a closure evidence profile (CEP). For a target phenomenon and scale, let \(n\) be the number of candidate constitutive constraints in the preregistered enablement graph, and let \(e_i\in[0,1]\), for \(i=1,\ldots,n\), quantify the empirical support that constraint \(i\) is (a) causally relevant to the phenomenon, (b) regenerated within the candidate organization, and (c) not persistently supplied by the experimenter or environment. Report the vector
\begin{equation}
\mathbf{CEP} = (e_1,\ldots,e_n),
\end{equation}
rather than collapsing heterogeneous evidence into an weighted binary sum. A closure claim is strengthened when the enablement graph remains robust across reasonable decompositions and when intervention on any indispensable constraint impairs both the target phenomenon and the regeneration of other constraints.

\subsection{Measurement--Control Complementarity}
The second metric is Measurement--Control Complementarity (MCC). Its purpose is to estimate whether sensing and action are mutually embedded in viability-maintaining loops, rather than functioning as separable input--output channels. This distinction is crucial because ordinary machines can possess sensors and effectors without being semantically closed \citep{cariani1998towards}. Biological systems, by contrast, often produce the very architectures through which measurement becomes action and action reshapes future measurement \citep{cariani2009strategies}. Let \(\Delta>0\) be a preregistered prediction lag chosen for the target process and time scale. As suggested by~\citet{salge2017empowerment}, a first approximation to MCC can be defined as the degree to which sensing improves the prediction of future action and action improves the prediction of future sensing, conditional on internal organization and viability:

\begin{equation}
    \mathrm{MCC} = \frac{1}{2} \left[ \mathcal{I}(S_t;A_{t+\Delta}\mid I_t,V_t) + \mathcal{I}(A_t;S_{t+\Delta}\mid I_t,V_t) \right]. \label{eq:MCC}
\end{equation}

In eq.~(\ref{eq:MCC}), $\mathcal{I}$ stands for the conditional mutual information. The overall quantity represents the average intrinsic value or information gain of a system. It supports a measurement--control interpretation only when interventions on $S_t$ and $A_t$ selectively alter downstream viability-relevant behavior and when those channels are themselves maintained by the candidate organization. However, this quantity should not be interpreted as sufficient for agency. High measurement--control complementarity could occur in engineered control systems \citep{salge2017empowerment}. What matters is whether the loop is also internally maintained, viability-relevant, and plastic under perturbation. MCC therefore becomes informative only when paired with the SCI and the Anticipatory Modulation Index. 

\subsection{Anticipatory Modulation Index}
The Anticipatory Modulation Index (AMI) measures whether present action depends on internally generated future-oriented structure above and beyond current stimuli, past state, and externally imposed feedback. Its purpose is to distinguish ordinary predictive regulation from endogenous anticipation. A regulator may predict future states using a fixed model supplied by a designer. An agentive system, in the stronger biological sense, produces and revises the internal organization that makes such future-oriented modulation possible. This magnitude was inspired by (and can be conceived as an operationalization of)~\citet{levin2019computational}'s \emph{cognitive light cone}. Let \(I_t\) denote internal variables that encode or approximate future viability-relevant possibilities, whether directly observed or inferred as latent variables in the ADBN. Let \(H_t\) the relevant history of prior states. A first approximation of AMI is

\begin{equation}
    \mathrm{AMI} = \mathcal{L}(A_t \mid S_t,E_t,H_t,V_t) - \mathcal{L}(A_t \mid S_t,E_t,H_t,V_t,I_t), \label{eq:AMI}
\end{equation}

where \(\mathcal{L}\) is a suitable prediction loss function (e.g., Kullback-Leibler Divergence) that must be chosen based on the experimenter's objectives. According to eq.~(\ref{eq:AMI}), the interventionist test is stronger than the correlational one: perturbing \(I_t\) should selectively impair adaptive action under novel conditions, especially when current stimuli alone are insufficient to explain the behavior. Given the generality of \eqref{eq:AMI}, we propose the following instance. Let $\mathcal L_0$ be cross-validated predictive loss for future viability using present environmental and reactive variables, and $\mathcal L_1$ the loss after adding candidate internal variables $I_t$. Define the predictive gain $G_{\mathrm{pred}}=\mathcal L_0-\mathcal L_1$. Let $G_{\mathrm{int}}$ be the decline in future viability-selective action caused by an intervention that disrupts $I_t$ while matching energetic and nonspecific damage controls. The anticipatory modulation signature is the ordered pair
\begin{equation}
\mathrm{AMS}=(G_{\mathrm{pred}},G_{\mathrm{int}}).
\end{equation}
Notice that both components are required: prediction without causal leverage may be an epiphenomenal correlation, while causal leverage without future-sensitive prediction may be ordinary feedback.

\subsection{Affordance Reconstruction Rate}
The Affordance Reconstruction Rate (ARR) estimates whether the system creates new relevant possibilities rather than merely choosing among predefined options. An affordance, in the present framework, is not simply an environmental opportunity, nor merely an internal representation. It is a relational possibility for action generated by the coupling between organism and environment \citep{hirota2024reality}. Thus, affordance reconstruction occurs when the system changes the variables, boundaries, sensor--effector relations, or constraints through which the environment matters to it \citep{kauffman2021world}. Let \(d(\mathcal{G}_{t+\Delta},\mathcal{G}_t)\) be an appropriate graph distance measuring changes in nodes, edges, causal directions, or functional roles. A first approximation of ARR is

\begin{equation}
    \mathrm{ARR} = \frac{1}{\Delta} \cdot d(\mathcal{G}_{t+\Delta},\mathcal{G}_t) \cdot V_{t:t+\Delta}. \label{eq:ARR}
\end{equation}

In eq.~(\ref{eq:ARR}), the viability term $V_{t:t+\Delta}$ is essential. Without it, random breakdown, noise, or pathological instability could be mistaken for affordance reconstruction. ARR should increase only when graph-level reorganization expands or preserves viable possibilities. This metric is therefore meant to distinguish genuine open-ended relevance construction from mere behavioral variability. Naturally, in certain cases the graph edit distance alone does not distinguish meaningful reconstruction from estimation error or destructive disorganization. For these cases, define an event as a validated affordance reconstruction only when the system acquires a new measurement--control relation that (i) was absent from its previous repertoire, (ii) is reproducible across matched trials, (iii) depends on internally maintained changes, and (iv) expands or reorganizes viable action. Then
\begin{equation}
\mathrm{ARR}=\frac{N_{\mathrm{validated\ reconstructions}}}{\Delta }.
\end{equation}
The accompanying report should specify whether the change concerns kernel parameters, graph structure, or the variable repertoire itself.

\subsection{Syntactic Open-endedness}
The fifth metric, Syntactic Open-endedness (SO), has already been developed and tested elsewhere \citep{lopez2026characterizing}. Its role in the present framework is to capture whether the effective state space, rule space, context space, or phenotypic space of a system expands over time. A fixed dynamical system may generate many trajectories, but it remains bounded by a predefined space of possibilities. Strong open-endedness requires the system to alter the effective rules by which future states become reachable. In the present paper, SO should be interpreted as a measure of formal openness. It does not replace semantic closure or anticipatory modulation. Rather, it asks whether the system can expand the syntactic repertoire through which future organization is explored. 

Syntactic open-endedness should be retained only as an independent, previously defined test of whether novelty continues to exceed the repertoire generated by a fixed finite description. It is evidence about the persistence of novelty, not by itself a measure of agency. Its strongest use here is conjunctive: sustained novelty should coincide with closure, anticipatory modulation, and validated semantic reconstruction. When combined with SCI, MCC, AMI, and ARR, SO helps distinguish systems that are merely adaptive within a fixed space from systems that transform the space in which adaptation occurs.

\subsection{Viability-Corrected Skill Acquisition}
The final metric is Viability-Corrected Skill Acquisition (VCSA). Its purpose is to extend~\citet{chollet2019measure}'s skill-acquisition efficiency beyond predefined artificial tasks. In standard AI evaluation, a system can acquire skill over a fixed benchmark without constructing the task space, preserving its own viability, or revising its own measurement--control architecture. Biological intelligence, however, requires a broader measure that quantifies how efficiently a system acquires new competencies while maintaining the conditions of its own continued existence \citep{pezzulo2026bootstrapping}. Let \(\Delta K\) denote improvement in competence over a family of tasks or affordance relations, \(C_{\mathrm{data}}\) the informational or experiential cost, \(C_{\mathrm{energy}}\) the energetic cost, \(C_{\mathrm{dev}}\) the developmental or structural cost, \(C_{\mathrm{time}}\) the elapsed-time cost, and \(\overline{V}\) the average viability preserved during acquisition. A first approximation of VCSA is

\begin{equation}
    \mathrm{VCSA} = \overline{V} \cdot \frac{ \Delta K }{ C_{\mathrm{data}} + C_{\mathrm{energy}} + C_{\mathrm{dev}} + C_{\mathrm{time}} }. \label{eq:VCSA}
\end{equation}

Eq.~(\ref{eq:VCSA}) preserves the spirit of Chollet's skill-acquisition efficiency while adding the biological dimensions absent from it \citep[p. 41--42]{chollet2019measure}. A system that learns rapidly by destroying its own viability should not count as biologically intelligent in the relevant sense. Conversely, a system that acquires modest skill while preserving and reconstructing its own measurement--control organization may be more biologically agentive than a high-performing but externally specified machine \citep{jaeger2023artificial}. Because energetic, developmental, temporal, and data costs have different units, they must not be added without normalization. For \(j\in\{\mathrm{data},\mathrm{energy},\mathrm{dev},\mathrm{time}\}\), let \(\widetilde C_j=C_j/C_j^{\mathrm{ref}}\) be the dimensionless cost relative to a preregistered positive baseline \(C_j^{\mathrm{ref}}>0\), and let \(\lambda_j\geq0\) be preregistered dimensionless weights satisfying \(\sum_j\lambda_j=1\). Let \(\epsilon>0\) prevent division by zero, and let \(\Delta P_{\mathrm{viable}}\) be the out-of-sample improvement in the prespecified performance or competence measure that is credited only while the system meets the declared viability criterion. Thus, \(\Delta P_{\mathrm{viable}}\) is the viability-qualified version of \(\Delta K\) (for example, \(\overline V\Delta K\) when viability is normalized to \([0,1]\)). Then
\begin{equation}
\mathrm{VCSA}
=
\frac{\Delta P_{\mathrm{viable}}}
{\epsilon+\sum_j\lambda_j\widetilde C_j}.
\end{equation}
The accompanying report should specify the cost vector alongside the scalar score so that conclusions are not artifacts of the weights.

\subsection{How the metrics work together?}
None of these metrics is sufficient on its own. A system may have high MCC without semantic closure, high AMI without self-produced interpreters, high ARR because it is unstable, or high VCSA within a fully externally defined task space. The metrics become meaningful only as a profile. Autonomy is primarily associated with SCI; goal-directedness with viability-selective compensation; agency with the joint presence of SCI, MCC and AMI; open-endedness with ARR and SO; intelligence with viability-corrected skill acquisition across changing affordance spaces. This metric suite therefore turns the taxonomy of the previous section into an empirical research program. It predicts that ordinary machines, standard predictive regulators, biological cells, synthetic living machines, biohybrid systems, and multicellular collectives should occupy different regions of metric space, as constantly suggested by~\citet{sole2019liquid, sole2022evolution, sole2026cognition}. A conventional machine may score high in task skill but low in semantic closure. A self-maintaining cell may score high in closure and viability maintenance but lower in affordance reconstruction. A morphogenetic collective may score high in reconstruction if it can reorganize its measurement--control relations under perturbation while preserving viability.

Furthermore, along with these metrics, the proposed theory is falsifiable in several ways. First, if systems classified as strongly agentive show only parameter tuning within fixed measurement--control graphs, then the claim of endogenous anticipation is weakened. Second, if perturbing the internal anticipatory structure has no specific effect on adaptive action beyond what is explained by external feedback, then the AMI interpretation fails. Third, if candidate interpreters are not internally produced or maintained, then semantic closure has not been demonstrated. Fourth, if apparent affordance reconstruction does not preserve viability, then it should be interpreted as instability rather than open-ended agency. These criteria also prevent inflationary use of agency language. As suggested by~\citet{mitchell2023debate}, a system does not become an agent merely because it is complex, adaptive, predictive, or difficult to model. It becomes agentive to the extent that it produces and revises the measurement--control organization through which it maintains viability and modulates its coupling to possible futures. The purpose of the proposed metric suite is therefore not to assign a single agency score, but to identify the organizational dimensions along which agency can emerge, fail, or become open-ended.

Thus, the operational program developed here remains preliminary. Its value lies in specifying what must be measured if the preceding theory is correct. Before clarifying the status of Markov blankets and active inference, however, it is useful to ask how these operational handles behave in concrete domains where ordinary distinctions between machines, organisms, cells, collectives, and nervous systems become unstable. The next section therefore treats the metric suite not as a finished experimental apparatus, but as a set of proof-of-concept lenses for comparing artificial chemical self-reproduction, unicellular coordination, synthetic living machines, multicellular agency, and neural organization.

\section{Proof of Concept}
Once again, the metric suite introduced above should not be interpreted as an absolute benchmark for biological agency. Its function at this stage is more modest and more useful. It provides a way to ask what would count as evidence for the conditions previously derived. A proof of concept, in this context, is not a decisive experimental verification of the theory, but a test of discriminative power. If the proposed framework is useful, it should allow us to distinguish systems that merely self-organize, systems that self-maintain, systems that reproduce, systems that regulate, systems that reconstruct their organism--environment couplings, and systems whose future possibilities become open-ended. This section considers several domains where such distinctions become experimentally meaningful. The cases are deliberately heterogeneous: abiotic self-reproducing vesicles, \emph{Physarum polycephalum}, biobots, multicellular collectives, and nervous systems. These are not offered as instances of the same kind of agency. On the contrary, their importance lies in the fact that they should occupy different regions of the metric space described in the last section. The goal is therefore not to call all of them agents, but to show how a formal, multidimensional, biologically grounded theory of agency can avoid both inflationary agency talk and reductive dismissal.

\subsection{Abiotic self-reproduction and the lower bound of organization}
A first proof-of-concept domain is provided by fully abiotic systems capable of growth, reorganization, and self-reproduction. Recent work by~\citet{katla2025self} describes polymeric vesicles generated through photo-induced polymerization-induced self-assembly, where feedback between polymerization, degradation, chemiosmotic gradients, and vesicle growth yields a self-reproducing population of synthetic compartments. These systems are especially important for the present theory because they occupy a liminal zone between ordinary chemical self-organization and biological reproduction. They are not cells, they do not possess genuine semantic closure, and they should not be described as full agents. Yet, they exhibit a kind of autonomous material reorganization that is richer than a passive dissipative structure.

In terms developed in the previous section, these vesicles are useful because they force us to separate self-reproduction from agency. As demonstrated by~\citet{katla2025self}, a vesicle population may show persistence, multiplication, and even heritable variation without thereby possessing endogenous anticipatory modulation. SCI should remain low unless the system constructs or maintains interpreters that mediate measurement--control relations, such as those developed by~\citet{soria2026primitive}. MCC may be nonzero if internal gradients and boundary dynamics reciprocally affect future material organization. AMI, however, should remain weak unless present reorganization is demonstrably biased by internally generated variables that approximate future viability \citep{egbert2023behaviour}. Such systems therefore help establish the lower bound of the partial order: organized reproduction and primitive viability are not yet agency, but they may provide the material substrate from which agency-relevant measurement--control relations can later evolve. 

This case also clarifies why the framework does not reduce agency to self-replication. In~\citet{lopez2025closing}, self-replication marked a crucial threshold in the origin of open-ended evolution because it introduced a self-referential loop capable of preserving and varying organization through time \citep{pattee2012evolving}. Nevertheless, the present theory requires a stronger threshold for agency: the organization must not only reproduce or persist, but modulate its coupling to the environment through internally maintained anticipatory structure \citep{machatzke2026dna}. Abiotic vesicles therefore function as a negative-control-like case. They show how far self-organizing chemistry can go, while also clarifying what must be added before agency language becomes justified.

\subsection{Unicellular coordination without a nervous system}
A second proof-of-concept domain is provided by unicellular organisms that coordinate behavior without a nervous system. \emph{Physarum polycephalum} is particularly useful because its adaptive behavior depends on the coupling of physical, biochemical, and mechanical processes distributed across a single cell \citep{beekman2015brainless}. Recent work by~\citet{gyllingberg2026self} suggests that self-sustained calcium oscillations, diffusion, and mechanics can explain diverse behaviors by linking local perturbations to large-scale coordination. This case is valuable because it challenges any account of agency that requires centralized control, nervous representation, or explicit symbolic cognition, such as~\citet{david2008sense} or~\citet{arnellos2015multicellular}.

In the present framework, \emph{Physarum} should be analyzed as a system in which measurement and control are distributed across a deformable, materially continuous body. The relevant question is not whether \emph{Physarum} “has a mind”, but whether its internal oscillatory organization modulates future action in ways that preserve viability under changing environmental conditions \citep{reid2023thoughts}. MCC should be particularly informative here, because sensing and action are not easily decomposed into separate channels. The same mechanochemical processes that register perturbations also participate in movement, growth, and reorganization. In this sense, \emph{Physarum} is a natural test case for whether measurement--control relations can be inferred without presupposing a nervous input--output architecture.

Certainly, MCC should be complemented by metrics such as AMI and ARR. AMI would ask whether present motion or morphological reorganization is better predicted by internal oscillatory states than by current external stimuli alone. ARR would ask whether the organism merely follows predefined attractors or whether it reconstructs the relevant field of possible actions by changing its morphology, internal transport, and boundary relations. A strong result would show that perturbing calcium dynamics or mechanical feedback selectively disrupts adaptive action, even when environmental gradients remain available \citep{alim2017mechanism}. In this way, \emph{Physarum} can help bridge the gap between minimal material self-organization and genuine biological agency.

\subsection{Xenobots and the machine--organism boundary}
A third proof-of-concept domain is provided by \emph{xenobots} and other related synthetic biobots, such as those recently introduced by~\citet{gumuskaya2025morphological}. Xenobots are built from living cells, but they are not programmed in the same sense as conventional robots. Their behavior emerges from the self-organization of cellular components, bioelectric and mechanical coordination, and the developmental capacities of the living substrate \citep{levin2019computational, blackiston2021cellular, seifert2023reinforcement}. This makes them theoretically valuable not because they settle the question of agency, but because they destabilize the inherited distinction between organism and machine \citep{barwich2024rage}.

A conventional machine has externally specified sensors, effectors, task spaces, and success criteria. Even when it learns, it normally learns within a space of variables and goals established by designers \citep{cariani1989design}. A biobot is different if the relevant measurement--control architecture is partly generated, maintained, or reorganized by the biological material itself \citep{cariani2009strategies}. Xenobots therefore invite a more careful vocabulary. They should not simply be called organisms, robots, or agents. They are better treated as candidate \emph{neomachines}: synthetically constructed systems whose agency-relevant properties depend on self-organizing measurement--control relations rather than on fixed syntactic programming alone.

The metrics introduced in the last section provide a way to make this distinction precise. A xenobot may have high task performance while still scoring low on semantic closure if its relevant sensor--effector architecture is externally imposed. It may show goal-directedness if it preserves a stable collective morphology or behavioral attractor under perturbation \citep{kriegman2020scalable}. It becomes more agentive only if present action depends on internally maintained anticipatory organization, and more open-ended only if it can reconstruct its own field of relevant affordances. This is why xenobots should be treated as stress tests for the framework rather than as decisive evidence for it. They force us to ask which parts of their organization are engineered, which are self-produced, and which are transformed by the system's own activity.

This also clarifies the status of “programming” in biological substrates. In ordinary computation, programming consists in specifying syntactic operations over a predefined set of states. In biological organization, by contrast, interventions may modulate bioelectric, mechanical, or developmental conditions that bias the system toward some attractor without fully specifying the trajectory by which that attractor is reached. Such interventions are not irrelevant to agency, but neither do they exhaust it \citep{barwich2024rage}. The key operational question is whether the biological collective can repair, reroute, or reconstruct the measurement--control relations through which the intervention becomes meaningful \citep{cariani1993evolve}.

\subsection{From multicellular agency to ecologies of relational systems}
The transition to multicellularity adds a further complication. Organizational accounts have usually been developed with unicellular organisms in mind, where autonomy and agency are tightly coupled \citep{mossio2017makes}. However, multicellular systems show forms of coordination, plasticity, and collective behavior that do not always satisfy the strict requirements of single-organism closure \citep{bich2019understanding}.~\citet{arnellos2015multicellular} argue that multicellular agency requires a constitutive--interactive closure mediated by a regulatory center, but~\citet{newman2025agency} suggest that such criteria may exclude many systems that display plausible forms of multicellular agency, including placozoans, slugs, and biobots. This tension is exactly where a graded metric space becomes useful.

The present framework can accommodate this disagreement by refusing to treat agency as all-or-nothing. A multicellular collective may be weakly autonomous at the level of the whole while strongly agentive in some interactional dimension, or strongly autonomous in developmental organization while weakly anticipatory in behavior \citep{fulda2023agential}. The question is not whether the collective satisfies a single binary definition, but how closure, viability maintenance, anticipatory modulation, and affordance reconstruction are distributed across levels, scaling from individuals to collectives. Some collectives may inherit measurement--control relations from their constituent cells; others may generate genuinely new collective measurement--control relations not reducible to any cell taken alone.

This suggests a future relational program for multicellularity. Once a taxonomy of semantically closed or partially closed systems has been developed (e.g., \citep[Fig. 4]{hofmeyr2021biochemically}), one can ask what happens when two or more such systems occupy the same environment \citep[p. 25]{cornish2020contrasting}. The coupling of relational entities may produce ecologies of agency in which each system modifies the affordance space of the others \citep{cardenas2018rosennean}. At the multicellular scale, this may take the form of developmental fields, morphogenetic coordination, bioelectric patterning, or tissue-level constraints that reorganize what individual cells can do. The task is not to force multicellularity into a single-cell diagram, but to identify which diagrams become coupled, nested, or transformed when agency-relevant closure appears across scales \citep[p. 13]{hofmeyr2021biochemically}. Additionally to the metrics proposed here, when dealing with multicellular organizations, other relevant quantities such as~\citet{poledna2026towards}'s Distributed Integration or~\citet{krakauer2020information}'s individuality measures must be taken into account. Multicellular systems are therefore not merely illustrative examples; they are a domain in which the theory must show whether it can discriminate degrees and kinds of agency better than ordinary feedback language alone.

\subsection{Neural agency and the semiotics of the brain}
\citet{cariani2001symbols}'s application of Pattee's symbol--matter complementarity principle to brain dynamics makes possible to extend the present framework, cautiously, to nervous systems. Neural spikes can be interpreted as discrete functional switching states, while membrane potentials, fields, and network dynamics provide the rate-dependent physical substrate in which those switching states are embedded \citep{cariani2001symbols}. In this view, the brain is not merely a dynamical system and not merely a computational device, but a self-maintaining semiotic organization in which symbols and dynamics continually constrain one another \citep{pinotsis2025ephaptic}. This interpretation, however, should be updated in light of contemporary neuroscience. 

Recent discussions of causation in neuroscience emphasize that brain explanation cannot be reduced to correlation \citep{ross2024causation}, nor even to a single narrow notion of mechanism \citep{potter2025beyond}. Lesions, stimulation, neuroimaging, electrophysiology, development, plasticity, and organism-level behavior all reveal different aspects of causal organization \citep{siddiqi2022causal}. This pluralism is compatible with the present framework, because relational agency is not expected to appear at one privileged level \citep{noble2012theory, noble2019biological}. Instead, it should appear in the way measurement--control loops are stabilized, revised, and coupled across neural, bodily, and environmental scales \citep{ball2023distinguishes}.

The brain therefore offers a demanding test of the presented theory. If neural agency is reduced to fixed computation, we lose the self-maintaining, developmental, and embodied dynamics that make neural processes biologically meaningful \citep{cariani1999temporal, cariani2001temporal}. If it is reduced to undifferentiated dynamical holism, we lose the functional specificity of symbolic switching, memory, and control \citep{craver2007explaining}. A relational theory of agency should instead ask which neural processes function as interpreters, which constraints are maintained by the organism, how internal anticipatory variables modulate action, and how neural systems reconstruct affordance spaces through learning, development, and plasticity.

It is very important to emphasize that this does not imply that the same closed diagram used for a self-manufacturing cell should be imposed unchanged on the brain or any other multicellular structure. Different scales may require different relational architectures \citep[p. 13]{hofmeyr2021biochemically}. A cell, a tissue, a nervous system, and a social organism may realize different forms of closure and different modes of anticipatory organization. What unifies them is not a single diagram, but a common explanatory demand: to understand how material systems produce, maintain, and revise the constraints through which their future possibilities become meaningful \citep[p. 27]{pattee2013epistemic}. As we did here with extended $(F, A)$-systems, we can derive the temporal parametrizations and joint distributions for weak versions of any relational model, glimpsing the emergence of internal models that guide adaptive action not just in Rosen's original $(M, R)$-systems, but across the entire zoo of potential systems closed to efficient causation \citep[Fig. 4]{hofmeyr2021biochemically}. 

\subsection{Some empirical expectations}
The proposed cases discussed above can be summarized as a set of expected metric profiles. Abiotic self-reproducing vesicles should display material self-organization and perhaps weak viability-like persistence, but low semantic closure and weak anticipatory modulation. Unicellular organisms, such as \emph{Physarum}, should display stronger measurement--control complementarity and possibly nontrivial anticipatory modulation through distributed mechanochemical organization. Xenobots and other biobots should test whether biological collectives can reconstruct sensor--effector relevance under perturbation. Furthermore, multicellular collectives should test whether agency can be distributed across levels without requiring a centralized nervous system. Similarly, neural systems should test whether symbolic switching, embodied dynamics, and organism-level viability can be integrated into a single causal account. Some of the empirical expectations are shown in Table~\ref{tab:proof-of-concept}.

\begin{table*}[t] \centering \begin{adjustbox}{max width=\textwidth} \begin{tabular}{p{3.1cm}p{3.8cm}p{4.1cm}p{4.4cm}} \toprule \textbf{System} & \textbf{Expected organizational status} & \textbf{Most informative metrics} & \textbf{Main experimental question} \\ \midrule Abiotic self-reproducing vesicles & Self-organization and reproduction without full semantic closure & SCI, MCC, SO & Are interpreters internally produced, or is reproduction still driven mainly by externally imposed chemistry? \\ \emph{Physarum polycephalum} & Unicellular distributed coordination without nervous control & MCC, AMI, ARR & Do internal oscillatory variables modulate future action beyond current stimuli? \\ Xenobots / synthetic living machines & Constructed biological collectives at the organism--machine boundary & SCI, AMI, ARR, VCSA & Can the collective repair or reconstruct measurement--control relations under novel perturbations? \\ Multicellular collectives & Coupled relational systems with distributed agency across scales & SCI, ARR, SO & Does agency emerge at the collective level, or only from constituent cells? \\ Neural systems & Multiscale semiotic organization linking switching states and dynamics & MCC, AMI, VCSA & How do neural symbols, bodily dynamics, and viability constraints co-produce action? \\ \bottomrule \end{tabular} \end{adjustbox} \caption{Proof-of-concept domains for the proposed metric suite. The expected profiles are not intended as final classifications, but as hypotheses about how different systems should occupy different regions of the agency metric space.} \label{tab:proof-of-concept} \end{table*}

Table~\ref{tab:proof-of-concept} is not meant to settle the empirical questions in advance. Its purpose is to make the framework testable. If abiotic vesicles score as strongly agentive under these metrics, then the metrics are too permissive. If biological collectives capable of robust reorganization score no differently from fixed regulators, then the metrics are too weak. If neural systems cannot be distinguished from arbitrary dynamical networks, then the framework has failed to preserve the functional specificity of measurement and control. In each case, the theory becomes stronger precisely by exposing itself to possible failure. The proof-of-concept cases considered here show how the operational theory can be applied without overextending it. Abiotic vesicles, \emph{Physarum}, xenobots, multicellular collectives, and neural systems are not placed on a single ladder of increasing intelligence. They are different domains in which closure, viability, anticipation, affordance reconstruction, and semantic plasticity may dissociate. This is why the framework should be understood as a graded organizational theory rather than a universal attribution of agency to all adaptive systems.

Furthermore, the cases above point toward a broader design concept, labeled here as \emph{neomachines}. The term names a proposed class of constructed systems and is unrelated to ``new mechanism'' or ``neomechanism'' in the philosophy of science. A neomachine is not simply a machine made from biological parts, nor a robot with adaptive control. It is a constructed or constructible system whose relevant measurement--control organization is partly self-produced, historically revisable, and viability-sensitive. Such systems may be fully biological, hybrid, chemical, cellular, or eventually artificial. What distinguishes them from conventional machines is not complexity alone, but the degree to which they can generate, maintain, and transform the semantic relations through which their actions matter. Such a transition from our current notion of the machine to neomachines should be conceived as the shift from clocks to engines in the 19th century, or from engines to computers in the 20th century.

This notion also clarifies what current artificial systems lack. A contemporary AI system may acquire impressive skill over symbolic or statistical task spaces while lacking semantic closure, intrinsic viability, and self-produced interpreters. It may adjust parameters, discover correlations, or optimize policies without constructing the organism--environment relation in which success becomes meaningful for itself. The point is not that artificial systems can never become more life-like. The point is that doing so would require architectures that incorporate measurement, control, self-maintenance, and the capacity to revise their own semantic primitives \citep{cariani1989design,cariani1998towards, cariani2009strategies}. 

Our neomachine concept therefore turns the present theory into a design constraint. To build life-like agency, one should not merely add more computation to a passive substrate. One must construct systems in which the substrate participates in the production and repair of the constraints that guide computation. This requires stochasticity, material turnover, measurement--control complementarity, and the possibility of reorganizing sensor--effector relations under perturbation \citep{cariani1993evolve}. In this sense, neomachines are not just technological artifacts; they are experimental tests of whether the organizational principles identified here can be physically realized.

Having clarified the experimental scope of the theory, we can now return to the formal language that has appeared throughout the paper. ADBNs, Markov blankets, and active inference are useful only if their status is carefully restricted \citep{di2022laying}. They do not provide the ontological foundation of agency. They are derived redescriptions of temporal dependency structures already grounded in organizational closure, measurement, and control \citep{nave2025drive}. The final theoretical task is therefore to explain how the present framework can use these tools without inheriting the substance metaphysics and universalizing tendencies often associated with the famous Free Energy Principle \citep{friston2009free, friston2013life, kirchhoff2018markov}.

\section{Markov Blankets and Computational Enactivism Without Substance Metaphysics}
The proof-of-concept cases discussed above clarify the empirical scope of the theory, but they also force a final conceptual distinction. Throughout this paper, we have used the language of asynchronous dynamic Bayesian networks, joint distributions, generative models, Markov blankets, and active inference. These terms are useful only if their explanatory status is carefully restricted. They should not be treated as first principles from which life, autonomy, or agency are deduced \citep{nave2025drive}. Rather, they are derived formalisms: probabilistic redescriptions of temporal dependency structures that become meaningful only after a living organization has already been demarcated in terms of closure, measurement, control, and material self-maintenance \citep{di2022laying}. This restriction is not a minor philosophical caution, but what prevents the present framework from collapsing into the universalizing tendencies often associated with the Free Energy Principle (FEP), introduced by \citet{friston2009free}. The claim is not that any system with a Markov blanket is alive, autonomous, or agentive. Nor is the claim that organisms are literally Bayesian networks. The stronger and more careful claim is that, once a semantically closed organization has been temporally unfolded, its changing dependency structure can be redescribed using probabilistic tools. In this order of explanation, organizational closure comes first; Bayesian description comes second.

\subsection{Closure before inference}
The first principle of the present framework is organizational closure, not active inference. A living organization is not identified because it minimizes free energy, occupies expected states, or possesses a statistical boundary, as in~\citet{ramstead2023bayesian}. In our approach, a living organization is identified because its efficient causes are produced, maintained and transformed by the organization itself, while remaining materially open to the environment \citep{rosen1991life, montevil2015biological, hofmeyr2021biochemically}. Only after this organizational unity has been identified does it become legitimate to ask how its temporal dependencies can be represented probabilistically. This order directly addresses one of~\citet{nave2025drive}'s central worries about Markov-blanket realism. 

If Markov blankets are treated as ontological foundations, they appear to explain autonomy by presupposing the very organization that they are meant to identify \citep{biehl2021technical}. A cell membrane, metabolic network, or sensorimotor boundary is produced and repaired by the activity of the organism; a Markov blanket, by contrast, is a statistical structure defined relative to a model. It cannot, by itself, explain why a living boundary exists, why it persists under material turnover, or why its maintenance matters to the system \citep{nave2022boundaries}. The present framework reverses this explanatory direction. The Markov blanket is not the source of the organism's boundary, but a possible redescription of dependency relations among processes already embedded in a self-maintaining organization \citep{virgo2025good}. This makes it possible to use the formal strengths of Markov blankets without converting them into metaphysical foundations. A blanket may help us describe how internal, sensory, and active variables are conditionally related; it does not explain why those variables belong to a living unity in the first place \citep{clark2017knit}.

The ADBN induced by the temporal parametrization of an $(F, A)$-system (and of any other closed relational model) should be understood in the same restricted way. Following~\citet{korbak2023self}, each constitutive mapping was associated with a characteristic timescale, supporting the idea that constraints and the processes that produce or maintain them operate across different temporal regimes \citep{montevil2015biological}. This temporal unfolding does not add a new substance or hidden cognitive principle to the relational model. It simply makes explicit that a closed organization cannot be operationally understood as a timeless diagram \citep{varela1975calculus}. The resulting ADBN is therefore not the organism's essence, but a formal reconstruction of how the organization constrains its own future states through asynchronous dependencies. Its nodes represent variables associated with components, processes, or mappings; its edges represent conditional dependencies generated by the temporal unfolding of the organization. In this sense, the ADBN is a model of the organization as seen through a probabilistic lens, not a replacement for the relational organization itself.

This distinction matters because the ADBN can be mistaken for a fixed internal model. That would be misleading. The relevant generative structure is not imposed from outside, nor is it assumed as a pre-given veridical model of the world. It is produced by the organization's own history of material interactions, and it is continuously revised as the organization adapts. Thus, the internal model associated with a living organization is not a static representation stored somewhere inside the organism, but a processually maintained dependency structure whose variables, probabilities, couplings, and possibly topology may change as the organization reconstructs its own conditions of existence \citep{guenin2024contextuality}. This is the point at which the present framework can retain something valuable from active-inference language while avoiding its strongest overextensions. A temporally unfolded organizationally closed organization can be described \emph{as if} it were maintaining and revising a generative model of its own organism--environment coupling \citep{virgo2025good}. But this model is not the metaphysical ground of life. It is the probabilistic trace of a more basic organizational fact: the system produces and maintains the constraints through which environmental perturbations become meaningful for its own continuation \citep{fabregas2026organism}.

\subsection{Nested Markov blankets in an unfolded organization}
\citet{kiverstein2022problem} propose that Markov blankets can be used to model sensorimotor autonomy by considering the enabling relations among component processes. In their account, each component process may be associated with its own blanket. Two components stand in an enabling relation when the active states of one component help maintain the sensory states required for the viability of another. If these relations are reciprocally organized, the system as a whole can maintain itself as a precarious unity \citep[p. 9]{kiverstein2022problem}. This proposal is valuable because it shifts attention away from a single global blanket and toward nested relations among component processes. Our temporally parametrized $(F, A)$-system provides a natural candidate for formalizing this idea. Instead of asking whether the organism has one Markov blanket, we can ask which subsets of variables in the ADBN form candidate blankets for particular component processes or closure loops. Given a time-indexed graph $\mathcal{G}_t$ with variables $V_t$, the Markov blanket of a subset $U \subset V_t$ can be written schematically as

\begin{equation}
\operatorname{MB}_{\mathcal{G}_t}(U) = \operatorname{Pa}(U) \cup \operatorname{Ch}(U) \cup \operatorname{Pa}(\operatorname{Ch}(U)) \setminus U , 
\end{equation}

where $\operatorname{Pa}(U)$ denotes the parents of $U$ and $\operatorname{Ch}(U)$ denotes the children of $U$. Notice this definition is purely graphical. It becomes biologically meaningful only when the resulting dependencies correspond to real measurement--control relations in the organization. This gives a precise way to connect~\citet{kiverstein2022problem}'s proposal with relational biology. For each component process, the candidate blanket should not be identified merely by graph topology. It should be tested against the causal and functional role of that process in the organization \citep[p. 45]{korbak2023self}. Does it receive perturbations through genuine measuring structures? Does it affect the rest of the organization through genuine control structures? Does its continued activity contribute to the maintenance of the processes that maintain it? 

If the answer is yes, then the blanket is not merely a statistical convenience; it is a formal redescription of an organizationally relevant interface. The resulting picture is not one global Markov blanket surrounding a pre-given organism, but a heterarchy of nested, partially overlapping blankets associated with the temporal maintenance of component processes. Some blankets may correspond to metabolic loops, others to sensing and response, others to internal fabrication and assembly, and others to higher-order regulatory dependencies. What matters is not that every blanket is ontologically real, but that the pattern of nested dependencies can reveal how the organization maintains its own viability through reciprocal enabling relations \citep{bruineberg2022emperor}.

\subsection{Epistemic cuts, active inference, and anticipatory control}
The above also clarifies the relation between Markov blankets and Pattee's epistemic cuts \citep{pattee2001physics}. A Markov blanket does not, by itself, generate syntax, symbols, or semantic interpretation. It gives an interface of conditional dependence. Syntax, in the biosemiotic sense, requires more: relatively rate-independent symbol vehicles, an interpreter capable of treating differences among those vehicles as differences that matter, and an organization in which such interpretation has functional value \citep{pattee2013epistemic}. Without these ingredients, a blanket is only a statistical partition \citep{williams2022markov}. This point is crucial for avoiding a common inflationary move. One might be tempted to say that because a Markov blanket separates internal from external states, it already captures the epistemic cut. That is too fast. The epistemic cut is not merely a boundary between variables. It is a functional distinction between symbolic vehicles and the rate-dependent dynamics they measure or control. A Markov blanket may help redescribe that distinction once it exists, but it cannot explain the origin of symbolic vehicles, interpreters, or semantic closure \citep{facchin2023extended}.

At most, Markov-blanket language can help formulate the conditions under which syntactic vehicles become stable within a living organization. First, there must be a timescale separation that allows some internal structures to behave as relatively rate-independent tokens with respect to the dynamics they regulate \citep{montevil2015biological}. Second, there must be a dedicated interpretive organization that treats differences among those tokens as functionally relevant \citep{jaeger2024naturalizing}. Third, that interpreter must itself be produced or maintained by the organization \citep{pattee2012cell}. These are not consequences of Markov blankets alone. They are biological conditions that a blanket-based redescription must respect. In this sense, the epistemic cut precedes the Markov blanket in the order of biological explanation. The blanket can help us identify how the cut is embedded in a larger dependency structure, but the cut is grounded in measurement and control, not in conditional independence alone \citep{pattee2012does}. The same point applies to syntax. A system does not acquire syntax because its variables are separated by a blanket. It acquires syntax when discrete, relatively inert vehicles become available for combinatorial control within a self-producing semiotic organization \citep{waters2012neumann}.

Once this restriction is in place, active inference can be used in a more disciplined way. In the present framework, active inference does not define life. It redescribes one possible functional consequence of temporally unfolded closure. A semantically closed organization with sensing and response capacities can use present perturbations to modulate future organism--environment coupling. This is why the extended $(F, A)$-system naturally suggests a primitive anticipatory structure: environmental signals do not merely trigger reactions, but become integrated into a self-maintaining organization whose present dynamics bias future viable trajectories \citep{egbert2023behaviour}. This is not the same as saying that every organism explicitly minimizes free energy. Nor is it necessary to interpret free energy as a universal metaphysical principle governing all self-organizing systems. The safer interpretation is methodological: active-inference formalisms may provide a useful language for describing how some organisms regulate their coupling to the environment under uncertainty \citep{baltieri2025bayesian}. The deeper biological explanation remains organizational. The organism acts in light of possible futures because it is a temporally extended, materially precarious, semantically closed organization whose constraints must be continuously repaired and revised.

Remarkably, Bickhard's interactivism provides an especially close non-Fristonian neighbor to the present proposal \citep{bickhard2009interactivism}. It begins from process naturalism rather than substance metaphysics and distinguishes systems that are merely self-maintaining from systems that are recursively self-maintaining. A recursively self-maintaining system can switch among possible interactions as environmental conditions change. Each interaction carries an implicit presupposition about the conditions under which it will contribute to continued maintenance, and that presupposition can fail. Normativity therefore emerges from the difference between interactions that sustain the far-from-equilibrium organization and interactions that undermine it. This analysis sharpens the transition from goal-directedness to agency proposed here. Viability-biased restoration is not yet sufficient. Agency requires endogenous selection among possible organism--environment couplings on the basis of anticipatory presuppositions whose success or failure matters to the organization itself. The temporally realized $(F,A)$-system supplies the constitutive account that interactivism leaves comparatively schematic: it identifies how the sensors, effectors, symbol vehicles, and constraints enabling such selection are materially fabricated and maintained. Interactivism, in turn, supplies a naturalized account of representational error and normative content without treating an internal model as a static encoding or invoking the Free Energy Principle as a universal law.

The above also clarifies why the internal model in the present theory avoids the circularity criticized by~\citet{nave2025drive}. In many active-inference narratives, the generative model appears already in place, and learning becomes a matter of updating parameters within a pre-established state space. In the present account, the generative structure has an origin. It is chemically, materially, and organizationally generated through closure, then temporally unfolded into a revisable dependency structure. The model is not given first and used to explain life; it is produced by living organization as one of the consequences of temporalized self-maintenance. The strongest version of this claim, however, remains future work. It is not enough to say that the associated ADBN joint distribution changes over time. Genuine open-ended agency requires that the organization can modify more than probabilities. It must be able, at least in principle, to transform variables, alphabets, interpretive roles, measurement repertoires, and coupling topologies. The present framework points toward this possibility through formal openness in $(F, A)$-systems, but a complete mathematical account of topology-changing ADBNs remains to be developed. To be more specific, the stochastic realization introduced in Section \nameref{sec:temporalFA} can change in more than one sense. At each index \(t\), let \(\mathcal M_t=(\mathcal V_t,G_t,K_t)\) denote the stochastic model used to redescribe the organization, where \(\mathcal V_t\) is the repertoire of random variables and their state spaces, \(G_t\) is the directed dependency graph over that repertoire, and \(K_t=\{K_{V,t}:V\in\mathcal V_t\}\) is the collection of conditional Markov kernels. It is useful to distinguish three modes:

\begin{align}
\mathcal M_t &= (\mathcal V_t,G_t,K_t),\\
\text{Mode I: }& K_t\rightarrow K_{t+\tau}\quad\text{with fixed }(\mathcal V,G),\\
\text{Mode II: }& (G_t,K_t)\rightarrow(G_{t+\tau},K_{t+\tau})\quad\text{with fixed }\mathcal V,\\
\text{Mode III: }& (\mathcal V_t,G_t,K_t)\rightarrow(\mathcal V_{t+\tau},G_{t+\tau},K_{t+\tau}).
\end{align}

The arrows denote changes in the stochastic redescription between indices \(t\) and \(t+\tau\), not ordinary state transitions inside one fixed model. ``Fixed'' means equality of the named objects across the comparison interval: \(\mathcal V_t=\mathcal V_{t+\tau}=\mathcal V\) and, where stated, \(G_t=G_{t+\tau}=G\). Mode I corresponds to parameter or posterior updating inside a fixed model class. Mode II corresponds to structural adaptation among an already specified repertoire of variables. Mode III represents reconstruction of the repertoire itself: new measurement channels, effectors, interpretive roles, or state variables become relevant. These modes are alternatives that may occur in different contexts, not ranks in an evolutionary sequence. The present manuscript gives an organizational rationale for all three modes and a standard probabilistic representation for Modes I--II. It does not yet provide a complete mathematics of Mode III. Claims about open-ended reconstruction should therefore be presented as a research program constrained by the relational model, rather than as a result already contained in conventional Bayesian-network formalism. This clarification also restricts the connection to active inference. As we will explain below, a generative model is not postulated as the source of autonomy. Once a self-producing measurement--control organization has been independently identified, probabilistic inference can redescribe how its histories bias future viable trajectories. Relational closure comes first; stochastic redescription comes second.

\subsection{From instrumentalism to constrained realism}
This restricted use of ADBNs, Markov blankets, and active inference also clarifies the sense in which the present framework is \emph{computational enactivist}. The term, coined by~\citet{korbak2021computational}, is potentially misleading because it has been associated with attempts to unify enactivism and the FEP. Our use of the term is different. It does not begin from Fristonian metaphysics, nor does it treat free-energy minimization as the foundation of life or mind. It begins from~\citet{rosen1991life}'s relational biology,~\citet{pattee2001physics}'s symbol--matter complementarity,~\citet{hofmeyr2021biochemically}'s $(F, A)$-systems, and~\citet{cariani1989design}'s account of measurement, control, and emergent semantic functions. The computational side of the framework lies in the fact that living organizations do manipulate relatively discrete, rate-independent structures. They use symbolic vehicles, codes, and syntactic relations to construct, repair, and regulate their own functional organization \citep{waters2021behavior}. The enactive side lies in the fact that these symbolic processes are not detached from material embodiment, thermodynamic precariousness, or organism--environment coupling \citep{clark2017knit}. Computation is not an autonomous layer floating above life, but a functional regime inside a broader architecture of measurement, control, self-production, and adaptive action \citep{cariani2011semiotics}.

This position also avoids the opposite extreme. Purely anti-computational versions of enactivism risk losing the formal structure needed to explain how organisms stabilize memory, coordinate action, and preserve functional distinctions through time \citep{di2022laying}. Conversely, purely computational approaches risk reducing life to syntax and losing the pragmatic dimension that makes symbols matter \citep{pattee2019simulations}. The present framework occupies the middle path: syntactic operations are real and indispensable, but they are biologically meaningful only when embedded in a self-maintaining organization that measures, controls, and reconstructs its own conditions of viability \citep{lopez2025closing}. In this sense, computational enactivism should not be understood as a compromise between two already complete theories, but a reconstruction of both \citep{rodriguez2026every}. Computationalism is corrected by requiring measurement, control, embodiment, and semantic closure. Enactivism is sharpened by providing a formal account of how self-maintaining organization can generate revisable anticipatory structures. The result is neither a Turing-machine theory of life nor a vague appeal to organismic holism, but a process-ontological account of how living systems become capable of meaningful action \citep{dupre2020life}.

There remains a final interpretive question: should the ADBN and its blankets be understood instrumentally, realistically, or something in between? A purely instrumental reading would say that these formalisms are useful descriptions imposed by an observer (e.g.,~\citet{bruineberg2022emperor}). A strong realist reading would say that the organism literally is an ADBN, literally has Markov blankets, and literally performs active inference (e.g.,~\citet{friston2013life}). Both readings are misleading. The present framework requires a constrained realism, such as~\citet{mitchell2023bearable}'s \emph{metaphysical pragmatism} or~\citet{weckstrom2023natural}'s \emph{perspectival realism}. The probabilistic model is observer-constructed, but it is constrained by real organizational relations of measurement, control, and closure. This is similar to the status of the epistemic cut. The cut is not an ontological dualism between matter and symbol, but neither is it arbitrary \citep{pattee2013epistemic}. It reflects the fact that some functional distinctions disappear when everything is reduced to microscopic dynamics \citep{pattee2007laws}. Likewise, the ADBN is not an extra substance inside the organism, but it is not arbitrary either. It is a formal redescription of dependencies generated by temporally organized biological processes. Its adequacy depends on whether it tracks the real functional architecture of the system.

This constrained realism also protects the theory from pan-inferential overreach. Many systems can be described statistically. Many systems can be partitioned into conditionally dependent variables. Many systems can be interpreted \emph{as if} they were minimizing some quantity. None of that is sufficient for agency. Agency requires that these dependencies be embedded in a materially precarious organization whose measurement--control relations help maintain and revise the conditions of its own existence. The ADBN matters only because it is derived from such an organization. The present section has restricted the status of the probabilistic language used throughout the paper, but it has not solved every problem. Three open questions are especially important. 

First, a general method is still needed for identifying biologically meaningful blankets inside a temporally unfolded relational organization. Graph-theoretic blankets are easy to define; biologically meaningful blankets require independent evidence of measurement, control, and closure. Second, a stronger formalism is needed for cases in which adaptation changes not only probabilities but variables, alphabets, interpretive roles, and graph topology. Third, the relation between local component blankets and organism-level agency must be clarified in multiscale systems. These open questions should not be treated as weaknesses of the framework. They are the point at which the theory becomes experimentally and mathematically productive. If Markov blankets are derived formalisms, then the task is not to discover one universal blanket of life, but to identify which dependency structures are produced and maintained by living organizations, how those structures change across timescales, and when such changes become sufficient for autonomy, goal-directedness, agency, and open-endedness. In this way, the framework turns the philosophical debate over active inference into a concrete research program.

The resulting picture is therefore deliberately non-Fristonian. Life is not explained by free-energy minimization. Agency is not explained by the mere possession of a blanket. Cognition is not explained by treating organisms as generic inferential machines. Instead, free-energy, blanket, and inference languages become useful only when subordinated to a more basic account of biological organization: materially open, semantically closed, temporally unfolded, and capable of revising the measurement--control relations through which its future possibilities become meaningful. This completes the theoretical arc of the paper. The earlier sections established why biological agency cannot be reduced to stability, goal-directedness, or mechanism alone. The middle sections derived a structural partial order of organizational conditions and proposed operational handles for testing their distinct profiles. The proof-of-concept section showed how these handles may apply across protocells, unicellular behavior, synthetic living machines, multicellular systems, and neural organization. The present section has clarified the status of the probabilistic and active-inference vocabulary used along the way. What remains is to summarize the theory, its limits, and the research program it opens.

\section{Discussion}
The aim of this paper has not been to add one more definition of agency to an already crowded theoretical landscape, but to show how several concepts that are often conflated can be systematically distinguished once living organization is treated as temporally unfolded semantic closure. The central result is a graded organizational account in which autonomy, goal-directedness, agency, and open-endedness form nested but non-equivalent thresholds. Autonomy arises when a system maintains the efficient causes of its own organization under material openness \citep{rosen1991life, montevil2015biological}. Goal-directedness appears when such precarious organization becomes viability-biased, so that some trajectories count as preserving the system while others count as breakdown \citep{veloz2021goals, heylighen2023meaning}. Agency begins when viability-biased organization acquires an endogenous anticipatory structure that selectively modulates organism--environment coupling \citep{moreno2018minimal, newman2025agency}. Open-endedness begins when that anticipatory organization can reconstruct the very space of future possibilities through which viability, action, and relevance are defined \citep{pattee2019evolved, lopez2026characterizing}. 

This partial ordering helps avoid two symmetrical mistakes. On the one hand, agency should not be identified with any form of self-organization, robustness, or attractor-seeking behavior \citep{jaeger2023fourth, bourrat2026agency}. Many non-living systems display stability, cyclicity, or spontaneous order without producing the constraints that make their own persistence possible \citep{deacon2014transition, mossio2017makes}. On the other hand, agency should not be treated as an all-or-nothing property added mysteriously to already living matter \citep{sultan2022bridging, difrisco2025biological}. The decomposition identifies possible dependency relations: structured chemical dependencies need not entail closure; closure need not entail anticipatory modulation; and anticipatory modulation need not entail open-ended reconstruction. Some biological histories may instantiate these conditions in the focal order, while others may bypass, externalize, or lose them. Agency is therefore neither a primitive gift nor a merely observer-imposed metaphor, but an organizational achievement with identifiable, not universally sequential, precursors.

A second contribution of the paper is to show why temporality is not an optional supplement to relational biology. Relational models are powerful precisely because they capture organizational entailment without reducing living systems to fixed state-space dynamics \citep{louie2013more, hofmeyr2021biochemically}. Yet, the very self-reference that makes closure biologically distinctive cannot be operationally understood outside time \citep{varela1975calculus, reichel2011snakes}. A closed organization must appear alternately as producer and produced, interpreter and interpreted, constraint and constrained process. By assigning characteristic timescales to the mappings of an $(F, A)$-system, we do not abandon relational biology for ordinary dynamics. Rather, we make explicit the temporal form in which closure can persist, fail, repair itself, and become anticipatory \citep{montevil2015biological, korbak2023self}. 

This point is especially important for the status of the ADBN formalism developed throughout the paper. The ADBN is not the organism itself, nor is it a universal law of self-organization. It is a derived redescription of the temporal dependency structure of a semantically closed organization. This restriction allows us to retain the useful language of generative models, Markov blankets, and active inference without treating them as ontological foundations \citep{bruineberg2022emperor, nave2025drive}. The organism is not alive because it has a Markov blanket. Rather, once a living organization has been identified through closure, measurement, control, and self-maintenance, Markov-blanket-like structures may help redescribe how its component processes enable one another across time, allowing us to experimentally track the development and evolution of such organization. 

The paper also clarifies the sense in which this theory is computational enactivist. The computational aspect lies in the fact that living systems depend on relatively discrete, rate-independent structures, symbolic vehicles, and syntactic relations that participate in the construction and regulation of biological function \citep{pattee2001physics, waters2021behavior}. The enactive aspect lies in the fact that these structures are meaningful only within materially precarious organizations that measure, control, act, and reconstruct their coupling with the environment \citep{cariani1989design, pattee2013epistemic}. It should be noted that the existence of both epistemic modes of description (and their interdependence) does not constitute an ontological dualism, but rather an epistemic necessity inherent in our modeling capacities \citep{weckstrom2023natural}. Thus, computation here is neither rejected nor universalized, but placed inside a broader architecture of semantic closure, measurement--control complementarity, and adaptive self-production \citep{rodriguez2026every}.

The operational section translated this organizational profile into a preliminary suite of measurable quantities built upon existing literature: Semantic Closure Index (SCI), Measurement--Control Complementarity (MCC), Anticipatory Modulation Index (AMI), Affordance Reconstruction Rate (ARR), Syntactic Open-endedness (SO), and Viability-Corrected Skill Acquisition (VCSA). These quantities should not be interpreted as fully validated metrics or collapsed into a single agency score. Their function is more modest and more important: they specify what would have to be measured if the preceding theory is correct. In this sense, the metric suite is an answer to~\citet{difrisco2025biological}'s worry that biological agency lacks a proper research program. The proposed theory becomes experimentally meaningful only insofar as it generates testable contrasts among systems that merely self-organize, systems that maintain closure, systems that anticipate, and systems that reconstruct their own future possibilities.

The proof-of-concept cases were introduced in this same spirit. Abiotic self-reproducing vesicles, \emph{Physarum polycephalum}, xenobots, multicellular collectives, and neural systems were not presented as confirmations of the theory. They were used as stress tests and, absolutely, do not constitute an exhaustive list of instances where our theory can be applied. Each case separates features that are often collapsed: reproduction without full semantic closure \citep{katla2025self}, coordination without a nervous system \citep{gyllingberg2026self}, constructed biological organization without ordinary machine programmability \citep{blackiston2021cellular}, multicellular agency without simple reduction to individual cells \citep{newman2025agency}, and neural symbol--dynamics relations without a single privileged causal level \citep{pinotsis2025ephaptic}. The value of these cases is that they show where the proposed metrics could succeed, where they could fail, and where the theory must be refined.

\begin{figure*}[t]
\centering
\includegraphics[width=0.97\textwidth]{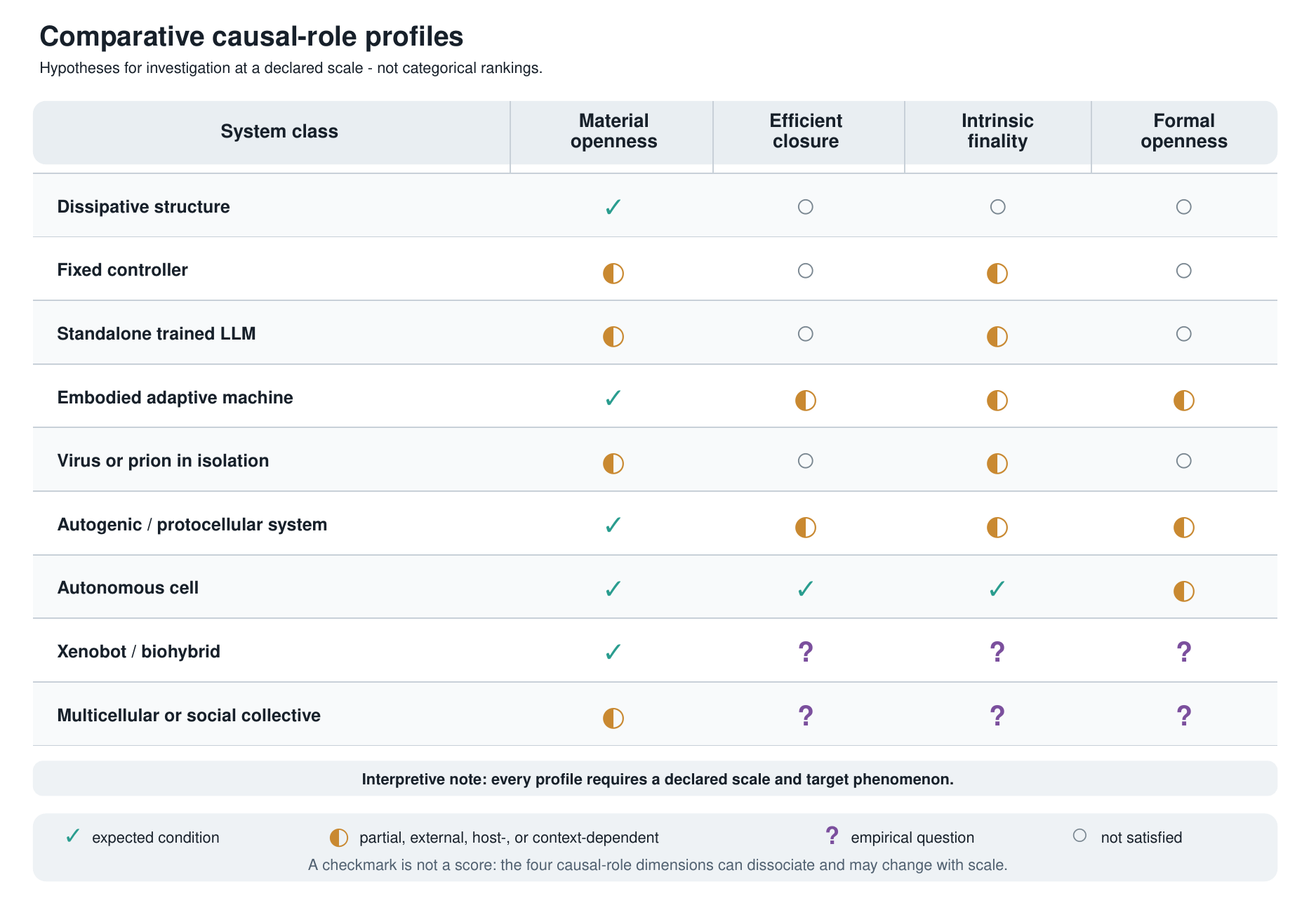}
\caption{Comparative causal-role profiles for biological, host-dependent, collective, and artificial systems. Entries are hypotheses for investigation at a declared scale, not categorical rankings or scores. A partial mark indicates external, host, or context dependence; question marks identify empirical targets rather than deficits inferred in advance.}
\label{fig:comparative-profiles}
\end{figure*}

While the present theory is compatible with several themes associated with developmental plasticity \citep{moczek2011role, beldade2011evolution}, niche construction \citep{laland2016introduction, constant2018variational}, reciprocal causation \citep{svensson2018reciprocal, rama2025reciprocal}, and eco-evolutionary feedback \citep{post2009eco, govaert2019eco}, it should not be presented as a replacement for population genetics or as a universal synthesis. Its contribution is narrower: it offers a relational vocabulary for asking when developmental and ecological feedbacks become constitutive of an agent's own organization. A niche-constructing organism changes selective conditions, but the relevant agency claim depends on whether the altered coupling is measured, controlled, and maintained through the organism's closure. Likewise, a holobiont or multicellular collective should not be assumed to be one higher-level agent merely because its members interact. Agency at the collective scale requires evidence for collective constraint regeneration, an intrinsic viability norm, and measurement--control relations that are not exhausted by the agency of the parts. 

Notably, our framework is substrate-neutral but organizationally demanding. A conventional artifact can exhibit sophisticated control while its constitutive constraints, repair, goals, and semantic interpretation remain externally anchored. A standalone trained large language model, considered as a token-prediction component, does not fabricate or maintain the hardware, energy supply, training regime, sensors, effectors, or viability norms that make its operation possible. It therefore does not satisfy semantic closure or biological autonomy in the present sense. This conclusion does not imply that artificial agency is impossible. An LLM could become one component in a larger embodied organization that maintains its own measurement--control interfaces and materially regenerates the constraints required for continued operation \citep{zhang2025darwin}; the agency claim would then concern the whole organization, not the model alone. 

The same caution applies to societies, cities, nations, and institutions. Such systems can possess memory, coordinated control, and anticipatory planning, but a claim of higher-level agency requires identifying the target phenomenon and demonstrating closure at that scale rather than inferring agency from collective complexity. When seeking a relational diagram that represents such collective organizations, it is important to recognize that their causal architecture would be entirely different from the one analyzed here \citep[p. 13]{hofmeyr2021biochemically}. Systems that construct other agents introduce a further distinction between \emph{productive capacity} and \emph{organizational inheritance}. A factory can build robots without itself being autonomous, while a reproducing organization transmits or reconstructs the constraints through which the offspring becomes capable of its own closure. This distinction should guide future work on multicellularity, synthetic swarms, and artificial systems that manufacture successor systems. Figure~\ref{fig:comparative-profiles} summarizes the resulting hypotheses for different classes of systems without placing them on one scale.

Several limitations follow from this. First, the mathematical account of topology-changing ADBNs remains incomplete. The present paper argues that open-ended agency requires more than changes in probability values within a fixed dependency graph. It requires changes in variables, alphabets, interpretive roles, measurement repertoires, and coupling topologies. However, a full formalism for such transformations has not been developed here. We need to reformulate part of Rosen’s relational biology in modern mathematical terms \citep{zafiris2012rosen, loregian2020rosen}, refining our theoretical tools and developing new experimental benchmarks for agency \citep{baltieri2026mathematical}. The present ADBN construction therefore captures the first step: how temporalized closure can be redescribed as a history-dependent probabilistic structure. The next step is to formalize how such a structure can transform its own space of variables without being reduced to a predefined transition system.

Second, the operational criteria remain perspectival \emph{sensu}~\citet{jaeger2026reengineering}. Any attempt to measure closure, viability, anticipation, or semantic reconstruction requires the investigator to choose observables, timescales, perturbations, and relevant organizational boundaries \citep{mossio2013emergence, weckstrom2023natural}. This does not make the theory arbitrary, but it does mean that empirical applications must be explicit about the explanatory scale at which closure is claimed \citep{winning2020mechanistic, cserban2026rethinking}, as well as looking for alternative ways of explanation in science \citep{juarrero1999dynamics, santos2023emergence}. A proposed closure relation should be treated as a hypothesis about a target organization, not as a metaphysical declaration \citep{winning2020mechanistic}. It is supported when intervention reveals internally maintained enablement relations; it is weakened when the allegedly internal constraint is externally supplied, non-functional, or irrelevant to viability-preserving organization.

Third, the extension beyond unicellularity remains programmatic. The paper has argued that the framework is not intrinsically cell-bound, since different relational diagrams may characterize different scales of self-referential organization \citep{cardenas2018rosennean, hofmeyr2021biochemically}. Yet, it has not supplied a full relational theory of multicellular or neural agency. Such a theory would need to specify how multiple semantically closed or partially closed organizations couple, compete, synchronize, or compose into higher-order units \citep{arnellos2015multicellular, newman2025agency}. In neural systems, it would also need to clarify how symbolic switching states, analog dynamics, embodiment, and organism-level viability form a coherent measurement--control architecture \citep{cariani2001symbols, craver2007explaining, ross2024causation}. These are not small extensions of the present theory, but major future projects to be developed.

Fourth, the framework has not solved the problem of realization \citep{rosen1971some}. It tells us what kinds of organization would be needed for life-like agency, but it does not yet construct such organization from scratch. This distinction is important. A simulation of an ADBN, a digital model of closure, or a Bayesian reconstruction of biological behavior is not yet a realization of agency \citep{pattee2019simulations}. Realization would require physical systems in which measurement, control, material turnover, constraint maintenance, and semantic reconstruction are implemented in the substrate itself \citep{cariani1989design, cariani1993evolve}. This is why the notion of neomachines remains a design constraint rather than an achieved technology. To build life-like agency, one must not merely add computation to matter; one must construct matter organized so that it participates in the production and transformation of the constraints guiding computation \citep{cariani1998towards, cariani2009strategies}.

These limitations also specify what would count as empirical progress. The theory would gain support if systems with internally maintained measurement--control organization display the predicted profile across the relevant metrics, while systems governed by externally imposed constraints do not. It would gain support if perturbations to inferred anticipatory structures selectively impair adaptive action, while perturbations to irrelevant correlations do not. It would gain support if unicellular organisms, biobots, and synthetic protocells occupy distinct regions of the agency metric space in ways that match independently established organizational capacities. Conversely, the theory would be weakened if ordinary dissipative structures, fixed controllers, and purely externally programmed machines become indistinguishable from living systems under the same closure-sensitive tests. Experimental verification of the present theory remains a task for future work.

Remarkably, the most important risk of falsification concerns open-endedness. If all the proposed metrics can be satisfied by systems that merely optimize within a fixed externally specified state space, then the theory has failed to capture what is distinctive about living agency \citep{difrisco2025biological}. A genuinely open-ended agent should not merely choose among options already provided by an external model. It should be able to modify which variables matter, which actions are possible, which affordances are available, and which future states count as viable \citep{longo2012no, erwin2017topology, pattee2019evolved}. The empirical challenge is therefore to distinguish parameter learning from semantic reconstruction, and robust control from the production of new measurement--control relations.

An illuminating, deliberately analogical comparison is Bu's generativism \citep{bu2026generativism}. On that account, generativity is a global profile of generative relations rather than a property inferred from isolated outputs. The notions of \emph{saturability} and \emph{degeneracy} sharpen two distinctions needed here. A system may generate many outputs while remaining inside a saturable profile, and multiple non-injective generative routes may converge on the same output, supporting robustness without enlarging the repertoire of possible organization. In the present framework, open-endedness requires the stronger reconstruction of the measurement--control relations that define the relevant generative profile. The comparison is structural rather than an identification of mathematical frameworks.

Despite these open problems, the framework developed here provides a unified research program. It brings together relational biology, physical biosemiotics, organismic biology, ecological affordances, active-inference formalisms, and process philosophy without reducing any of them to the others. Relational biology supplies the logic of closure; physical biosemiotics supplies the irreducibility of measurement and control; ecological psychology supplies the situated structure of affordances; active inference supplies a restricted formal language for anticipatory regulation; and process ontology prevents the resulting synthesis from hardening into a substance metaphysics. The result is not a final theory of all agency, but a principled way to ask when matter becomes organized so as to act on behalf of its own continued possibility.

\section{Conclusion}
This paper has proposed a temporally grounded, graded theory of biological agency based on the relational model introduced in~\citet{lopez2025closing}. The central claim is that agency is not identical to stability, computation, goal-directedness, or autonomy. 
Autonomy concerns the recurrent regeneration of constitutive constraints under material openness. Goal-directedness concerns viability-biased maintenance. Agency begins when a precarious self-producing organization becomes temporally structured in such a way that its internally generated anticipatory dynamics selectively modulate organism--environment coupling in light of possible futures. Open-endedness begins when the organization can reconstruct the variables, measurement relations, effectors, and norms through which its own future possibilities are defined. To defend this claim, we first argued that self-reference cannot be materially realized outside time. By temporally parametrizing $(F, A)$-systems, we can redescribe semantically closed organizations as history-dependent ADBNs whose joint distributions are not fixed once and for all, but depend on context, substrate, and prior interaction. This does not reduce life to a Bayesian network. Rather, it provides a formal redescription of how a closed organization unfolds, maintains coherence across timescales, and acquires a primitive anticipatory structure.

We then derived weaker temporalized organizations in order to reconstruct a structural partial order from elemental chemical transformations to situated semantic closure. This order distinguishes autonomy, goal-directedness, agency, and open-endedness as defeasible organizational conditions rather than synonyms or universal stages. It avoids both the inflation of agency into any adaptive behavior and the restriction of agency to fully developed cognitive systems. The operational part of the paper proposed a preliminary metric suite intended to make these distinctions empirically tractable. The Semantic Closure Index, Measurement--Control Complementarity, Anticipatory Modulation Index, Affordance Reconstruction Rate, Syntactic Open-endedness, and Viability-Corrected Skill Acquisition are not final measurements of agency. They are operational handles: quantities that indicate what must be tracked if the theory is to become experimentally testable. In this way, the framework attempts to convert biological agency from a suggestive vocabulary into a falsifiable research program, bridging process philosophy and experimental biology \citep{difrisco2026biology}.

Finally, we clarified the status of the probabilistic and active-inference vocabulary used throughout the paper. Markov blankets and active inference do not explain autonomy from first principles. They can redescribe dependency structures within organizations already identified by closure, measurement, and control. Likewise, computational enactivism is not here understood as Fristonian metaphysics, but as a process-ontological synthesis in which computation is embedded within materially grounded, semantically closed, and enactively situated organization. The theory remains incomplete in several respects. A full mathematics of topology-changing ADBNs has not yet been developed. The proposed metrics require calibration in concrete biological and synthetic systems. Multicellular and neural agency require their own relational diagrams, rather than a simple extrapolation from the self-manufacturing cell. Above all, the physical realization of neomachines capable of self-produced semantic reconstruction remains a future experimental goal. Still, these limitations are productive: they specify the next steps required for a general theory of agency to become a working scientific program.

With our approach, the problem of agency has not been solved once and for all, but has been placed on firmer ground. Matter becomes agentive when it is organized so as not merely to persist, but to maintain, interpret, and transform the constraints through which persistence remains possible. In that sense, agency is not a substance added to matter, nor a metaphor imposed by observers, nor a universal property of systems that minimize uncertainty. Agency is a temporal achievement of living organization. We now know that its synthetic realization is, quite literally, “a matter of time.”

\section{Author Contributions}
A.J.L.-D.: formal analysis, visualization, conceptualization, investigation, methodology, validation; C.G.: conceptualization, supervision, funding acquisition, validation.

\section{Acknowledgements}
The authors acknowledge Kate Nave, Johannes Jaeger, Manuel Baltieri, Avel Guénin--Carlut, Michael Levin, Chenyu Bu, Tomek Korbak, Gonçalo Fibra, and Martin Biehl for stimulating conversations and their valuable comments on earlier versions of this manuscript, as well as anonymous reviewers for constructive comments when submitting this work.

\section{Data Accessibility}
This article has no additional data.


\footnotesize
\bibliographystyle{apalike}
\bibliography{example} 

\end{document}